\documentclass[acmsmall,screen,bookmarksnumbered,unicode,colorlinks=true,linkcolor=black,urlcolor=blue]{acmart}
\renewcommand{\authorsaddresses}{}
\renewcommand\footnotetextcopyrightpermission[1]{}
\AtBeginDocument{%
  }

\acmJournal{JACM}

\usepackage{tabularx} 
\usepackage{array} 
\usepackage{longtable}   
\usepackage{booktabs}    
\usepackage{multirow}    
\usepackage{afterpage}   
\usepackage{xurl}        
\usepackage{url} 
\usepackage[table]{xcolor}
\hypersetup{colorlinks=true,linkcolor=black,urlcolor=blue,breaklinks=true}
\makeatletter
\renewcommand{\@dotsep}{2}
\renewcommand*\l@section[2]{%
  \vspace{3.5pt}%
  \@dottedtocline{1}{0em}{2.0em}{\bfseries #1}{#2}%
}
\renewcommand*\l@subsection[2]{%
  \vspace{1.8pt}%
  \@dottedtocline{2}{2.0em}{2.6em}{#1}{#2}%
}
\renewcommand*\l@subsubsection[2]{%
  \@dottedtocline{3}{4.6em}{3.2em}{#1}{#2}%
}
\makeatother

\usepackage{fontawesome5}
\usepackage{makecell} 
\usepackage{threeparttable,threeparttablex} 
\newcolumntype{P}[1]{>{\centering\arraybackslash}p{#1}}
\renewcommand{\arraystretch}{1.05}
\newcolumntype{C}{>{\centering\arraybackslash}X}

\usepackage{enumitem} 
\setlist{itemsep=3pt}

\usepackage{amsthm}   
\usepackage{xcolor}   
\usepackage{epigraph}
\usepackage{mdframed}
\newlength{\qwidth}

\newtheoremstyle{mydef}
  {\topsep}
  {\topsep}
  {\itshape}
  {}
  {\bfseries\color{blue}}
  {.}
  {0.5em}
  {\thmname{#1}\thmnumber{ #2}\thmnote{ (#3)}}

\theoremstyle{definition}
\newtheorem{definition}{Definition} 

\newenvironment{defbox}
  {\begin{mdframed}[roundcorner=5pt, linecolor=black, linewidth=0.5pt]}
  {\end{mdframed}}

\title{Beyond Pipelines: A Survey of the Paradigm Shift toward Model-native Agentic AI}

\author{Jitao Sang}
\authornote{Corresponding author.}
\affiliation{%
  \institution{Beijing Jiaotong University}
  \city{Beijing}
  \country{China}
}
\email{jtsang@bjtu.edu.cn}
\author{Jinlin Xiao}\email{24120395@bjtu.edu.cn}
\affiliation{%
  \institution{Beijing Jiaotong University}
  \city{Beijing}
  \country{China}
}
\author{Jiarun Han}\affiliation{%
  \institution{Beijing Jiaotong University}
  \city{Beijing}
  \country{China}
}\email{23111132@bjtu.edu.cn}
\author{Jilin Chen}\affiliation{%
  \institution{Beijing Jiaotong University}
  \city{Beijing}
  \country{China}
}\email{25125238@bjtu.edu.cn}
\author{Xiaoyi Chen}\affiliation{%
  \institution{Beijing Jiaotong University}
  \city{Beijing}
  \country{China}
}\email{25120333@bjtu.edu.cn}
\author{Shuyu Wei}\affiliation{%
  \institution{Beijing Jiaotong University}
  \city{Beijing}
  \country{China}
}\email{25115081@bjtu.edu.cn}
\author{Yongjie Sun}\affiliation{%
  \institution{Beijing Jiaotong University}
  \city{Beijing}
  \country{China}
}\email{23331011@bjtu.edu.cn}
\author{Yuhang Wang}
\affiliation{%
  \institution{Beijing Jiaotong University}
  \city{Beijing}
  \country{China}
}\email{21112020@bjtu.edu.cn}

\renewcommand{\shortauthors}{Sang et al.}

\begin{document}


\begin{abstract}
The rapid evolution of agentic AI marks a new phase in artificial intelligence, where Large Language Models (LLMs) no longer merely respond but act, reason, and adapt. This survey traces the paradigm shift in building agentic AI: from \textbf{Pipeline-based} systems, where planning, tool use, and memory are orchestrated by external logic, to the emerging \textbf{Model-native} paradigm, where these capabilities are internalized within the model's parameters.

We first position Reinforcement Learning (RL) as the algorithmic engine enabling this paradigm shift. By reframing learning from imitating static data to outcome-driven exploration, RL underpins a unified solution of $LLM + RL + Task$ across language, vision and embodied domains. Building on this, the survey systematically reviews how each capability---\textit{Planning}, \textit{Tool use}, and \textit{Memory}---has evolved from externally scripted modules to end-to-end learned behaviors. Furthermore, it examines how this paradigm shift has reshaped major agent applications, specifically the \textit{Deep Research agent} emphasizing long-horizon reasoning and the \textit{GUI agent} emphasizing embodied interaction. 

We conclude by discussing the continued internalization of agentic capabilities like \textit{Multi-agent collaboration} and \textit{Reflection}, alongside the evolving roles of the system and model layers in future agentic AI. Together, these developments outline a coherent trajectory toward model-native agentic AI as an integrated learning and interaction framework, marking the transition from constructing systems that apply intelligence to developing models that grow intelligence through experience. 

A curated list of the reviewed papers can be found at \faGithub: \url{https://github.com/ADaM-BJTU/model-native-agentic-ai}.
\vspace{0.5cm}
\end{abstract}



\maketitle

\clearpage

\pdfbookmark[1]{Contents}{toc} 
\tableofcontents

\clearpage


\settowidth{\qwidth}{\textit{``I hear and I forget. I see and I remember. I do and I understand.''}}

\setlength{\epigraphwidth}{\qwidth}

\epigraph{\textit{``I hear and I forget. I see and I remember. I do and I understand.''}}
{--- Confucius }

\section{Introduction}
In recent years, the field of Artificial Intelligence (AI) has been dominated by the rapid progress of generative AI, which excels at producing human-like text, images, and other modalities~\cite{chatgpt2022,suno2023,brooks2024video}. Yet, the outputs of generative AI remain largely reactive: it produces content as prompted but does not pursue goals, sustain long-horizon reasoning, or interact with environments. To move beyond passive generation toward autonomous action, researchers have increasingly focused on agentic AI, which emphasizes self-directed behavior, complex reasoning abilities, and environment interaction. The rise of agentic AI is widely regarded as the next stage in the evolution of AI systems.

Across both academia and industry, three core capabilities are consistently highlighted as central to agentic AI:
\begin{itemize}
    \item \textbf{Planning}: to decompose high-level goals into coherent, multi-step strategies.
    \item \textbf{Tool use}: to invoke and coordinate external resources such as APIs, databases, or other models.
    \item \textbf{Memory}: to retain, retrieve, and manage information over extended horizons.
\end{itemize}
The development of agentic AI is deeply intertwined with the evolution of how its core capabilities are implemented, a process that has been undergoing a profound paradigm shift.

\subsection{Paradigms}
\paragraph{\textbf{Pipeline-based Paradigm.}} \hspace{2mm}
Early attempts for constructing agents can be characterized as a ``pipeline'' paradigm, where the agent's core capabilities were largely facilitated by external structures in a workflow-style architecture:
\begin{itemize}
    \item \textbf{Planning} in early systems often relied on external symbolic planners such as PDDL (Planning Domain Definition Language), e.g., LLM+P~\cite{liu2023llmp} generated plan by domain-independent engines rather than by the model itself. Later, this evolved to eliciting reasoning chains directly from the model through prompts like Chain-of-Thought (CoT)~\cite{wei2022chain} and Tree-of-Thought (ToT)~\cite{yao2023tree}, where the model articulates its intermediate thinking process step-by-step.
   
    \item \textbf{Tool use} initially appeared as single-turn functional calls~\cite{openai2023functioncalling}, where the model generated a structured API request parsed and executed by the system. This later advanced to multi-turn frameworks like ReAct~\cite{yao2023react}, which prompts the model to interleave reasoning traces with action calls following an external \emph{Thought-Action-Observation} loop.
    
    \item \textbf{Memory} has long been addressed with external modules. Short-term memory was typically maintained through conversation summary~\cite{langchain:summarymemory}, where long interaction histories were summarized and reinserted to fit within the context window. Long-term memory usually relied on retrieval-based methods, most prominently through Retrieval-Augmented Generation (RAG)~\cite{lewis2020retrieval}, which stored past interactions in a vector database and retrieved them on demand.  
\end{itemize}

In this paradigm, the essential capabilities of agency were not intrinsic to the model, but rather engineered externally and imposed through handcrafted pipelines. We introduce the following formal definition:\vspace{0.5mm}
\begin{samepage}
\begin{definition}[\textbf{Pipeline-based Agentic AI}]\label{def:pipeline}
\begin{defbox}
Pipeline-based agentic AI conceptualizes the agent as a system where a Large Language Model (LLM) serves as a functional component orchestrated by external, handcrafted logic. Formally, the agent's policy $\pi_{\text{agent}}$ is a composite function, where its final action $a$ is constrained and manipulated by an external pipeline (either system workflow or model prompt) $f_{\text{pipeline}}$, acting upon the LLM's internal policy $\pi_{\theta}$: $a = f_{\text{pipeline}}(\pi_{\theta})$. 
\end{defbox}
\end{definition}
\end{samepage}

\vspace{1mm}
While the pipeline-based paradigm offers modularity and a degree of interpretability, its limitations were apparent: such systems heavily rely on carefully engineered pipelines, making them both rigid and brittle. Since the execution logic follows a pre-scripted procedure, these systems struggle to adapt when faced with unforeseen circumstances or dynamically changing environments. Pipeline-based paradigm treats the LLM as a passive, reactive tool rather than a proactive, autonomous decision-maker.

\paragraph{
\textbf{Model-native Paradigm.}} \hspace{2mm}
In response to the limitations of pipeline-based paradigm, agentic AI is undergoing a paradigm shift centered on the principle of ``model-native''. This represents a move away from building complex external agentic systems, toward training a powerful agentic model that effectively becomes the system itself. In the old \textit{pipeline}-based paradigm, an agent was conceptualized as a composite system linked together through prompts or workflow scripts. By contrast, the emerging \textit{model-native} paradigm envisions the agent as a single, unified model that, through end-to-end training, has learned to autonomously perform high-level functions. Planning, tool use, and memory management are no longer external scripts or templates but progressively internalized within the model itself:
\begin{itemize}
    \item \textbf{Planning} was first internalized in OpenAI’s reasoning model \textit{o1}~\cite{openai_o1_2024} through large-scale reinforcement learning, demonstrating the feasibility that LLM could learn to ``think'' and plan autonomously. This line of work was advanced by DeepSeek’s \textit{R1}~\cite{deepseek_r1_2025}, where reinforcement learning with outcome-based rewards was sufficient to train reasoning and planning behaviors, significantly reducing the need for costly, step-by-step supervision.
    
    \item \textbf{Tool use} has likewise moved inside the model. OpenAI’s \textit{o3} model~\cite{openai_o3_2025} exemplified the integration of tool use into reasoning process, learning when and how to invoke a diverse set of tools as part of its internal policy. Moonshot’s \textit{K2}~\cite{kimi_k2_2025} further scaled this direction by synthesizing large-scale tool-use trajectories combined with a multi-stage RL process, strengthening agentic tool use and multi-step decision making.

    \item \textbf{Memory} has also begun to shift from external modules to model-native mechanisms. For short-term memory, Qwen-2.5-1M~\cite{qwen25_1m_2025} leveraged synthetic long-sequence data to extend the native context windows, addressing the challenge of ``remembering'' long-horizon information. Beyond remembering, recent works attempt to effectively ``using'' the information, e.g., MemAct~\cite{zhang2025memact} reframed context management as a tool the agent learns to call, proactively deciding when and how to store or retrieve information based on dynamic state and environmental feedback. For long-term memory, representative works like MemoryLLM~\cite{wang2024memoryllm} pioneered parameterizing memory directly, where a set of latent memory tokens is continuously updated as part of the model's forward pass, resulting in automatically updated internal knowledge. 
\end{itemize}

It is easy to see that, in contrast to the pipeline-based paradigm, this emerging model-native paradigm regards LLMs as autonomous decision-makers which learn to generate plans, invoke tools, and manage memory as intrinsic behaviors. This shift leads to the following definition:\vspace{1mm}\begin{samepage}
\begin{definition}[\textbf{Model-native Agentic AI}]\label{def:model_native}
\begin{defbox}
Model-native agentic AI refers to the paradigm where the core capabilities of agency are progressively internalized within the LLM’s parameters. Formally, the agent's policy is monolithic and synonymous with the LLM's internal policy: $\pi_{\text{agent}} \equiv \pi_{\theta}$. The model directly learns to map a state $s$ to a final action $a$ by optimizing its policy $\pi_{\theta}(a|s)$ to maximize a task-oriented objective. 
\end{defbox}
\end{definition}\end{samepage}

\subsection{Applications}
The paradigm shift in how core agentic capabilities are acquired has also transformed the way agent applications are developed. Currently, agent applications have evolved along two major lines: (1) the \textit{Deep Research agent}, which acts as the ``brain'' and excels at complex reasoning and analysis, and (2) the \textit{GUI agent}, which acts as the ``eyes and hands'' and simulates human interaction with graphical environments. The Deep Research agent is designed for knowledge-intensive tasks, such as writing literature review, conducting market analysis, where long-horizon reasoning, information retrieval, and critical synthesis are essential. By contrast, the GUI agent is well suited to operation-intensive tasks such as automated software testing, workflow automation, where the agent is required to click, type, and manipulate graphical elements with high precision.  

\paragraph{
\textbf{Deep Research Agent.}} \hspace{2mm}
An early form of the Deep Research agent was AI search, exemplified by systems like Perplexity~\cite{perplexity_search}, which constructed an agent pipeline involving steps such as query expansion, information retrieval, and answer generation. Google was the first to introduce a ``Deep Research'' agent~\cite{google_deepresearch}, upgrading the single-turn AI search to a multi-turn, iterative research process. However, early versions of Google's Deep Research agent still relied on a carefully engineered pipeline to manage the multi-turn search and final report generation.

The paradigm shifted with OpenAI's introduction of the first model-native Deep Research agent~\cite{openai_deepresearch}, which was fine-tuned based on its agentic foundation model \textit{o3}. This approach, where the model's internal policy learns to strategize the entire research process, significantly enhanced long-horizon consistency and the depth of information discovery, establishing Deep Research as an advanced assistant capable of tackling rigorous professional tasks. More recently, Tongyi Lab's WebAgent series have advanced the open-source implementation of model-native Deep Research agents. This includes WebSailor~\cite{websailor_2025}, which addressed the critical challenge of synthesizing high-quality trajectory data for uncertain tasks, and culminated in the Tongyi DeepResearch model~\cite{tongyi_deepresearch}, which is capable of executing complex, multi-step research tasks in dynamic web environments. 

Compared with pipeline-based systems, model-native Deep Research agents demonstrate long-horizon consistency, deeper exploration, and greater adaptability to diverse information environments. Nevertheless, two key research challenges remain. First, operating on the open web exposes the agent to pervasive information noise, and outcome-driven RL may amplify hallucinations by rewarding spurious correlations rather than factual grounding. Second, defining rewards for open-ended research tasks is inherently difficult: unlike tasks with verifiable answers, research outputs are judged by subjective qualities such as insightfulness and critical analysis. Developing reward models that capture these nuanced criteria remains a key frontier.

\begin{figure}[t] 
  \centering 
  \includegraphics[width=0.95\textwidth]{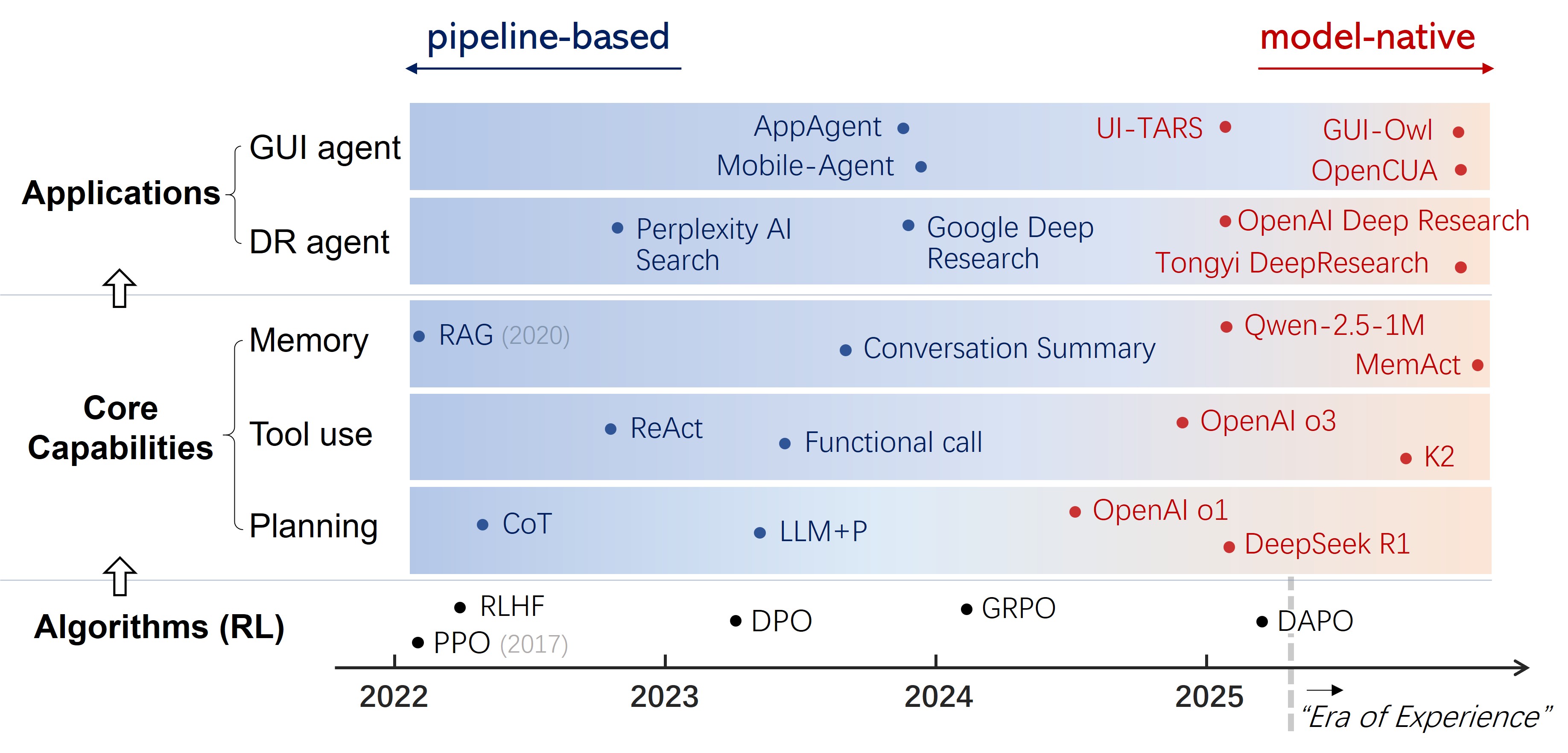} 
  \caption{From \textit{Pipeline} to \textit{Model-native}: RL-driven evolution of agentic capabilities and applications.} 
  \label{fig:1} 
\end{figure}

\paragraph{
\textbf{GUI Agent.}} \hspace{2mm}
Early GUI agents often adopted a pipeline-based approach, built a workflow that orchestrated powerful, closed-source LLMs. For instance, AppAgent~\cite{zhang2023appagent} builds a pipeline where the LLM is orchestrated through XML view-hierarchy information, using structured metadata to perceive UI elements and generate tap/swipe actions. Mobile-Agent~\cite{wang2024mobileagent,wang2024mobileagentv2} developed a multimodal agent pipeline, invoking specialized perception tools such as object detection and OCR to ground UI elements directly from screenshots. These systems typically paired a general-purpose model with specialized tools for perception and action, relying on an external workflow to guide the agent's behavior.

More recently, the trend has shifted towards developing model-native solutions, which internalize perception, planning, grounding, and action execution into a unified policy. UI-TARS~\cite{qin2025ui_tars} represents an early step in model-native GUI agent design. Instead of wrapping a model with orchestrated pipelines, UI-TARS was trained end-to-end to predict low-level actions from visual and UI context under supervised trajectory datasets. GUI-Owl~\cite{ye2025mobileagentv3} and OpenCUA~\cite{wang2025opencuaopenfoundationscomputeruse} advanced this paradigm further by fully internalizing GUI planning and execution via reinforcement learning. By optimizing outcome-based rewards over long horizons, they gained robustness, adaptability, and the ability to decide when and how to act beyond imitation.

Compared to pipeline-based systems, model-native paradigm enables GUI agents to tackle more complex and flexible tasks without brittle external scripts. Nevertheless, model-native GUI agents face unique challenges. First, unlike high-level text-based tasks, their inputs and outputs are inherently fine-grained and low-level, requiring the agent to reason over pixel-level visual cues, widget hierarchies, and precise action sequences such as taps, swipes, and text entries. Small perception or grounding errors can easily cascade into task failure. Second, GUI environments evolve as dynamic interface states, e.g., the same webpage may differ across time due to layout changes, pop-ups, or backend updates. This non-stationarity makes parallel exploration and reinforcement learning particularly difficult, since trajectories collected once may not generalize to later executions of the same task.

\subsection{Algorithms}
A central driver behind the paradigm shift from pipeline-based toward model-native agentic AI is the application of large-scale Reinforcement Learning (RL) in LLM training. Following the release of DeepSeek-R1 technical report~\cite{deepseek_r1_2025}, significant progress has been made in end-to-end RL for LLMs, demonstrating that core agentic capabilities can be acquired through exploration without requiring costly, step-by-step supervision. This highlights the transformative potential of RL for enabling LLMs to evolve their own behavioral policies and adapt to novel environments. Fig.~\ref{fig:1} illustrates the paradigm shift from pipeline-based to model-native agentic AI, contextualized by the evolution of RL algorithms. 

\paragraph{
\textbf{From SFT to RL}} \hspace{2mm} 
Before RL became the focus, Supervised Fine-Tuning (SFT) was the primary method for enhancing LLM capabilities. The premise of SFT is to train the model to imitate a dataset of ground-truth trajectories. However, unlike perception-level tasks such as object classification where humans can easily and reliably provide labels, agentic tasks usually operate at the cognitive and executive level. Real-world agent tasks, such as writing a research report, involve multi-step reasoning, iterative information retrieval before drafting the final report. Constructing complete trajectories for such tasks is prohibitively expensive and, in many cases, practically infeasible for human annotators.

Reinforcement learning elegantly bypasses the need for explicit procedural supervision by fundamentally reframing the learning problem. It shifts the objective from imitating a static dataset of how to act to exploring within a dynamic environment to learn what actions lead to success. This process can be formalized as a Markov Decision Process (MDP): at each step, the model observes a task context (state) and produces an action—a sequence of text or a specific decision. The environment then provides a feedback signal in the form of a reward, reflecting the quality or utility of the model’s action with respect to the overall task. The model's objective is to learn a policy $\pi_\theta$ that maximizes the expected cumulative reward over the task horizon. By learning from the relative value of different trajectories, the model can gradually refine its policy through experience without step-by-step instruction. This transformation is critical because it allows the model to discover novel and potentially more optimal strategies that may not exist in human-curated data, turning it from a passive imitator into an active explorer.

\paragraph{
\textbf{RL for LLMs.}} \hspace{2mm} Early RL methods for LLMs largely treated the model as a static sequence generator, with the objective of aligning the output to human preference data. A representative example is Proximal Policy Optimization (PPO)~\cite{schulman2017ppo} and Direct Preference Optimization (DPO)~\cite{rafailov2023dpo}, which were widely applied in the Reinforcement Learning from Human Feedback (RLHF)~\cite{ouyang2022training} framework. In this setup, a separate reward model is first trained to convert human preferences into numerical rewards, which are then used to fine-tune the LLM via PPO/DPO. Such methods proved highly effective at optimizing single-turn behaviors, like enabling models to better follow instructions and generate content aligned with human values.

Yet, these RL methods are insufficient to train agentic models. Agents operate in multi-turn interactions and dynamic environments, often entailing long-term dependencies and sparse rewards. PPO/DPO and RLHF, which rely heavily on dense and step-level supervision, cannot efficiently optimize policies for long-horizon tasks.

To address these issues, a new set of outcome-driven RL algorithms has emerged to address the practical challenges of training stability and efficiency in long-horizon tasks. To counter the inefficiency of PPO's large critic network, Group Relative Policy Optimization (GRPO)~\cite{zhu2024deepseekmath} was proposed. It introduces a lightweight evaluation paradigm that computes advantages based on the relative rewards within a group of sampled responses, circumventing the need for an absolute value critic and improving training stability. Further refining this, Decoupled Clip and Dynamic sAmpling Policy Optimization (DAPO)~\cite{yu2025dapo} improves performance in multi-turn interactions by decoupling the clipping mechanism for positive and negative advantages and employing dynamic sampling strategies, making it particularly effective for training long-horizon agents.

Following these advancements, other RL innovations~\cite{yang2025mashost, wang2025spa-rl, li2025chain} have continued to emerge, further enabling large-scale, high-efficiency training on LLMs. This progression has culminated in what is now considered a unified training solution: $LLM + RL + Task$, where a base model is enhanced via a RL learning algorithm within a well-defined task environment. Together, these algorithmic advances collectively form the central driver behind the model-native paradigm of agentic AI.  

\subsection{Survey Structure}
The remainder of this survey is organized as follows. Section~\ref{sec:2}  begins by discussing the necessity of RL for training based on the unique characteristics of agent tasks, and compares classical RL with RL for LLMs to establish its feasibility. Section~\ref{sec:3},~\ref{sec:4}, and~\ref{sec:5} will systematically review the paradigm shift from pipeline-based to model-native approaches for each of the three core agentic capabilities: planning, tool use, and memory. Section~\ref{sec:6}, concurrent with the evolution of core capabilities, will examine the evolution of agent applications, specifically outlining the progression of Deep Research agents and GUI agents. Finally, Section~\ref{sec:7} will discuss developing model-native trends for other agentic capabilities, such as multi-agent collaboration and reflection, and explore the evolving distinct roles of the system layer and the model layer in building agentic AI.

\section{Algorithm: RL for LLM}\label{sec:2}

\subsection{Necessity: The Shortage of Procedural Data}
Consider an LLM parameterized by $\theta$, whose base policy can be expressed as a conditional distribution $\pi_\theta(a |q)$, where $q$ denotes the input instruction (query), and $a$ denotes the final answer. In this formulation, the model directly maps an instruction to an answer without modeling intermediate procedural steps. In the following, by analyzing how pipeline-based methods elicit model to generate procedural behaviors, we discuss the necessity of using RL to internalize the agentic capabilities within the LLM.

\paragraph{\textbf{Pipeline as an External Scaffold.}} \hspace{2mm}
Using the planning capability as an example, we can examine how the CoT prompting-based pipeline guides the base policy $\pi_\theta$ to perform multi-step reasoning. 

We first formalize that:
\begin{itemize}
    \item A LLM is trained to generate an answer $a$ given a query $q$, such that $a \sim \pi_\theta(\cdot|q)$.
    \item A CoT prompt consists of $K$ few-shot exemplars, denoted as $E = \{(q^{(i)}, r^{(i)}_{1:T_i}, a^{(i)})\}_{i=1}^K$, where each exemplar consists of an instruction $q^{(i)}$, a reasoning trajectory $r^{(i)}_{1:T_i}$, and a final answer $a^{(i)}$. 
    \item The full prompt fed to the model is a concatenation $x_{\text{prompt}} = [E; q]$.
\end{itemize}

When the model's policy $\pi_\theta$ processes the concatenated prompt $x_{\text{prompt}}$, the probability of generating the subsequent reasoning chain $R=r_{1:T}$ and answer $a$ is the product of the conditional probabilities for each token:
\begin{equation}
    P(R, a | [E; q]) = \prod_{t=1}^{T} P_\theta(r_t |[E; q], r_{<t}) \cdot P_\theta(a |[E; q], r_{1:T}),
\end{equation}
The presence of the exemplar $E$ in the conditioning context creates a strong pattern, significantly increasing the probability of generating a sequence that begins with a reasoning chain $R$ that resembles $r^{(i)}_{1:T_i}$. In other words, CoT prompting externally injects procedural structure into the input, allowing $\pi_\theta$ to mimic multi-step reasoning.  The generated reasoning chain is therefore not an internalized behavior, but rather an elicited one.

\paragraph{\textbf{The Out-of-Distribution Gap.}} \hspace{2mm}
This reliance on in-context pattern matching is precisely why the pipeline-based paradigm is brittle. The model has not learned why the reasoning steps are logical or effective, but only learned that they are textually plausible in the given context. The effectiveness of such handcrafted pipelines is thus limited because large-scale natural corpora, used for pre-training, rarely contain the kind of structured, procedural data that would directly support learning the conditional probability.

Specifically, the conditional distribution invoked by CoT prompting,
$P_\theta(R,a|[E;q])$, often deviates from the distribution that the LLM was exposed to during pretraining. Pretraining primarily optimizes $P_\theta(a|q)$ over naturalistic instruction-answer pairs. Since the model rarely encountered trajectories of the form $(q, r_{1:T}, a)$, the structured prompt context $x_{\text{prompt}} = [E; q]$ is often Out-of-Distribution (OOD)  with respect to the data model was trained.

The model’s ability to follow CoT patterns is therefore fragile: it may succeed when the test query is similar to exemplar queries, but fails to generalize in OOD cases. Without an internalized planning policy, the model may fail to follow the reasoning structure, and produce incoherent reasoning steps and ungrounded answers.

\paragraph{\textbf{The Necessity of RL}}  \hspace{2mm} 
As discussed above, pipelines such as CoT prompting do not update the model's parameters $\theta$. Instead, they manipulate the input sequence to make the desired output pattern more probable. To bridge the OOD gap and create an internalized capability, the model’s parameters must be explicitly optimized.

Still taking planning as the example, we can reframe the planning problem from a probabilistic perspective. Instead of directly mapping a query to an answer, a model with native planning ability will first reason and then answer. This can be viewed as marginalizing the final answer probability over the space of all possible reasoning trajectories $R$:
\begin{equation}
P(a |q) = \int_{R} P(R |q) \, P(a |R, q) \, dR,
\end{equation}
where $R = (r_{1:T})$ denotes the latent reasoning trajectory. To internalize this planning process, the model's policy must learn to model both terms: the policy over reasoning trajectories, $P(R|q)$, and the conditional generation of the answer given the reasoning trajectory, $P(a|R, q)$.

One approach to achieve this is through SFT during post-training. This requires datasets of $(q, R, a)$ triples obtained via either human annotation or data synthesis. 
However, as previously discussed, human annotation of optimal reasoning paths is prohibitively expensive and sometimes even impossible for complex tasks. While, synthetic data generation is ultimately limited by the capabilities of the base model.

RL addresses this by allowing the policy $\pi_\theta$ to 
explore complete reasoning trajectories $\tau$ in an environment, and to update its parameters 
based on outcome-driven rewards $\mathcal{R}(\tau)$ without requiring full procedural supervision. 
Formally, the optimization objective can be expressed as:
\begin{equation}
\theta^* = \arg\max_\theta \; \mathbb{E}_{\tau \sim \pi_\theta} \big[ \mathcal{R}(\tau) \big],
\end{equation}
where $\tau = (r_{1:T}, a)$ denotes a reasoning trajectory consisting of intermediate steps 
$r_{1:T}$ and a final answer $a$.

Furthermore, RL offers two fundamental advantages over SFT for internalizing the agentic capabilities: \hspace{1mm}(1) RL enables a shift from static data feeding to dynamic sample generation. In SFT, the model is a passive recipient of a fixed dataset. In contrast, RL enables adaptive and exploratory learning: the model continuously updates its behavior through interaction, generating new trajectory samples $\tau \sim \pi_\theta$ as its policy $\pi_\theta$ evolves. \hspace{1mm}(2) RL provides a more flexible feedback mechanism, moving from absolute ground-truth fitting to relative value learning. Unlike SFT, which minimizes loss against a single, pre-defined ``correct'' trajectory, RL optimizes for relative outcomes by rewarding trajectories that lead to better task performance even without explicit ground truth. In this sense, RL transforms the model from a passive imitator into an active explorer, providing the fundamental mechanism for the model-native internalization of agentic capabilities.

\subsection{Feasibility: Classical RL \textit{vs.} RL for LLM }
Large-scale pretraining endows LLMs with extensive world knowledge and structured priors, 
which fundamentally reshape how RL can be applied. 
These pretrained priors not only provide a knowledgeable starting point that improves 
\textit{exploration efficiency}, but also supply a \textit{universal interface} for 
representing environments, actions, and rewards across diverse tasks. 
Consequently, RL for LLMs is no longer confined to narrow, domain-specific settings, 
but becomes a general mechanism for internalizing agentic capabilities.

\paragraph{\textbf{Exploration Efficiency: from Random Search to Prior-guided Exploration}}  \hspace{2mm} 
In classical RL, the agent begins with a random policy and 
learns solely through trial and error. The policy $\pi_\theta(a|s)$ is initialized randomly and gradually refined 
by maximizing expected cumulative rewards across trajectories sampled from an 
initially unstructured distribution. 
This process is computationally expensive and sample-inefficient, 
as the agent must repeatedly explore a large amount of low-value or irrelevant states 
before converging to an optimal policy.

By contrast, an LLM pretrained on massive corpora already encodes extensive factual and 
procedural knowledge, which acts as a strong prior over both the state and action spaces.  As a result, the model’s policy is no longer initialized randomly but 
anchored by a structured prior $\pi_{\text{prior}}(a|s, \mathcal{K})$, 
where $\mathcal{K}$ represents world knowledge embedded in the pretrained weights. 
The RL objective thus becomes:
\begin{equation}
\theta^* = \arg\max_\theta \; 
\mathbb{E}_{\tau \sim \pi_\theta(\tau \mid \mathcal{K})} \big[ \mathcal{R}(\tau) \big],
\end{equation}
which can be viewed as fine-tuning the knowledge-conditioned policy 
to better align with task-specific reward signals $\mathcal{R}(\tau)$. 
This prior-guided exploration significantly improves sample efficiency, 
enables meaningful trajectory discovery in early training, 
and prevents the policy from degenerating into random or repetitive behaviors. 
In essence, RL for LLMs does not learn from scratch, 
but rather performs exploratory refinement of a 
pretrained world model through feedback-driven optimization.

\paragraph{\textbf{Generalization across Tasks: Universal Environment, Action and Reward Interfaces}}  \hspace{2mm} 
Classical RL operates within narrowly defined environments, where both the state and action spaces are explicitly specified and the reward function is handcrafted for a single objective. Each policy $\pi_\theta(a|s)$ is trained to optimize performance within this constrained setup, which is tightly coupled to its environment. As a result, classical RL agents, including systems like \textit{AlphaGo}~\cite{alphago_nature2016} and robotic manipulation controllers, are highly specialized and rarely generalize beyond their predefined domains.

By contrast, RL for LLMs operates within an open-ended, 
language-mediated environment where every element of the RL tuple 
is represented through text or symbolic tokens. 
The state $s_t$ corresponds to the evolving textual or multimodal context, capturing task descriptions, retrieved evidence, and interaction history. 
The action $a_t$ is expressed as generated text, tool invocations, or GUI operations, 
serving as a universal and compositional control interface that allows one policy to act across diverse domains. 
Crucially, the reward $\mathcal{R}(\tau)$ can also be defined in flexible and semantic terms: success in reasoning, factual correctness, user preference alignment, or even programmatic verification (e.g., passing unit tests or symbolic proofs).

This language-based representation collapses the boundaries between distinct RL tasks, 
creating a unified state-action-reward interface for different forms of reasoning and interaction. Through this abstraction, LLM serves as both the policy learner and an implicit world model, transforming RL 
from a domain-specific algorithm into a general optimizer, 
bridging reasoning, decision-making, tool use, and memory management 
within one integrated agentic framework. 

\paragraph{\textbf{Unified Solution: LLM + RL + Task}}  \hspace{2mm} 
Recently, it is widely discussed that the development of AI has entered ``the second half''~\cite{yao2025secondhalf}. Rather than designing solutions for specific problems, research now starts from a unified methodology, LLM with RL, and seeks appropriate tasks through which to evaluate and further enhance model capabilities. The proposal of challenging benchmarks has thus become critical for advancing LLMs and agents. This includes  GAIA~\cite{gaia2024} for assessing general-purpose agent capabilities, SWE-Bench~\cite{jimenez2024swebench} for evaluating project-level programming, AndroidWorld~\cite{rawles2024androidworld} for real-world GUI operations, BrowseComp~\cite{wei2025browsecompsimplechallengingbenchmark} for challenging deep research tasks, MCP-Universe~\cite{mcp-universe2025} for evaluating multi-agent collaboration and planning, and most recently, FutureX~\cite{zeng2025futurex} for examining the predictive abilities of LLM agents.

The unified paradigm of model-native agentic model training can be formulated as: 
\begin{equation}
\underbrace{\mathcal{M}_{\text{base}}}_{\text{Base Model}}
+ \underbrace{\mathcal{A}_{\text{learn}}}_{\text{Learning Algorithm}}
+ \underbrace{\mathcal{E}_{\text{task}}}_{\text{Task Environment}}.
\end{equation}
where $\mathcal{M}_{\text{base}}$ provides general world knowledge and reasoning priors, $\mathcal{A}_{\text{learn}}$ (e.g., RL or preference optimization) adapts and refines these capabilities through interaction and optimization, and $\mathcal{E}_{\text{task}}$ defines the environment, tool set, and reward signals that contextualize learning. 

This formula effectively encapsulates the converging trend of recent AI research, particularly in agent training. Within this framework, contributions from the community tend to focus on one of several key facets, which are often deeply interrelated with the design of the ``Task'':
\begin{itemize}
    \setlength{\itemsep}{0pt}
    \setlength{\parskip}{0pt}
    \item \textbf{Data Synthesis}: Creating high-quality, large-scale interaction data to support the RL agent's need for experience, enabling it to learn under diverse constraints. 
    \item \textbf{Reward Function Design}: Crafting sophisticated reward functions for complex tasks where outcomes are not easily verified, often balancing outcome-based rewards with process-based guidance.
    \item \textbf{Environment and Benchmark Construction:} Building the simulation environments and reproducible benchmarks that provide the interactive, verifiable, and challenging scenarios necessary for both training and evaluation.
\end{itemize}
``Task'' ($\mathcal{E}_{\text{task}}$) is not merely the final objective, but defines the entire learning world allowing the \textit{Learning Algorithm} ($\mathcal{A}_{\text{learn}}$) to effectively optimize the \textit{Base Model} ($\mathcal{M}_{\text{base}}$).

This trend toward a unified methodology is quite similar to the development of classical physics. Before Newton, physical subfields were fragmented: Celestial mechanics was governed by Kepler's laws describing planetary motion, Terrestrial mechanics was pioneered by Galileo's studies of falling bodies, and fields like Optics and Fluid dynamics had their own disparate empirical and geometric principles. Newton's \textit{``Three Laws of Motion''}, \textit{``Law of Universal Gravitation''}, together with the powerful mathematical tool of \textit{Calculus}, provided a universal framework that unified these experimental research under a single set of principles, achieving an unprecedented unification for Physics.

A similar dynamic can be observed in AI today with the emergence of the ``LLM + RL'' methodology. The LLM provides a unified model of world knowledge and foundational reasoning, analogous to a set of foundational principles. Meanwhile, RL offers a dynamic, goal-oriented optimization framework, analogous to a general-purpose problem-solving engine. 

Following this parallel, the rise of this methodology ``singularity'' is shifting the focus of AI research, much as Newton shifted physics. In physics, this led to an expansion of applications into areas like fluid dynamics and celestial mechanics, a new focus on enhancing capabilities to solve complex challenges like the many-body problem, and eventually the theoretical innovations such as Quantum theory. Similarly, the practical focus for AI now ranges from the application expansion into practical domains like healthcare, scientific discovery, and social system simulation, to the capability enhancement required for challenges such as continual learning, safety, and alignment, and ultimately toward possibly new theories on general intelligence.

\subsection{A Data Synthesis Perspective}
\paragraph{\textbf{Compute–Intelligence Conversion in AI Development}}  \hspace{2mm}  To further understand the role of RL in training agentic models, it is useful to review the evolution of AI development. With computational power growing at a rate of $3\text{-}4x$ per year, the total increase in compute over the last two decades has been on the order of trillion-fold, making compute a non-negligible driving force. An intriguing view holds that, an underlying thread of AI development over this period has been the effort to \textit{convert the ever-growing compute into intelligence gains with maximum efficiency}~\cite{chung2024dontteachincentivize}. 

\begin{figure}[t] 
  \centering 
  \includegraphics[width=0.9\textwidth]{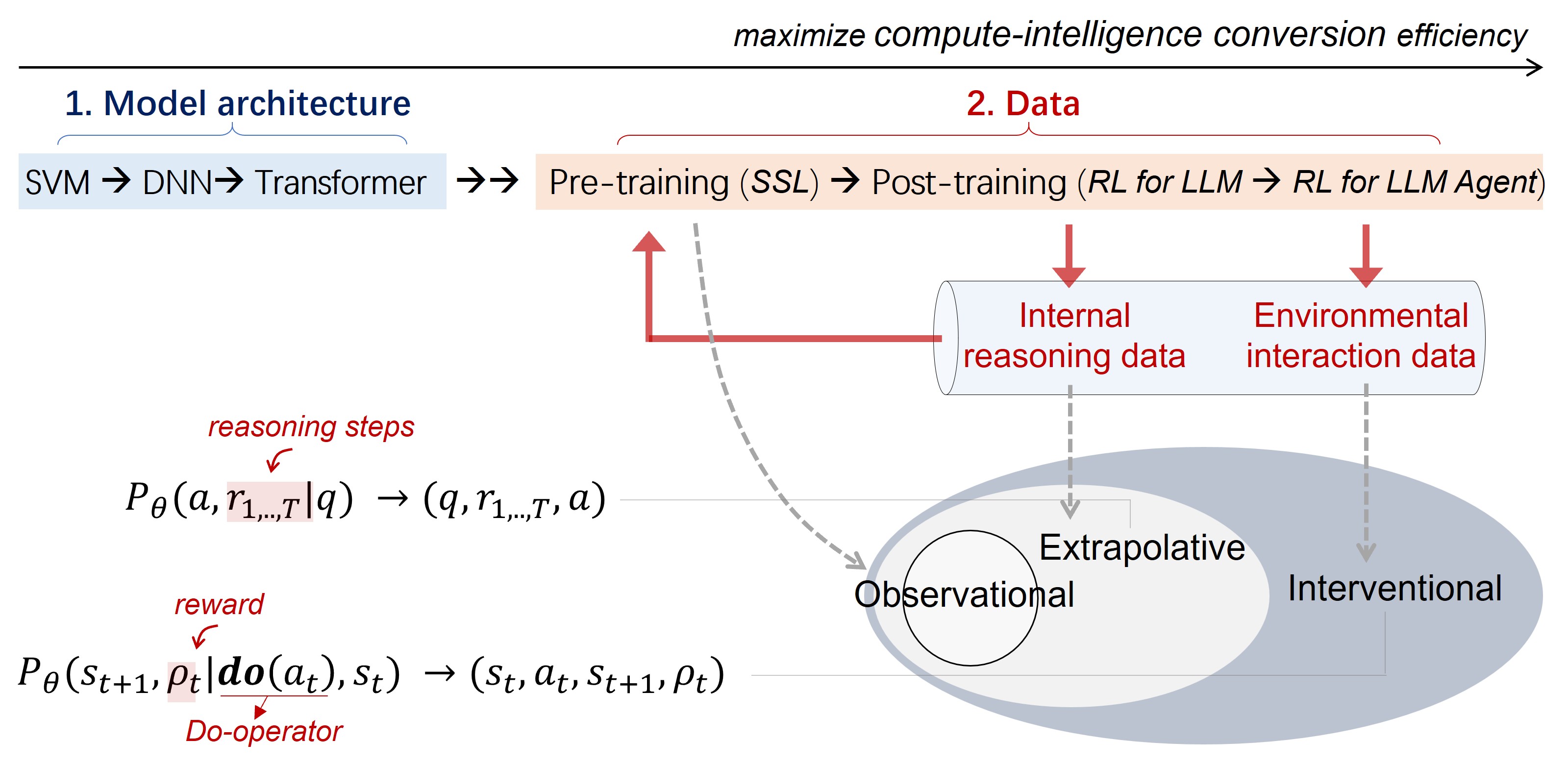} 
  \caption{RL-driven data synthesis as the engine of compute-intelligence conversion.} 
  \label{fig:3} 
\end{figure}

As illustrated in the upper half of Fig.~\ref{fig:3}, this thread has unfolded in two major stages:
\begin{itemize}
   \item \textbf{Model architecture advancement}. \hspace{1mm} During the first ten years (approx. 2010-2020), progress was mainly driven by architectural innovations that enabled the efficient consumption of massive compute and data. The evolution from Support Vector Machines (SVMs)~\cite{cortes1995support} to Deep Neural Networks (DNNs)~\cite{krizhevsky2012imagenet} and then to Transformer~\cite{vaswani2017attention} was creating models capable of effectively utilizing and encoding data into an ever-increasing number of parameters.
   \item \textbf{Data-centric scaling}.  \hspace{1mm} Over the last five years, the focus has shifted to the data side. Self-Supervised Learning (SSL), particularly next-token prediction~\cite{radford2018improving}, first unlocked the ability to use the entire internet as a pretraining corpus. The current frontier is the use of RL in post-training to convert compute into high-quality synthetic data, which can be further divided in two sub-stages:  \hspace{1mm}(1) \textit{RL for LLMs (Internal Reasoning)}: The synthesized data is internal without environment interaction, i.e., the procedural data such as reasoning trajectories that do not exist in the pre-training corpus.
    \hspace{2mm}(2) \textit{RL for LLM Agents (Environmental Interaction)}: 
    This involves interactive RL learning through tool invocation and environmental feedback, which generates interaction data that captures the consequences of the agent's actions.
\end{itemize}

We can see that the ever-growing computational capacity has been systematically converted into more powerful model intelligence by scaling both model parameters and data volume. Focusing on the post-training stage, both forms of RL-driven data synthesis rely on strong pretrained base models but also create new procedural and interaction data that can, in turn, enhance the next round of pretraining. This establishes a positive feedback loop linking pretraining, post-training, and inference, in which RL serves as the critical mechanism continually transforming compute into intelligence. From this perspective, the involved three scaling laws in pretraining, post-training and inference should no longer be viewed as independent, one-way relationships, but as interconnected components of a cyclic system driving the self-improving evolution of agentic intelligence.

\paragraph{\textbf{Extrapolative and Interventional Data Synthesis}}  \hspace{2mm} 
As previously discussed, RL mitigates the shortage of procedural data and alleviates OOD issues by synthesizing new data beyond the natural corpora used in pretraining. Understanding the characteristics of these synthesized data provides an alternative perspective on why RL is critical for acquiring agentic capabilities. As illustrated in the lower half of Fig.~\ref{fig:3}, RL generates two types of synthetic data, extrapolative and interventional data through internal reasoning and environmental interaction, respectively. 

\textit{Extrapolative data} arise from internal cognitive tasks performed by the LLM itself. The model is incentivized to produce procedural data that are not present in original pretraining corpus, with a reward signal then used to select and amplify high-quality samples.
For example, in mathematical reasoning, internet-scale corpora may contain queries $q$ and answers $a$, along with the requisite knowledge (e.g., axioms, theorems), but they rarely contain complete, step-by-step solution trajectories $(q, r_{1:T}, a)$. RL encourages the model to explore within its existing knowledge space via $P_{\theta}(a,r_{1:T}|q)$, combining known concepts to generate previously unseen reasoning paths. When a path leads to a correct answer, it is positively reinforced. This process is essentially a form of structured extrapolation from the model's pre-trained knowledge.

\textit{Interventional data} is generated when an agent is trained to interact with an external environment to complete tasks. Natural data are typically observational (for instance, \textit{(screenshot, click\_position)} pairs from human GUI operation logs), which merely reveal behavioral correlations: what humans tend to do in certain contexts. In contrast, RL allows the agent to actively perform interventions $do(a_t)$ that change the environment from state $s_t$ to $s_{t+1}$, receiving a reward $\rho_t$. Through learning from interventional data $P_{\theta}(s_{t+1},\rho_t|do(a_t), s_t)$ rather than passive observations $P(s_{t+1}|a_t, s_t)$, the agent acquires a causal mapping from action to outcome, thereby learning to predict the consequences of its action. In essence, synthesizing interventional data through RL-driven environment interaction embodies what Rich Sutton describes as ``experience'': \textit{the agent acquires experience by ``doing things,'' and through that experience, it builds its understanding of the world}~\cite{silver2025era}.

\section{Core Capability: Planning}\label{sec:3}
\subsection{Overview}
Planning is formulated as an automated reasoning process that, given an initial state, seeks a feasible sequence or strategy of actions to achieve one or more specified goals, leading the system to a desired state. Traditional symbolic planning methods offer strong interpretability but rely on manually constructed, explicit environment models, leading to high domain-specific customization costs and limited applicability. The advent of LLMs has fundamentally transformed this paradigm. Leveraging implicit world knowledge acquired from vast datasets, LLMs can directly interpret natural language instructions and utilize their inherent commonsense and reasoning capabilities for task decomposition and dynamic adjustment. 

Within this context, early explorations primarily adopted a ``pipeline'' mode. For instance, approaches such as LLM+P~\cite{liu2023llmp} and LLM+PDDL~\cite{10.5555/3666122.3669581} pioneered a hybrid pathway by using the LLM as a front-end for automatic generation of formal planning descriptions (e.g., PDDL~\cite{1998PDDL}), which are then solved by an external planner. Such neuro-symbolic methods integrate the strengths of both paradigms to some extent. Another category of pipeline-based methods, such as CoT~\cite{wei2022chain} and ToT~\cite{yao2023tree}, elicits the model's step-by-step reasoning capability during decoding through prompt engineering to accomplish planning. Although these approaches differ in implementation, they share a common characteristic: the LLM remains within an external collaborative framework, dependent on external planner or prompting strategies to complete the planning process.

\begin{figure}[t] 
  \centering 
  \includegraphics[width=0.98\textwidth]{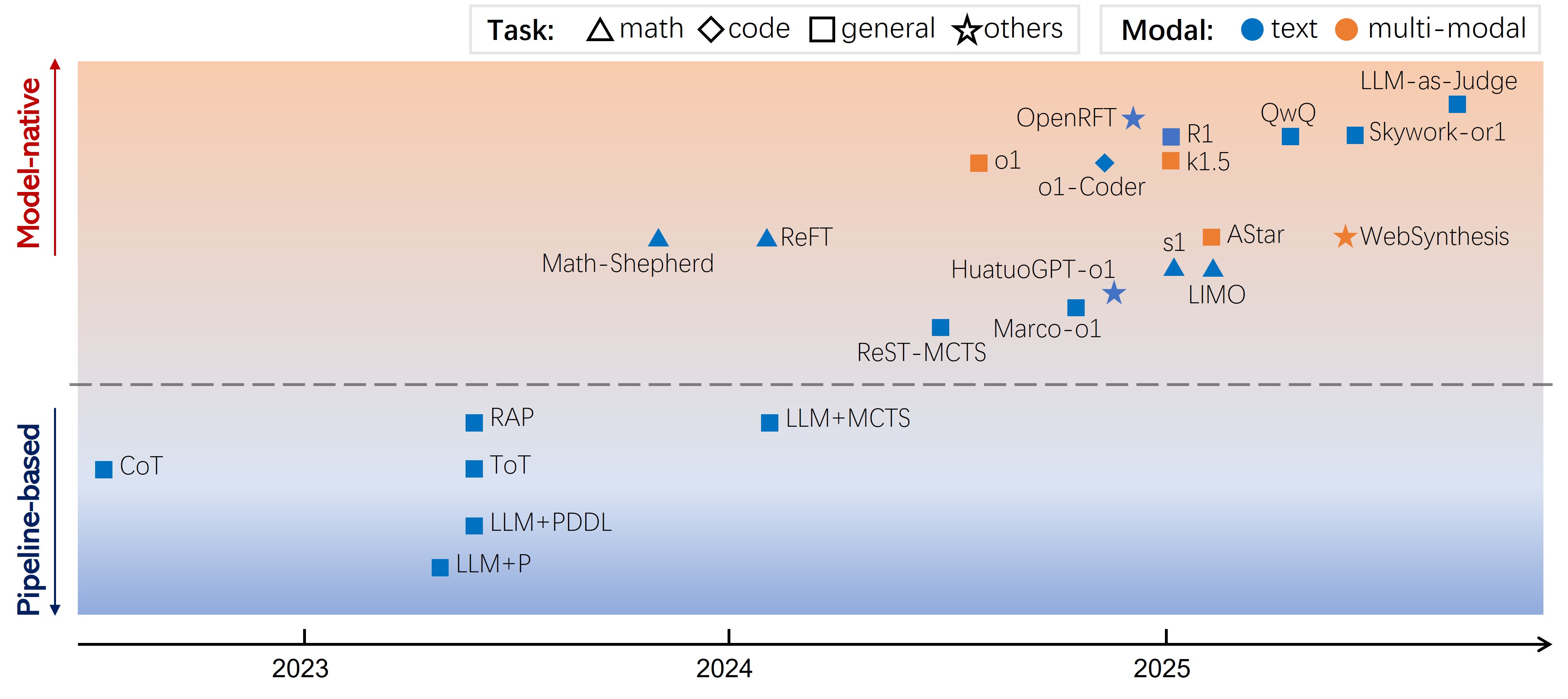} 
  \caption{Overview of agentic planning methods.} 
  \label{fig:plan} 
\end{figure}

However, pipeline-based paradigm fails to fully exploit the LLM's intrinsic potential. As research progresses, the focus of the planning paradigm is gradually shifting from reliance on manual design toward harnessing the model's internal reasoning capabilities. Planning, as a goal-directed decision-making process, fundamentally relies on the model's reasoning abilities in comprehension, decomposition, logic, and causality. Inspired by this, a clear trend has emerged: the internalization of planning capacities into the model. Through techniques such as supervised fine-tuning and reinforcement learning, complex planning and reasoning abilities are directly embedded into the model parameters, enabling more flexible and robust autonomous planning in open-world environments and ultimately eliminating dependence on external frameworks. An overview of agentic planning methods is illustrated in Fig.~\ref{fig:plan}.



\subsection{Pipeline-based Paradigm}\label{sec:3.2}
\subsubsection{\textbf{Symbolic Planning}} \hspace{2mm}
Traditional symbolic planning is characterized by its dependence on explicit, manually constructed action models and logical deduction. Classical planners, exemplified by STRIPS~\cite{FIKES1971189} and PDDL~\cite{1998PDDL}, operate on symbolic state representations and perform state-space search through logical inference based on pre-defined action preconditions and effects, thereby producing a plan to achieve a goal. These methods exhibit robustness and are readily verifiable in closed domains where prior knowledge is sufficient and amenable to precise modeling. However, the translation of flexible natural language tasks into formalized symbolic domains and action schemas necessitates domain expertise, leading to high modeling and maintenance costs and a pronounced sensitivity to modeling errors. 

The emergence of LLMs has partially alleviated the burdens associated with semantic extraction and environment modeling. This has prompted the development of integrated pathways such as LLM+PDDL~\cite{10.5555/3666122.3669581}, which combines LLMs with external planners. In this framework, the LLM automatically induces objects, predicates, and action templates from unstructured text or example trajectories, generating a consistent PDDL domain file for solution by a symbolic planner. Although these hybrid methods demonstrate strong performance and interpretability in closed domains, their capacity for cross-domain generalization and their operational robustness remain constrained in open, dynamic, or noisy environments.

\subsubsection{\textbf{Prompt-based Planning}} \hspace{2mm}
LLM-based planning methods regard planning as a sequence generation task for the model, leveraging the world knowledge and reasoning capabilities acquired during pretraining. This approach significantly surpasses traditional planners relying on rigid symbolic logic in terms of generalization and flexibility. Among them, prompt-based planning, an early and widely explored category, requires no additional training. It operates exclusively by designing appropriate input prompts to guide the model in generating planning paths. Based on the topology of path generation, these methods can be roughly categorized into two types: linear and non-linear. 
\paragraph{\textbf{Linear-structured Planning Methods}}\label{linear} These methods decompose tasks into a sequence of reasoning steps executed in order. A representative approach is Chain-of-Thought (CoT) prompting, which guides the model to produce intermediate reasoning steps. This is achieved by providing few-shot exemplars or zero-shot cues such as “Let’s think step by step,” enabling the model to progressively arrive at the final answer. This approach breaks down complex problems into logically coherent sub-steps, improving the transparency and interpretability of reasoning and helping reduce errors caused by the model omitting steps. However, the linear structure is inherently a local, sequential reasoning mode, lacking a global perspective on the overall task, making it difficult to handle complex planning scenarios that require multi-path exploration or backtracking.

\paragraph{\textbf{Nonlinear-structured Planning Methods}}\label{non-linear} These methods generate multiple possible ``thought nodes'' during reasoning, forming a tree or graph-based search space, and incorporate external or internal evaluation mechanisms for path selection and optimization. For example, Tree-of-Thought (ToT)~\cite{yao2023tree} structures reasoning as a tree, expanding multiple candidate reasoning paths at each step and selecting the optimal one via self-evaluation or external scoring. LLM+MCTS~\cite{yan2024planning} integrates Monte Carlo Tree Search with LLMs, leveraging simulation and backtracking to balance exploration and exploitation. RAP~\cite{hao-etal-2023-reasoning} formalizes reasoning as a planning problem in state space, guided by policy and value functions to generate reasoning paths. Non-linear methods significantly enhance the model’s global planning capability by introducing search mechanisms during reasoning. However, their performance heavily depends on the quality of the external evaluator, and the generation and evaluation of numerous candidate paths lead to a substantial increase in computational cost.

In summary, prompt-based planning methods, characterized by training-free, ease of implementation, and high flexibility, have provided a crucial foundation for the exploration of LLM-based planning. Nevertheless, their limitations are evident: they are highly sensitive to prompt design, exhibit unstable performance in complex tasks, and incur high token consumption costs due to the generation of extensive intermediate content. Consequently, the current research trend is shifting toward internalizing planning capacities into model parameters, a shift that allows models to perform reasoning without relying on explicit search or external evaluators, thereby maintaining strong planning capabilities while significantly improving reasoning efficiency and stability.

\subsection{Model-native Paradigm}\label{sec:3.3}
Based on the source of learning signals, model-native methods can be classified into two  categories: (1) \textit{supervised learning}, which acquires planning capabilities by imitating existing reasoning trajectories; and (2) \textit{reinforcement learning}, which optimizes decision-making policies through self-exploration and learning from rewards.
\subsubsection{\textbf{Supervised Learning}} \hspace{2mm}
Through supervised learning, models acquire planning capabilities by learning from high-quality reasoning process data. This paradigm is fundamentally constrained by its reliance on offline datasets, with final model performance being a direct function of data quality. A critical challenge is the scarcity of such procedural data containing sequential reasoning logic in natural web text, coupled with the high cost and lack of scalability of manual annotation. To address this data bottleneck, contemporary research has pursued two primary directions for constructing training sets: data synthesis and data distillation.
\paragraph{\textbf{Data Synthesis}}\label{synthesis} \hspace{2mm}
Synthetic data denotes reasoning-enriched, long Chain-of-Thought (long-CoT) data generated automatically via programmed rules, LLM-based self-generation, or search mechanisms, eliminating labor-intensive manual annotation. The core challenge lies in systematically generating diverse, challenging, and verifiable reasoning traces, replacing costly human labeling with automated quality control. Currently, the synthesis of high-quality Chain-of-Thought data primarily relies on two approaches: multi-path reasoning process sampling and tree-search methods, both requireing integration with effective quality filtering strategies.
\begin{itemize}
    \item \textbf{Multi-path reasoning-trajectory sampling.} \quad Multiple complete reasoning paths are sampled from the model and then filtered and corrected under pre-defined rules to obtain high-quality data. For example, LIMO~\cite{ye2025limoreasoning} demonstrates that selecting high-quality reasoning trajectories is pivotal for boosting model capability. s1~\cite{muennighoff2025s1simpletesttimescaling} further shows that a small, carefully curated dataset can fine-tune models more effectively than a large, unfiltered one. BOLT~\cite{pang2025boltbootstraplongchainofthought} uses in-context learning to elicit detailed reasoning and filters low-quality responses via outcome-based rewards, thereby automating quality control. In vertical domains, HuatuoGPT-o1~\cite{chen2024huatuogpto1medicalcomplexreasoning} targets verifiable medical problems, sampling correct paths and refining erroneous ones to construct reliable reasoning chains.
    \item \textbf{Tree search.} \quad These typically combine Monte Carlo estimation or process reward models to enable fine-grained control over step-level quality~\cite{li2025fastmctssimplesamplingstrategy,li2025enhancingreasoningprocesssupervision}. For instance, Marco-o1~\cite{zhao2024marcoo1openreasoningmodels} employs Monte Carlo Tree Search (MCTS), using model confidence to guide exploration and synthesize high-quality reasoning data. WebSynthesis~\cite{gao2025websynthesisworldmodelguidedmctsefficient} introduces a world model in Web-UI agent scenarios, combined with MCTS to generate web interaction trajectories at scale, supported by a two-stage curriculum learning strategy to enhance policy capability. ReST-MCTS*~\cite{zhang2024restmctsllmselftrainingprocess} integrates a process reward model with MCTS to collect higher-quality trajectories by assessing the value of each reasoning step. Regarding multimodal reasoning, AStar~\cite{wu2025boostingmultimodalreasoningautomated} defines an executable visual action space and leverages MCTS to obtain high-quality visual reasoning data.
\end{itemize}

To sum up, multi-path sampling focuses on the holistic evaluation and selection of complete reasoning paths, whereas tree search methods enable finer, step-level quality control through structured search and process rewards. Together, these two data synthesis approaches facilitate an effective balance between the quality and scale of synthetic data.

\paragraph{\textbf{Data Distillation}} \hspace{2mm}
Data distillation refers to the process of extracting high-quality reasoning chains and final answers from a strong, System-2 teacher model with advanced reasoning capabilities. After verifying answer correctness and applying simple filtering, this data is used to train a student model. The goal is to transfer the teacher’s reasoning style and step-by-step strategies to a more lightweight student model, thereby internalizing stable planning capabilities. The performance of this approach is highly dependent on the reasoning ability of the teacher model. 
In DeepSeek-R1~\cite{deepseek_r1_2025}, for instance, the reasoning capability of the base model was first improved through direct RL, and then followed by data distillation. Only simple verification and filtering were required to obtain high-quality reasoning process data. After the release of DeepSeek-R1, thanks to its powerful reasoning performance, many studies have attempted to use DeepSeek-R1 as the teacher model to distill high-quality reasoning data~\cite{guha2025openthoughtsdatarecipesreasoning,openr1,2025synthetic1,bespoke_stratos,slam-distillation-from-r1}. A common characteristic of these works is the achievement of scalable and automated data production, significantly reducing the cost of manual annotation.

\subsubsection{\textbf{Reinforcement Learning}} \hspace{2mm}
In the training of reasoning models, supervised learning remains effective but is constrained by its heavy reliance on large volumes of high-quality reasoning data. Moreover, the learned patterns often exhibit limited generalization~\cite{chu2025sftmemorizesrlgeneralizes,Wu2025OnTG}, making them less adaptable to complex and dynamic scenarios. As foundation models continue to improve, they increasingly possess strong inherent reasoning capabilities, which means the key challenge lies in how to effectively elicit and activate these abilities. This has made RL-based optimization a viable and promising path. Early academic explorations primarily focused on process reward, aiming to shape the model's reasoning trajectory through fine-grained supervision. However, the release of DeepSeek-R1~\cite{deepseek_r1_2025} has profoundly revealed the practical challenges associated with process reward, while successfully demonstrating the potential of using mere outcome reward. Accordingly, our discussion of reinforcement learning will also center around these two types of rewards.
\paragraph{\textbf{Process Reward}} \hspace{2mm}
Process reward refers to the mechanism of evaluating the correctness of each intermediate step in a model's multi-step reasoning process and providing corresponding reward signals. The core idea is to ensure the reliability of the entire reasoning path through dense reward guidance. By providing fine-grained supervision over intermediate steps, Process Reward Model (PRM) is expected to not only identify and correct errors more effectively but also enhance the model's performance on unseen problems.
However, its effectiveness hinges on one critical prerequisite: the availability of large-scale, high-quality step-level supervision data. This reintroduces the  bottleneck of supervised learning, i.e., the need for extensive, granular annotation, which represents the central challenge in PRM development. Based on the source and method of supervision signals, we categorize process reward into explicit process reward and implicit process reward. 
\begin{itemize}
    \item \textbf{Explicit process reward.} \setlength{\parindent}{1em} Explicit process reward relies on high-quality human annotations or reliable automatic labeling methods to obtain precise step-level reward signals. For instance, ProcessBench~\cite{zheng2025processbenchidentifyingprocesserrors} constructs specialized step-level benchmarks through expert human annotation, providing reliable metrics for model performance. 
    
    \indent To address scalability limitations of manual labeling, many studies employ Monte Carlo (MC) estimation or LLM-as-Judge~\cite{li2025whosjudgedetectabilityllmgenerated} methods, sampling and comparing multiple reasoning paths to assess step quality. For example, Math-Shepherd~\cite{wang2024mathshepherdverifyreinforcellms} assigns correctness scores to each mathimatical inference step via MC estimation, eliminating dependency on human annotation. OmegaPRM~\cite{luo2024improvemathematicalreasoninglanguage} enhances this approach by combining MC estimation with Monte Carlo Tree Search (MCTS), using binary search to locate the first erroneous step in reasoning chains, thereby improving data collection efficiency. Qwen-2.5-Math-PRM~\cite{zhang2025lessonsdevelopingprocessreward} notes that both MC estimation and LLM-as-Judge introduce noise in automatic labeling, while human annotation remains cost-prohibitive, thus proposing a hybrid approach to reduce noise and improve accuracy. 
    
   \indent  In executable domains like code generation, execution results can serve as process supervision signals: PRLCoder~\cite{ye2025processsupervisedreinforcementlearningcode} determines line correctness through compilation and test execution, enabling precise line-by-line labeling. ORPS~\cite{Yu2024ReasoningTE} integrates execution feedback directly into the reasoning loop, with each step evaluated based on program execution results. The advantage of execution feedback lies in its absolute objectivity (pass/fail test criteria), making this method particularly suitable for tasks with clear verification standards like code generation and theorem proving.
   
    \item \textbf{Implicit process reward.} \setlength{\parindent}{1em} Instead of relying on explicit process-level annotations, implicit process reward methods derive stepwise supervision directly from the model’s internal dynamics. The key idea is to approximate process rewards by using the language model’s own likelihood changes as a proxy signal, i.e., measuring how each intermediate reasoning step increases or decreases the final outcome probability. This allows for the efficient, on-the-fly generation of dense rewards without requiring any process-level labels.
    
    \indent The foundational work on Implicit PRM~\cite{pmlr-v267-yuan25c} first demonstrated this principle. It showed that an Outcome Reward Model (ORM), trained only on final outcome labels, could effectively generate process-level rewards during inference by calculating log-likelihood ratios. Building on this, PRIME~\cite{cui2025processreinforcementimplicitrewards} integrated this mechanism into a full online reinforcement learning loop. Instead of just using the implicit rewards for inference-time evaluation, PRIME utilizes these implicitly generated, per-token dense rewards to directly compute advantage estimates for policy updates during training. This approach avoids the need to train a separate, explicit PRM, significantly reducing development costs and complexity, and has shown notable improvements in performance on mathematical and code reasoning tasks.    
\end{itemize}

Overall, PRMs offer a viable way for enhancing the controllability and generalization of reasoning models by providing fine-grained, step-level supervision, thereby effectively mitigating the challenge of reward sparsity in multi-step reasoning. However, as highlighted by DeepSeek-R1~\cite{deepseek_r1_2025}, several critical challenges remain. First, the correctness of intermediate steps is often difficult to define and evaluate unambiguously, as step-quality criteria tend to be subjective and task-dependent. Second, obtaining high-quality process supervision data is challenging: manual annotation is prohibitively expensive, while automated labeling methods are generally prone to noise. Furthermore, PRMs are susceptible to reward hacking, where models may learn to exploit loopholes in the reward function rather than genuinely improving reasoning quality. These challenges collectively constrain the further development and application of PRMs, underscoring the need for future breakthroughs in evaluation frameworks, data construction, and training mechanisms.

\paragraph{\textbf{Outcome Reward}} \hspace{2mm}
Outcome reward, also referred to as outcome supervision, provides reward signals based solely on the correctness of the final answer. It offers a sparse yet unambiguous reward signal, grounded in the core hypothesis that a powerful foundation model can, through feedback on end results, autonomously explore and internalize effective reasoning policy. Exemplified by the RLVF approach adopted in DeepSeek R1~\cite{deepseek_r1_2025}, outcome reward completely bypasses the complex evaluation of intermediate steps and only requires verifying the correctness of the final answer~\cite{qwq32b,skywork-or1-2025}. In tasks with easily verifiable answers, such as multiple-choice questions and mathematical problems, RL is typically applied using verifiable outcome rewards to enhance the model's reasoning capability~\cite{trung-etal-2024-reft,zhang2024openrftadaptingreasoningfoundation,deepscaler2025}. Similarly, in the domain of code generation, methods like o1-Coder~\cite{zhang2024o1codero1replicationcoding} and RLEF~\cite{gehring2025rlefgroundingcodellms} validate answer correctness based on program execution feedback. 

The primary advantages of outcome reward are its low cost and high reliability. The annotation cost is minimal because only the final answer needs verification, making the approach highly scalable. Furthermore, the reward signal is accurate and verifiable, as the correctness of a final answer usually meets clear, objective standards, thereby avoiding the subjectivity and inconsistency issues that challenge process reward. As a result, outcome reward has become one of the mainstream approaches for strengthening reasoning abilities in automatically verifiable tasks such as mathematics and code generation.

Beyond learned outcome models, a parallel line of work designs explicit \textit{rule-based rewards} that score outputs or intermediate reasoning steps according to deterministic rules. These rules may enforce output formats, verify structural consistency, or constrain reasoning-chain length to mitigate redundant computation and balance accuracy with efficiency. Unlike learned reward models, this approach uses explicit rules to constrain and shape reasoning behaviors, offering a simple and scalable method for supervision. Widely-used rules include: (1) format reward~\cite{deepseek_r1_2025}, which enforces a specific output structure to ensure the stable extraction and verification of reasoning processes and final answers, and (2) length-based reward~\cite{aggarwal2025l1controllinglongreasoning,luo2025o1prunerlengthharmonizingfinetuningo1like,shen2025dastdifficultyadaptiveslowthinkinglarge,yeo2025demystifyinglongchainofthoughtreasoning}, which imposes constraints on the length of reasoning chains to control for redundant computation (``overthinking'') or to achieve a controllable trade-off between accuracy and computational cost. Similar to outcome rewards, rule-based rewards are characterized by their simplicity in design and ease of scalability. Their key function, however, is to provide auxiliary structural supervision that guides reasoning beyond the final answer's correctness. Therefore, they are often combined with outcome rewards to create a more holistic objective function that jointly guides the optimization of the model's reasoning behavior.

\newcolumntype{C}[1]{>{\centering\arraybackslash}p{#1}}
\afterpage{
\clearpage 
\footnotesize

\begin{longtable}{C{1.8cm}|l|c|c|c|c|c} 
\caption{Overview of agentic planning methods.}\label{tab:planning}\\ 
\toprule
\rowcolor{blue!5}
\textbf{} & \textbf{Method} & \textbf{Task} & \textbf{Modal} & \textbf{Affiliation} & \textbf{Access} & \textbf{Date} \\
\midrule
\endfirsthead

\multicolumn{7}{c}{\tablename~\thetable\ -- \textit{Overview of agentic tool use methods. }}\\
\toprule
\textbf{Domain} & \textbf{Date} & \textbf{Name} & \textbf{\#Sample} & \textbf{Format} & \textbf{Type} & \textbf{Link} \\
\midrule
\endhead

\midrule
\multicolumn{7}{r}{\textit{continued on next page}}\\
\endfoot

\bottomrule
\endlastfoot

\rowcolor{black!5}\multicolumn{7}{c}{\textit{\textbf{Pipeline-based Paradigm}}}\\
\midrule

\multirow{4}{*}{\makecell[c]{\textit{\textbf{Symbolic}}\\ \textit{\textbf{Planning}}}}
& STRIPS~\cite{FIKES1971189}& General& Text& Academia& No& 1971 \\
& PDDL~\cite{1998PDDL}& General & Text & Academia& No& 1998 \\
& LLM+P~\cite{liu2023llmp} & General & Text& Academia& \href{https://github.com/Cranial-XIX/llm-pddl}{Yes}& 23.04 \\
& LLM+PDDL~\cite{10.5555/3666122.3669581} & General & Text& Academia& \href{https://guansuns.github.io/pages/llm-dm/}{Yes}& 23.05 \\
\midrule

\multirow{4}{*}{\makecell[c]{\textit{\textbf{Prompt-based}}\\ \textit{\textbf{Planning}}}} 
& CoT~\cite{wei2022chain} & General& Text& Academia& No& 22.01 \\
& ToT~\cite{yao2023tree} & General & Text & Academia& \href{https://github.com/princeton-nlp/tree-of-thought-llm}{Yes}& 23.05 \\
& RAP~\cite{hao-etal-2023-reasoning} & General & Text& Academia& No & 23.05 \\
& LLM+MCTS~\cite{yan2024planning} & General & Text& Academia& No & 24.09 \\
\midrule

\rowcolor{black!5}\multicolumn{7}{c}{\textit{\textbf{Model-native Paradigm}}}\\
\midrule
\multirow{14}{*}{\makecell[c]{\textit{\textbf{Supervised}}\\\textit{\textbf{Learning}}}}
& ReST\text{-}MCTS*~\cite{zhang2024restmctsllmselftrainingprocess} & General & Text      & Academia & \href{https://rest-mcts.github.io/}{Yes} & 24.06 \\
& Marco\text{-}o1~\cite{zhao2024marcoo1openreasoningmodels} & General & Text        & Industry & \href{https://github.com/AIDC-AI/Marco-o1}{Yes} & 24.11 \\
& HuatuoGPT-o1~\cite{chen2024huatuogpto1medicalcomplexreasoning}    & Others  & Text        & Academia & \href{https://github.com/FreedomIntelligence/HuatuoGPT-o1}{Yes} & 24.12 \\
& Bespoke\text{-}Stratos~\cite{bespoke_stratos} & General & Text  & Industry & \href{https://huggingface.co/datasets/bespokelabs/Bespoke-Stratos-17k}{Yes}  & 25.01 \\
& s1~\cite{muennighoff2025s1simpletesttimescaling}              & Math    & Text        & Academia & \href{https://github.com/SimpleScaling/s1}{Yes} & 25.01 \\
& R1\text{-}Distill\text{-}SFT~\cite{slam-distillation-from-r1} & General & Text      & Industry & \href{https://huggingface.co/datasets/ServiceNow-AI/R1-Distill-SFT}{Yes} & 25.01 \\
& LIMO~\cite{ye2025limoreasoning}            & Math    & Text        & Academia & \href{https://github.com/GAIR-NLP/LIMO}{Yes} & 25.02 \\
& BOLT~\cite{pang2025boltbootstraplongchainofthought}            & General & Text        & Industry & No  & 25.02 \\
& AStar~\cite{wu2025boostingmultimodalreasoningautomated}           & General & Multi-modal & Academia & No  & 25.02 \\
& FastMCTS~\cite{li2025fastmctssimplesamplingstrategy}        & General & Text        & Academia & No  & 25.02 \\
& OpenThoughts~\cite{guha2025openthoughtsdatarecipesreasoning}    & General & Text        & Academia & \href{https://www.openthoughts.ai/}{Yes} & 25.02 \\
& OpenR1\text{-}Math\text{-}220k~\cite{openr1} & Math & Text      & Industry & \href{https://github.com/huggingface/open-r1/tree/main}{Yes} & 25.02 \\
& SYNTHETIC\text{-}1~\cite{2025synthetic1} & General & Text     & Industry & \href{https://huggingface.co/datasets/PrimeIntellect/SYNTHETIC-1-SFT-Data}{Yes} & 25.02 \\
& WebSynthesis~\cite{gao2025websynthesisworldmodelguidedmctsefficient}    & Others  & Multi-modal & Academia & \href{https://github.com/LucusFigoGao/WebSynthesis}{Yes} & 25.07 \\
\midrule

\multirow{22}{*}{\makecell[c]{\textit{\textbf{Reinforcement}}\\ \textit{\textbf{Learning}}}}
& Math-Shepherd~\cite{wang2024mathshepherdverifyreinforcellms} & Math & Text & Academia & \href{https://achieved-bellflower-4d6.notion.site/Math-Shepherd-Verify-and-Reinforce-LLMs-Step-by-step-without-Human-Annotations-41b6e73c860840e08697d347f8889bac}{Yes} & 23.12 \\
& ReFT~\cite{trung-etal-2024-reft} & Math & Text & Industry & \href{https://github.com/lqtrung1998/mwp_ReFT}{Yes} & 24.01 \\
& OmegaPRM~\cite{luo2024improvemathematicalreasoninglanguage} & Math & Text & Academia & No & 24.06 \\
& OpenAI o1~\cite{openai_o1_2024} & General & Multi-modal & Industry & No & 24.09 \\
& RLEF~\cite{gehring2025rlefgroundingcodellms} & Code & Text & Industry & No & 24.10 \\
& o1-Coder~\cite{zhang2024o1codero1replicationcoding} & Code & Text & Academia & \href{https://github.com/ADaM-BJTU/O1-CODER/tree/main}{Yes} & 24.11 \\
& Implicit PRM~\cite{pmlr-v267-yuan25c} & Math & Text & Academia & No & 24.12 \\
& ORPS~\cite{Yu2024ReasoningTE} & Code & Text & Academia & \href{https://github.com/zhuohaoyu/ORPS}{Yes} & 24.12 \\
& OpenRFT~\cite{zhang2024openrftadaptingreasoningfoundation} & Others & Text & Academia & \href{https://github.com/ADaM-BJTU/OpenRFT}{Yes} & 24.12 \\
& DeepSeek R1~\cite{deepseek_r1_2025} & General & Text & Industry & \href{https://github.com/deepseek-ai/DeepSeek-R1}{Yes} & 25.01 \\
& Qwen-2.5-Math-PRM~\cite{zhang2025lessonsdevelopingprocessreward} & Math & Text & Industry & \href{https://huggingface.co/Qwen/Qwen2.5-Math-PRM-72B }{Yes} & 25.01 \\
& Kimi k1.5~\cite{kimiteam2025kimik15scalingreinforcement} & General & Multi-modal & Industry & No & 25.01 \\
& O1-Pruner~\cite{luo2025o1prunerlengthharmonizingfinetuningo1like} & General & Text & Academia & \href{https://github.com/StarDewXXX/O1-Pruner}{Yes} & 25.01 \\
& PRIME~\cite{cui2025processreinforcementimplicitrewards} & General & Text & Academia & No & 25.02 \\
& DeepScaleR~\cite{deepscaler2025} & Math & Text & Academia & \href{https://github.com/rllm-org/rllm}{Yes} & 25.02 \\
& PRLCoder~\cite{ye2025processsupervisedreinforcementlearningcode} & Code & Text & Academia & No & 25.02 \\
& L1~\cite{aggarwal2025l1controllinglongreasoning} & General & Text & Academia & \href{https://cmu-l3.github.io/l1/}{Yes} & 25.03 \\
& DAST~\cite{shen2025dastdifficultyadaptiveslowthinkinglarge} & General & Text & Industry & \href{https://github.com/AnonymousUser0520/AnonymousRepo01}{Yes} & 25.03 \\
& QwQ~\cite{qwq32b} & General & Text & Industry & \href{https://huggingface.co/Qwen/QwQ-32B}{Yes} & 25.03 \\
& Skywork or1~\cite{skywork-or1-2025} & General & Text & Industry & \href{https://github.com/SkyworkAI/Skywork-OR1}{Yes} & 25.05 \\
& Demystify-long-cot~\cite{yeo2025demystifyinglongchainofthoughtreasoning} & General & Text & Academia & \href{https://github.com/eddycmu/demystify-long-cot}{Yes} & 25.05 \\
& LLM-as-Judge~\cite{li2025whosjudgedetectabilityllmgenerated} & Others & Text & Academia & No & 25.09 \\


\end{longtable}
}

\subsection{Summary and Discussion} \label{sec:3.4}
This section outlines the shift in LLM planning from the external framework-dependent ``pipeline'' paradigm toward the autonomous capability-oriented ``model-native'' paradigm. The core distinction lies in where planning capability resides: the former positions the LLM as a front-end or collaborator, whose planning efficacy depends on integration with external symbolic planners or complex prompt engineering---yet remains constrained by the high costs of knowledge formalization and prompt design; the latter aims to directly encode planning and reasoning abilities into the model's parameters, transforming the model into an independent, end-to-end planning agent. The reviewed representative studies are summarized in Table~\ref{tab:planning}. 

Within the evolution of the model-native paradigm, we observe two major shifts.
First, there is a transition from SFT to RL. While SFT serves as the foundation for ability internalization, its effectiveness is limited by the scarcity and high cost of high-quality annotated data. As base models become more capable, RL emerges as a more effective post-training paradigm, alleviating the dependency on costly process-level supervision.
Second, within RL itself, there is a shift from process reward to outcome reward. Although process reward offers dense, step-by-step guidance, it faces serious challenges such as ambiguous definitions of intermediate step correctness, high labeling costs, and reward hacking. Exemplified by DeepSeek-R1~\cite{deepseek_r1_2025}, research has established that sparse yet objective outcome rewards are sufficient to guide capable models toward effective reasoning, making this approach a mainstream choice for verifiable domains like mathematics and code. It is often combined with rule-based rewards (e.g., for formatting or length constraints) to achieve more stable optimization.

It is worth noting that this evolution from ``pipeline'' to ``model-native'' is not confined to linguistic planning. A similar trajectory is observed in multi-modal reasoning: early multi-modal tasks often relied on external visual tools or complex prompting chains to connect vision and language modules, whereas current research aims to enable models to inherently process and reason over multi-modal information through end-to-end training, achieving genuine "what you see is what you think" reasoning~\cite{lin2025mindeyeslanguagereasoning}.

Looking forward, research on model-native planning capabilities will evolve along the following directions:
\begin{itemize}
    \item \textbf{From explicit to implicit reasoning}: Current explicit reasoning processes relying on Chain-of-Thought enhance interpretability but incur significant computational and storage overhead. A key future direction is achieving an implicit CoT, where the planning process occurs within the model's hidden activations rather than generating lengthy textual steps~\cite{wei2025simcotsupervisedimplicitchainofthought}. This could substantially improve computational efficiency and better approximate the intuitive, rapid decision-making patterns of human experts.
    \item \textbf{From supervised to self-supervised/unsupervised internalization}: To reduce reliance on external reward signals, future research will explore self-supervised or unsupervised learning mechanisms. For example, by designing intrinsic reward signals such as exploration rewards based on information-theoretic principles like information gain or entropy~\cite{zhang2025rightquestionhalfanswer}, models could be motivated to autonomously discover novel and complex reasoning strategies, enabling continuous learning without explicit external objectives.
    \item \textbf{From single-task to general and transferable planning}: Current models are often trained for planning within specific domains (e.g., mathematics or code). A significant future challenge lies in achieving cross-domain generalization and transfer of planning abilities~\cite{su2025crossingrewardbridgeexpanding,liu2025xreasonergeneralizablereasoningmodalities}. This requires models not only to learn problem-solving schemas for specific tasks but also to abstract general meta-cognitive planning abilities, such as task decomposition, subgoal management, and dynamic strategy adjustment, enabling improved generalization or rapid adaptation to new tasks.
\end{itemize}

\section{Core Capability: Tool Use}\label{sec:4}
\subsection{Overview}
While the previous section addressed the agent's high-level task planning capability---the strategic reasoning process of decomposing a complex goal into a sequence of logical sub-goals, this section focuses on a distinct, more tactical layer of planning that is inherent to tool use. Tool use refers to the decision-making process in which an intelligent agent invokes external tools to extend its capability and accomplish complex tasks. The challenge of tool use is not solved by high-level planning alone: each sub-goal must be executed. Therefore, for an agent to act upon the world, it must also be able to plan the concrete sequence of actions required to achieve these sub-goals.

To systematically analyze this capability, we deconstruct tool use into two layers: \textit{planning} and \textit{execution}. It is critical to distinguish the planning discussed here from the task planning in the previous section. Here, planning refers to the action-level orchestration of tool invocations: determining the timing and sequence of tool calls and refining this action-plan based on feedback. The execution layer, in contrast, is responsible for generating the specific, syntactically correct invocation commands and interacting with the environment. 

Early explorations into tool use were notably marked by the introduction of single-turn functional calls~\cite{openai2023functioncalling}, which largely focused on the execution layer. The model's task was confined to generating a single, structured API request in response to a direct user query. However, for a truly agentic system, tool use involves more sophisticated decision-making process with multi-turn interaction and dynamic adaption, i.e., the model must master both the strategic planning of when to use tools and the subsequent execution of those calls. This section will examine the paradigm shift in how both the planning and execution layers of tool use are implemented, from external pipeline-based systems to internalized, model-native policies.
 
Early research methods primarily followed the pipeline-based paradigm, characterized by the externalization of tool-use decision logic. Specifically, one category of approaches relies on hard-coded system workflows, embedding the model into predefined execution nodes. Another category, carefully designing prompts, grants the model a degree of autonomy by embedding the decision logic for tool use, covering invocation timing, tool selection, and parameter specification, within structured prompts to guide dynamic planning and execution during reasoning. 

\begin{figure}[t] 
  \centering 
  \includegraphics[width=0.98\textwidth]{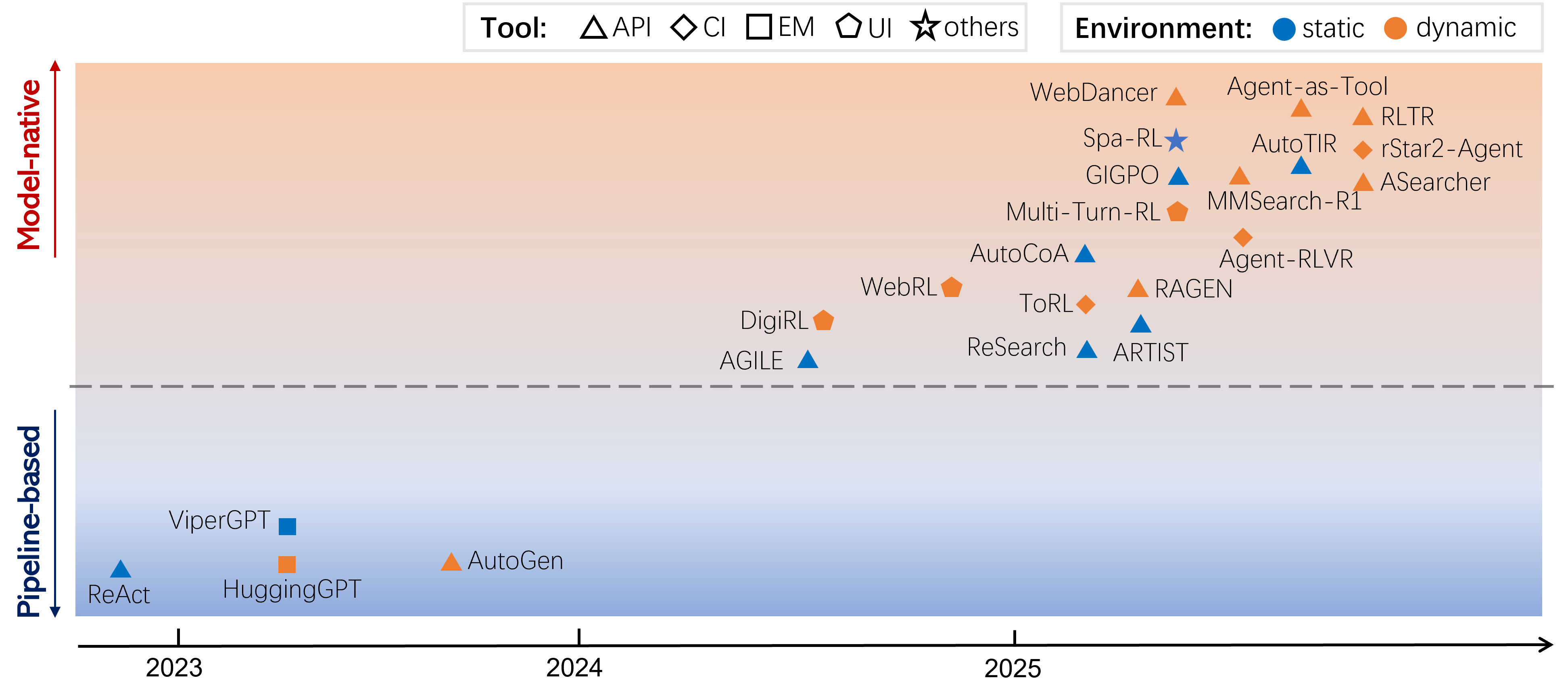} 
  \caption{Overview of agentic tool use methods.} 
  \label{fig:tool_use} 
\end{figure}

Recently, the research on agentic tool use has shifted toward the model-native paradigm, which internalizes tool-use decision-making capabilities within the model's parameters. This approach emphasizes the model's ability to autonomously perform tool selection and invocation during reasoning. Along the two layers of planning and execution, existing studies can be divided into two categories: (1) \textit{Modular training} decouples planning and execution, typically optimizing only the planner while delegating execution to external modules; (2) \textit{End-to-end training}, in contrast, emphasizes the joint optimization of planning and execution, imposing supervision signals on the model's multi-step planning and execution to cultivate a unified policy handling the complete process from decision-making to invocation. An overview of agentic tool use methods is illustrated in Fig.~\ref{fig:tool_use}.

\subsection{Pipeline-based Paradigm}\label{sec:4.2}

\subsubsection{\textbf{System-based Workflow}} \hspace{2mm}
System-based workflow methods follow a predefined sequence. The model is embedded at specific nodes within this flow, acting as a  subtask executor. Its responsibility is to generate specified outputs based on given inputs, without the authority to autonomously plan the overall task path or decide on tool usage. 
In the early implementations, the dialogue system BlenderBot 2.0 ~\cite{xu-etal-2022-beyond,komeili-etal-2022-internet} externalized long-term memory and web search modules as controllable components, enabling the dialogue engine to explicitly trigger queries and generate responses based on the retrieved results. HuggingGPT~\cite{Shen2023HuggingGPT} employs a pipeline of task planning, model selection, execution, and response generation, utilizing ChatGPT as a coordinator to select and orchestrate multi-modal expert models. Code as Policies ~\cite{Liang2023CodeAsPolicies} treats LLM-generated executable code as a policy interface, enabling robots to implement coded skills and controllers.

In summary, for system-based workflows, the advantages include high predictability due to the fixed process, stable execution, less prone to errors, and ease of debugging and replication. However, this hard-coded design also leads to insufficient flexibility, difficulty in handling exceptions outside the predefined flow or complex tasks, poor generalization capability, and potential need for extensive system restructuring to add new tools or functions, limiting the rapid iteration and expansion of agent capabilities.

\subsubsection{\textbf{Prompt-based Methods}} \hspace{2mm}
Prompt-based methods delegate the decision-making authority for tool use to LLM itself, where the overall control flow is dynamically generated by the model's reasoning chain. Based on the relationship between planning and execution, we categorize existing prompt-based methods into: (1) \textit{Plan-and-Execute Separation}: exemplified by Plan-and-Execute, where a planner generates a high-level plan, and an executor carries it out step-by-step. (2) \textit{Interleaved Planning and Execution}: exemplified by ReAct ~\cite{yao2023react}, where ``Thought'' and ``Action'' are alternately generated within a single reasoning sequence, and subsequent decisions are updated based on immediate ``Observation'',  forming a closed-loop control adaptive to environmental feedback.

\paragraph{\textbf{Plan-and-Execute Separation}} \hspace{2mm}
The model first formulates a complete action plan, subsequently followed by step-by-step execution by the system or the model itself. This mode offers clear logic but struggles to adapt to unexpected deviations during execution.
For instance, the PAL ~\cite{pmlr-v202-gao23f} method requires the model to write Python subroutines, which are then executed by an interpreter at runtime to obtain results. PoT ~\cite{chen2023program} further expresses the entire reasoning chain as program code, effectively decoupling the reasoning process from numerical computation. For multimodal and knowledge-intensive tasks, Chameleon ~\cite{lu2023chameleon} explicitly separates planning and execution: an LLM acts as the planner, composing tool sequences (e.g., retrieval, Python functions, vision models, heuristic modules) as needed, which are then executed uniformly to produce the answer.

\paragraph{\textbf{Interleaving Planning and Execution}} \hspace{2mm}
The model performs immediate reasoning at each step to decide the next action, observes the result after execution, and continues reasoning accordingly, forming a dynamic decision-making process capable of adapting to environmental feedback.
The ReAct framework ~\cite{yao2023react} structures each interaction round into ``Thought-Act-Observation'', forming a multi-turn reasoning-action loop. This facilitates simultaneous evidence gathering and plan adjustment during reasoning, significantly enhancing the model's adaptability in dynamic environments. It has demonstrated significant advantages in task success rate, reasoning transparency, and result reproducibility for knowledge-intensive QA and interactive environments. For example, Self-Ask ~\cite{press-etal-2023-measuring} explicitly deconstructs complex questions into ``selfask-retrieve-synthesize'' sub-stages, proving particularly adept at handling compositional reasoning tasks requiring multi-step information integration.

\vspace{2mm} In summary, prompt-based methods represent a paradigm shift in tool-augmented language models, transitioning system control from rigid architectures to dynamic LLM-driven reasoning. While plan-and-execute separation provides structural clarity, interleaved planning and execution offers superior adaptability through real-time feedback integration. The choice between these approaches depends on task requirements: the former suits well-defined problems with predictable execution paths, whereas the latter excels in dynamic environments requiring iterative refinement.

\subsection{Model-native Paradigm}\label{sec:4.3}
\subsubsection{\textbf{Modular Training}} \hspace{2mm}
Modular training decouples planning from execution: a compact, trainable planner selects actions and tool calls, while a frozen executor formats API invocations and interacts with the environment. This separation concentrates learning on decision making and mitigates credit-assignment issues introduced by execution noise. For example, Agent-as-Tool~\cite{zhang2025agentastoolstudyhierarchicaldecision} argues that end-to-end optimization dilutes the reward signal and complicates credit assignment. It therefore factorizes the agent into a trainable planner for high-level reasoning and a separate toolcaller for execution. RL is applied exclusively to the planner, with environmental feedback masked during training to isolate the learning signal to the decision-making policy. AI-SearchPlanner~\cite{mei2025aisearchplannermodularagenticsearch} assigns retrieval and call sequencing to a small trainable planner while delegating QA to a parameter-frozen large model (e.g., GPT-4, DeepSeek-R1), thereby avoiding simultaneous optimization of heterogeneous capabilities in a single loop. RLTR~\cite{li2025encouraginggoodprocessesneed} reaches a similar conclusion: since end-to-end training struggles to focus updates on planning, it optimizes only the planner, converting a multi-objective problem into a single-objective one. 

Overall, the modular training paradigm treats planning as the core of agent decision-making. In end-to-end optimization, structured parameter generation and interaction with the environment introduce additional noise, thereby weakening the effective propagation of learning signals. By contrast, the execution layer has a single, verifiable objective that can typically be realized robustly via templates, rules, or frozen models. On this basis, removing the execution layer from the training loop and modularizing it while concentrating on optimizing the planner not only conserves the training budget for the primary task, avoiding expenditure on format learning and noise handling, but also substantially improves sample efficiency and training stability.

\subsubsection{\textbf{End-to-end Training}} \hspace{2mm}
End-to-end training refers to coupling planning and execution under a single objective, such that one model jointly learns the abilities for multi-step planning and action execution. While modular training offers greater control by isolating individual components, the shift toward a fully end-to-end paradigm introduces two principal bottlenecks that the unified policy must now learn to handle directly: (1) cross-step credit assignment, which centers on the granularity of attribution---whether it is assigned at the trajectory level or the turn level; and (2) uncertainty in environmental feedback, which is largely determined by the type of environment---static or simulated environments typically being more controllable and less noisy, whereas dynamic or real-world environments exhibit greater stochasticity and noisier feedback. Accordingly, the discussion of end-to-end training methods is organized along two axes: the granularity of credit assignment and the type of the environment.
\paragraph{\textbf{Credit-assignment Granularity}}\label{axis:granularity} \hspace{2mm}
Credit assignment refers to attributing a task’s final return or verifiable outcome to each tool-related decision (and its preceding reasoning) along the action sequence, thereby producing learnable signals for policy optimization. The design of the credit-assignment scheme directly determines both the informativeness of these signals and the stability of training. Along this dimension, existing approaches can be organized into two categories.
\begin{itemize}
    \item \textbf{Trajectory-level credit assignment.} \setlength{\parindent}{1em} Rewards are computed solely from the final, verifiable outcome and applied uniformly across the entire trajectory, implicitly treating all intermediate steps as equal contribution. Representative works include Search-R1~\cite{jin2025searchr1trainingllmsreason}, R1-Searcher ~\cite{song2025r1}, and ReSearch~\cite{chen2025researchlearningreasonsearch}, which optimize multi-turn retrieval with simple outcome-based rewards and use retrieval-token masking to limit overfit. AutoCoA~\cite{zhang2025agentmodelsinternalizingchainofaction} internalizes a Chain-of-Action agent: while its SFT stage contains step-level action triggers, its RL phase adopts trajectory-level credit assignment. ARTIST~\cite{singh2025agenticreasoningtoolintegration} unifies agentic reasoning, tool integration, and RL in a single framework where the model autonomously decides when, how, and which tools to invoke; optimization relies on outcome-based RL without step-level supervision. 
    
    \indent In multimodal settings, trajectory-level supervision is likewise common: VTool-R1~\cite{wu2025vtoolr1vlmslearnthink} alternates textual reasoning with intermediate visual operations, integrates a Python image-editing tool into the RL loop, and trains with accuracy-linked outcome rewards; DeepEyes~\cite{zheng2025deepeyesincentivizingthinkingimages} explores “thinking with images” via a unified multimodal agent and emphasizes end-to-end, outcome-driven RL without cold-start SFT. Overall, trajectory-level methods are simple, scalable, and practical for large, real-world tasks. However, in long-horizon, multi-tool, multi-turn scenarios, reward sparsity and diffuse credit can weaken policy-improvement signals.
    
    \item \textbf{Step-level credit assignment.} The goal is to identify which intermediate steps or turns substantially contribute to the final outcome. In multi-turn agent RL, finer-grained credit enables targeted improvement of specific actions rather than averaging rewards across all steps. For example, RAGEN~\cite{wang2025ragenunderstandingselfevolutionllm} shows that robust reasoning fails to emerge in multi-turn agent RL without fine-grained rewards, underscoring the need for granular attribution. Several lines of work pursue finer credit in multi-turn RL: Multi-Turn-RL~\cite{zeng2025reinforcingmultiturnreasoningllm} estimates advantages directly at the turn level; SPA-RL~\cite{wang2025sparlreinforcingllmagents} decomposes the terminal reward into per-step progress; GiGPO~\cite{feng2025groupingrouppolicyoptimizationllm} combines trajectory- and step-level signals without adding a critic or extra sampling, significantly improving multi-turn training; StepSearch~\cite{wang2025stepsearchignitingllmssearch} frames retrieval and tool use through policy updates guided by information gain at each step, to mitigate policy drift from sparse global signals. Collectively, these methods shift reward from the terminal outcome to individual step, aligning gradients with effective actions and stabilizing training.
\end{itemize}

\paragraph{\textbf{Environment Type}}\label{axis:env}  \hspace{2mm}
Regarding tool use, feedback arising from agent-environment interactions is incorporated into the context and conditions subsequent reasoning. Because this feedback is not generated autonomously by the model and is affected by environmental noise and temporal non-stationarity, it often amplifies uncertainty in subsequent outputs. The magnitude of this uncertainty is largely determined by environmental characteristics: it tends to be lower in static or well-controlled simulated settings and substantially higher in dynamic or real-world environments.
\begin{itemize}
    \item \textbf{Static/simulated environments.} \hspace{2mm}  These environments are fixed and typically preprocessed, yielding stable feedback and low noise. As a result, training is more stable than in real settings. For instance, ZeroSearch~\cite{sun2025zerosearchincentivizesearchcapability} simulates a search engine with an LLM and applies curriculum RL to progressively degraded synthetic documents, substantially reducing external API costs and stochastic noise. MaskSearch~\cite{wu2025masksearchuniversalpretrainingframework} develops ``deep research'' capabilities under controlled-noise conditions by using an offline retriever. ReSearch~\cite{chen2025researchlearningreasonsearch} jointly optimizes reasoning and retrieval via RL and is commonly trained on offline or semi-online, controlled corpora. AutoCoA~\cite{zhang2025agentmodelsinternalizingchainofaction} internalizes a world model to reduce reliance on real-world interactions. In aggregate, this class of methods makes a pragmatic trade-off: sacrificing a degree of real-world adaptivity in exchange for the benefits of lower interaction costs, more stable training, and greater reproducibility.
    
    \item \textbf{Dynamic/real-world environments.} \hspace{2mm} \setlength{\parindent}{1em} The environments in this category evolve over time or change during interaction; feedback is unstable and can be delayed, which may destabilize training and thus necessitates additional stabilization techniques. Researchers have developed several lines of work to address these challenges, primarily focusing on either mitigating environmental noise or designing more robust training methodologies.

\indent A primary strategy involves directly handling the noise and complexity of feedback from live environments. On code-oriented tasks, rStar2-Agent~\cite{shang2025rstar2agentagenticreasoningtechnical} explicitly tackles tool- and environment-induced noise in the coding loop, showing that mitigating task-irrelevant disturbances improves efficiency and reliability. Similarly, to manage noise during online training with live retrieval, Search-R1~\cite{jin2025searchr1trainingllmsreason} couples autonomous multi-turn querying with retrieval-token masking and outcome rewards to stabilize learning. Other approaches use intermediary modules to preprocess the environment's feedback. For instance, DeepResearcher~\cite{zheng2025deepresearcherscalingdeepresearch} introduces a dedicated web-browsing agent that supplies the main agent with a preprocessed, filtered, and high-quality information stream. WebDancer~\cite{wu2025webdancerautonomousinformationseeking} adopts a similar principle by compressing verbose and noisy web page content into a concise “evidence + summary” format to reduce context length and interference.

\indent  A second line of work focuses on the training framework itself to manage the inherent instability of online interaction. SkyRL~\cite{cao2025skyrl}, for example, provides a training framework for real-world tasks that uses high-throughput asynchronous parallelism to execute environments more efficiently, thereby reducing latency effects and demonstrating the feasibility of online RL for long-horizon, multi-tool tasks. Taking a hybrid approach, SimpleDeepSearcher~\cite{sun2025simpledeepsearcherdeepinformationseeking} synthesizes offline datasets from the real web. This strategy aims to mitigate the instability of purely online training while achieving a closer match to real-world conditions than purely offline or simulated setups.

\indent  Overall, these methods offer stronger domain fit and closer alignment with end-use applications, but they do so at the cost of significantly higher training and engineering complexity.
\end{itemize}

Beyond credit-assignment granularity and environment type, the choice of base models also critically influences the internalization of tool-use capabilities. Compared with non-reasoning base models, Large Reasoning Models (LRMs) possess stronger long-horizon reasoning and planning abilities, enabling more effective acquisition of tool use under both end-to-end and modular optimization. 

While a majority of studies integrate tools on non-reasoning base models~\cite{jin2025searchr1trainingllmsreason,wei2025autotir,da2025agentrlvr}, several recent end-to-end training studies (AutoCoA~\cite{zhang2025agentmodelsinternalizingchainofaction}, WebDancer~\cite{wu2025webdancerautonomousinformationseeking}, SkyRL~\cite{cao2025skyrl}, etc.) have begun to explore training on more capable LRMs. For instance, ReTool~\cite{feng2025retoolreinforcementlearningstrategic} trains tool integration within the same tool-augmented RL framework using, respectively, a non-reasoning base (Qwen2.5-32B-Instruct~\cite{qwen25_1m_2025}) and a reasoning base (DeepSeek-R1-Distill-Qwen-32B~\cite{deepseek_r1_2025}). The results show that, under matched training conditions, the reasoning base performs better, corroborating that base-model reasoning capacity enhances the learning of tool-integrated reasoning.

In summary, achieving robust capability internalization in end-to-end training requires a coordinated strategy that addresses three key factors: (1) the chosen granularity of credit assignment, (2) the challenges to training stability posed by the environment type, and (3) the selection of a foundation model with reasoning strength and tool-use alignment appropriate for the task.

\afterpage{ 
\clearpage  
\footnotesize

\begin{longtable}{C{1.65cm}|l|c|c|c|C{1.0cm}|c|c|c} 
\caption{Overview of agentic tool use methods, including task, model, tool (\textit{CI} = code interpreter; \textit{EM} = external model), environment (\textit{Environ.}) and other attributes.}\label{tab:tool_use}\\ 
\toprule
\rowcolor{blue!5}
\textbf{} & \textbf{Method} & \textbf{Task} & \textbf{Model} & \textbf{Tool} & \textbf{Environ.} & \textbf{Affiliation} & \textbf{Access} & \textbf{Date}\\
\midrule
\endfirsthead
\multicolumn{9}{c}{\tablename~\thetable\ {Overview of agentic tool use methods (continued).}}\\
\toprule
\rowcolor{blue!5}
\textbf{} & \textbf{Method} & \textbf{Task} & \textbf{Model} & \textbf{Tool} & \textbf{Environ.} & \textbf{Affiliation} & \textbf{Access} & \textbf{Date}\\
\midrule
\endhead

\midrule
\multicolumn{9}{r}{\textit{continued on next page}}\\
\endfoot

\bottomrule
\endlastfoot

\rowcolor{black!5}\multicolumn{9}{c}{\textit{\textbf{Pipeline-based Paradigm}}}\\
\midrule

\multirow{7}{*}{\makecell[c]{\textit{\textbf{System-based}}\\ \textit{\textbf{Workflow}}}}
& WebGPT ~\cite{DBLP:journals/corr/abs-2112-09332}     & Retrieval & LLM & API    & Dynamic & Industry & {No} & 21.12 \\
& SayCan ~\cite{ahn2022saycan}           & Others    & LLM & Others & Dynamic & Industry & \href{https://github.com/google-research/google-research/tree/master/saycan}{Yes}     & 22.04 \\
& WebShop ~\cite{yao2022webshop}          & Retrieval & LLM & UI     & Static  & Academia & \href{https://github.com/princeton-nlp/WebShop}{Yes} & 22.07 \\
& Code as Policies ~\cite{Liang2023CodeAsPolicies}& Others    & LLM & CI     & Dynamic & Industry & \href{https://github.com/google-research/google-research/tree/master/code_as_policies}{Yes} & 22.09 \\
& HuggingGPT ~\cite{Shen2023HuggingGPT}      & General   & LLM & EM     & Dynamic & Academia & \href{https://github.com/microsoft/JARVIS}{Yes} & 23.03 \\
& AutoGen ~\cite{wu2023autogen}          & General   & LLM & API    & Dynamic & Industry & \href{https://github.com/microsoft/autogen}{Yes} & 23.08 \\
& SWE-agent ~\cite{yang2024sweagent}        & Code      & LLM & CI     & Dynamic & Academia & \href{https://github.com/SWE-agent/SWE-agent}{Yes} & 24.05 \\
\midrule

\multirow{7}{*}{\makecell[c]{\textit{\textbf{Prompt-based}}\\ \textit{\textbf{Methods}}}}
& Self-Ask ~\cite{press2022selfask} & Retrieval & LLM & API & Dynamic & Academia & \href{https://github.com/ofirpress/self-ask}{Yes} & 22.10 \\
& ReAct ~\cite{yao2023react} & Hybrid & LLM & API & Static & Academia & \href{https://github.com/ysymyth/ReAct}{Yes} & 22.10 \\
& PAL ~\cite{pmlr-v202-gao23f} & Code & LLM & CI  & Static  & Academia & \href{https://github.com/reasoning-machines/pal}{Yes} & 22.11 \\
& PoT ~\cite{chen2023program}& Math & LLM & CI  & Static  & Academia & \href{https://github.com/TIGER-AI-Lab/Program-of-Thoughts}{Yes} & 22.11 \\
& Reflexion ~\cite{NEURIPS2023_1b44b878} & General & LLM & Others & Static  & Academia & \href{https://github.com/NoahShinn024/reflexion}{Yes} & 23.03 \\
& ViperGPT ~\cite{suris2023vipergpt} & Hybrid & LMM & EM  & Static  & Academia & \href{https://github.com/cvlab-columbia/viper}{Yes} & 23.03 \\
& CRITIC ~\cite{gou2024critic} & Hybrid & LLM & CI  & Dynamic & Industry & \href{https://github.com/microsoft/ProphetNet/tree/master/CRITIC}{Yes} & 23.05 \\
\midrule

\rowcolor{black!5}\multicolumn{9}{c}{\textit{\textbf{Model-native Paradigm}}}\\
\midrule

\multirow{3}{*}{\makecell[c]{\textit{\textbf{Modular}}\\ \textit{\textbf{Training}}}}
& Agent-as-Tool ~\cite{zhang2025agentastoolstudyhierarchicaldecision}& Retrieval & LLM & API & Dynamic & Academia & {No} & 25.07 \\
& AI-SearchPlanner ~\cite{mei2025aisearchplannermodularagenticsearch}& Retrieval & LRM & API & Dynamic & Industry & {No} & 25.08 \\
& RLTR ~\cite{li2025encouraginggoodprocessesneed}& General & LRM & API & Dynamic & Academia & {No} & 25.08 \\
\midrule

\multirow{34}{*}{\makecell[c]{\textit{\textbf{End-to-end}}\\ \textit{\textbf{Training}}}}
& R1-Searcher ~\cite{song2025r1}& Retrieval & LLM & API & Static & Academia & \href{https://github.com/RUCAIBox/R1-Searcher}{Yes} & 25.03 \\
& ReSearch ~\cite{chen2025researchlearningreasonsearch}& Retrieval & LLM & API & Static & Academia & \href{https://github.com/Agent-RL/ReCall}{Yes} & 25.03 \\
& ToRL ~\cite{li2025torl} & Math & LLM & CI & Dynamic & Academia & \href{https://github.com/GAIR-NLP/ToRL}{Yes} & 25.03 \\
& Search-R1 ~\cite{jin2025searchr1trainingllmsreason}& Retrieval & LLM & API & Static & Academia & \href{https://github.com/PeterGriffinJin/Search-R1}{Yes} & 25.03 \\
& AutoCoA ~\cite{zhang2025agentmodelsinternalizingchainofaction}& Retrieval & LRM & API & Static & Academia & \href{https://github.com/ADaM-BJTU/AutoCoA}{Yes} & 25.03 \\
& ReTool ~\cite{feng2025retoolreinforcementlearningstrategic}& Math & LRM & CI & Dynamic & Industry & \href{https://github.com/ReTool-RL/ReTool}{Yes} & 25.04 \\
& ToolRL ~\cite{qian2025toolrl} & General & LLM & API & Dynamic & Academia & \href{https://github.com/qiancheng0/ToolRL}{Yes} & 25.04 \\
& OTC ~\cite{wang2025otc} & Retrieval & LLM & API & Static & Academia & {No} & 25.04 \\
& DeepResearcher ~\cite{zheng2025deepresearcherscalingdeepresearch}& Retrieval & LLM & API & Dynamic & Academia & \href{https://github.com/GAIR-NLP/DeepResearcher}{Yes} & 25.04 \\
& SWiRL ~\cite{swirl2025} & Retrieval & LLM & API & Static & Academia & {No} & 25.04 \\
& ARTIST ~\cite{singh2025agenticreasoningtoolintegration}& General & LLM & API & Static & Industry & {No} & 25.04 \\
& WebThinker ~\cite{li2025webthinker} & Retrieval & LRM & API & Dynamic & Academia & \href{https://github.com/RUC-NLPIR/WebThinker}{Yes} & 25.04 \\
& RAGEN ~\cite{wang2025ragenunderstandingselfevolutionllm}& General & LLM & API & Dynamic & Academia & \href{https://github.com/RAGEN-AI/RAGEN}{Yes} & 25.04 \\
& WebDancer ~\cite{wu2025webdancerautonomousinformationseeking}& Retrieval & LRM & API & Dynamic & Industry & \href{https://github.com/Alibaba-NLP/WebAgent}{Yes} & 25.05 \\
& Tool-N1 ~\cite{zhang2025tooln1} & General & LLM & API & Static & Industry & \href{https://github.com/NVlabs/Tool-N1}{Yes} & 25.05 \\
& Satori-SWE ~\cite{zeng2025satoriswe} & Code & LLM & CI & Dynamic & Academia & \href{https://github.com/satori-reasoning/Satori-SWE}{Yes} & 25.05 \\
& MaskSearch ~\cite{wu2025masksearchuniversalpretrainingframework}& Retrieval & LLM & API & Static & Academia & \href{https://github.com/Alibaba-NLP/MaskSearch}{Yes} & 25.05 \\
& SkyRL ~\cite{cao2025skyrl}& Code & LRM & CI & Dynamic & Academia & \href{https://github.com/NovaSky-AI/SkyRL}{Yes} & 25.05 \\
& ZeroSearch ~\cite{sun2025zerosearchincentivizesearchcapability}& Retrieval & LLM & API & Static & Industry & \href{https://github.com/Alibaba-NLP/ZeroSearch}{Yes} & 25.05 \\
& Agent RL Scaling~\cite{mai2025agentrlscaling} & Math & LLM & CI & Dynamic & Academia & \href{https://github.com/yyht/openrlhf_async_pipline}{Yes} & 25.05 \\
& GIGPO ~\cite{feng2025groupingrouppolicyoptimizationllm}& Retrieval & LLM & API & Static & Academia & \href{https://github.com/langfengQ/verl-agent}{Yes} & 25.05 \\
& VTool-R1 ~\cite{wu2025vtoolr1vlmslearnthink}& Others & LMM & API & Static & Academia & \href{https://github.com/VTool-R1/VTool-R1}{Yes} & 25.05 \\
& DeepEyes ~\cite{zheng2025deepeyesincentivizingthinkingimages}& Others & LMM & API & Static & Industry & \href{https://github.com/Visual-Agent/DeepEyes}{Yes} & 25.05 \\
& Multi-Turn-RL ~\cite{zeng2025reinforcingmultiturnreasoningllm}& Others & LLM & UI & Dynamic & Industry & \href{https://github.com/SiliangZeng/Multi-Turn-RL-Agent}{Yes} & 25.05 \\
& StepSearch ~\cite{wang2025stepsearchignitingllmssearch}& Retrieval & LLM & API & Static & Industry & \href{https://github.com/Zillwang/StepSearch}{Yes} & 25.05 \\
& Spa-RL ~\cite{wang2025sparlreinforcingllmagents}& General & LLM & Others & Staic & Academia & \href{https://github.com/WangHanLinHenry/SPA-RL-Agent}{Yes} & 25.05 \\
& O\textsuperscript{2}-Searcher ~\cite{mei2025o2searcher} & Retrieval & LLM & API & Static & Academia & \href{https://github.com/Acade-Mate/O2-Searcher}{Yes} & 25.05 \\
& MMSearch-R1 ~\cite{wu2025mmsearchr1} & Retrieval & LMM & API & Dynamic & Academia & \href{https://github.com/EvolvingLMMs-Lab/multimodal-search-r1}{Yes} & 25.06 \\
& Agent-RLVR ~\cite{da2025agentrlvr} & Code & LLM & CI & Dynamic & Industry & {No} & 25.06 \\
& AutoTIR ~\cite{wei2025autotir} & Retrieval & LLM & API & Static & Academia & \href{https://github.com/weiyifan1023/AutoTIR}{Yes} & 25.07 \\
& Agent Lightning ~\cite{luo2025agentlightning} & Hybrid & LLM & API & Staic & Industry & \href{https://github.com/microsoft/agent-lightning}{Yes} & 25.08 \\
& FunRL~\cite{hao2025exploringsuperiorfunctioncalls} & General & LLM & API & Dynamic & Academia & \href{https://github.com/BingguangHao/RLFC}{Yes} & 25.08 \\
& rStar2-Agent ~\cite{shang2025rstar2agentagenticreasoningtechnical}& Math & LRM & CI & Dynamic & Industry & \href{https://github.com/microsoft/rStar}{Yes} & 25.08 \\
& ASearcher ~\cite{gao2025asearcher} & Retrieval & LRM & API & Dynamic & Academia & \href{https://github.com/inclusionAI/ASearcher}{Yes} & 25.08 \\

\end{longtable}
}
\subsection{Summary and Discussion}\label{sec:4.4}
This section has reviewed the evolution of agentic tool use, elaborating its progression from the externally orchestrated pipeline-based paradigm to the internalized model-native paradigm. This shift represents a move away from predefined processes toward greater decision autonomy and adaptability in open environments. In the model-native approach, tool use is decomposed into two interdependent layers, planning and execution, presenting a multi-objective optimization problem where the model must produce logical action sequences while also reliably executing API calls and interpreting environmental feedback. The reviewed representative studies are summarized in Table~\ref{tab:tool_use}. 

The ongoing model-native tool use research faces two challenges. The first is \textit{credit assignment}: how to reliably attribute a final outcome to the specific decision steps in a long action sequence. The second is \textit{environmental noise}: how to maintain training stability when faced with uncertainties such as tool timeouts, inconsistent returns, and dynamic content.

To address these issues, research is likely to  progress along several directions. As a pragmatic solution, a trend has emerged toward back to modular training. By decoupling the planner from the executor, the learning of the core decision-making policy can be isolated from the noise generated by the execution layer, which has been shown to improve both sample efficiency and training stability. This can help distinguish between model decision errors and environmental interference.

Within end-to-end training, the refinement is expected from coarse, trajectory-level rewards toward more granular, step-level or turn-level credit assignment. Since tool use is often a multi-turn interactive process, relying solely on a final outcome signal dilutes the contribution of key intermediate decisions. Methods that decompose rewards or estimate advantages at a finer granularity help align the learning signal more directly with effective actions, thereby stabilizing training. Simultaneously, training environments are gradually shifting from low-cost, controllable, low-noise static simulators to dynamic real-world settings to narrow the simulation-to-reality gap. The latter provides more complex and diverse scenarios, facilitating more robust and generalizable policies.

Based on this analysis, future research on model-native tool use will likely advance along the following directions:
\begin{itemize}
    \item \textbf{Hybrid architectures.} ~While pure end-to-end training holds the highest theoretical performance ceiling, it often suffers from the training instabilities discussed. Conversely, fully modular designs, though easier to debug and optimize, may sacrifice synergy between components. A key research direction will be the development of hybrid architectures that strategically combine modular and end-to-end training to balance these trade-offs.

    \item  \textbf{Robust training in open environments.} ~The transition from simulated to real, open environments requires models to handle significant challenges related to latency, noise, and uncertainty. This necessitates the construction of more stable and sample-efficient training frameworks capable of learning reliable policies from inconsistent real-world feedback.

    \item \textbf{From tool user to tool creator.} ~Current research primarily focuses on using a predefined set of tools. A significant leap will be to enable agents to dynamically create, compile, and validate new tools based on task demands. This represents a fundamental shift from leveraging existing resources to creatively generating novel solutions.
\end{itemize}

\section{Core Capability: Memory}\label{sec:5}

\subsection{Overview}
In LLM-based agent systems, memory is evolving from a single external module into a set of capabilities integral to the entire task lifecycle. Its functions include preserving historical and world states, selectively organizing factual evidence within a limited context, and injecting this evidence to support planning, tool use, and reasoning. This section conceptualizes memory as a form of \textit{action-oriented evidence governance}: its role is not only to store information effectively but also to ensure that information is utilized proficiently to guide actions. Accordingly, this deconstructs the memory process into four core functions: \textit{storing}, the decision of what information to write; \textit{managing}, the organization and compression of that information; \textit{retrieval}, the extraction of relevant evidence when needed; and \textit{utilization}, the effective employment of that evidence in the agent's reasoning and actions.

Two primary paradigms have emerged to implement these memory functions. The first, the pipeline-based paradigm, relies on external workflows to orchestrate information. This approach uses techniques such as indexing, summarization, routing, and rule-based strategies to manage data before and after the model's reasoning process. While this method offers advantages in controllability, interpretability, and ease of implementation, it is often limited by a performance ceiling caused by error accumulation across modules and inflexibility in long-tail scenarios. The second, the model-native paradigm, aims to internalize memory capabilities directly within the model. This is achieved through methods like internal model editing or by integrating memory-related workflows into the model's training process. By learning a unified policy for storing, managing, and retrieving information, the model-native paradigm holds a higher potential for performance but requires more substantial data, computational resources, and training stability. In practice, these two paradigms are not mutually exclusive and often coexist within real-world systems.

From a developmental perspective, many advancements that enabled modern agent memory did not initially target it as an explicit goal. Instead, they emerged from parallel research in system optimization and model architecture, such as KV-cache management~\cite{streamingllm_project,h2o_2023,scissorhands_2023}, efficient attention mechanisms~\cite{longformer_paper,bigbird_neurips,flashattention_neurips}, and position encoding extrapolation~\cite{roformer_rope,ALiBiarxiv}. These foundational technologies effectively expanded the model's capacity for short-term information retention. Building on this, engineering practices established the RAG paradigm as the default baseline, using external pipelines to manage knowledge~\cite{densepassageretrieval20}. The recent paradigm shift is defined by the internalization of these processes, as models learn a unified policy for deciding what to store, how to manage it, and when to retrieve and utilize it through end-to-end training.

To systematically review this evolution, this section divides memory into two primary layers: \textit{short-term memory} and \textit{long-term memory}. Our discussion of short-term memory follows its own developmental trajectory. We begin with techniques for \textit{long context}, which address the fundamental challenge of processing large volumes of information within a single session~\cite{longformer_paper,flashattention_neurips,bigbird_neurips}. We then examine \textit{context management}, a more advanced capability focused on the selective organization and utilization of information to mitigate attention dilution and improve reasoning quality~\cite{selfrag_2023}. In contrast, the section on long-term memory reviews methods for persistently storing knowledge across sessions, either in external repository or within model parameters~\cite{awm2024,amem2025,mplus_2025,memorybank23}. Notably, since long-term memory content is typically not rewritable online, the paradigm shift in this area focuses more on internalizing retrieval and usage strategies rather than the entire storage process.

\subsection{Short-term Memory: Long Context}

This subsection examines how an agent, within a single inference session, can reliably process a large set of task-relevant evidence. The effectiveness of long-context processing is not determined solely by the absolute size of the context window but by the model's ability to use the information contained within it. We analyze this capability across three ascending levels of difficulty:
\begin{itemize}
    \item \textbf{Retrieval}. This is the foundational ability to accurately identify and pinpoint task-relevant information within a long and potentially noisy input.
    \item \textbf{Basic Reasoning}. This level involves performing basic, single-step operations on one or more localized pieces of evidence, such as extraction, summarization, or straightforward logical comparisons.
    \item \textbf{Complex Reasoning}. The most advanced level, this requires the model to integrate multiple, often distributed, pieces of evidence to construct a complete causal or logical chain for solving multi-step problems.
\end{itemize}

\begin{figure}[H] 
  \centering 
  \includegraphics[width=0.98\textwidth]{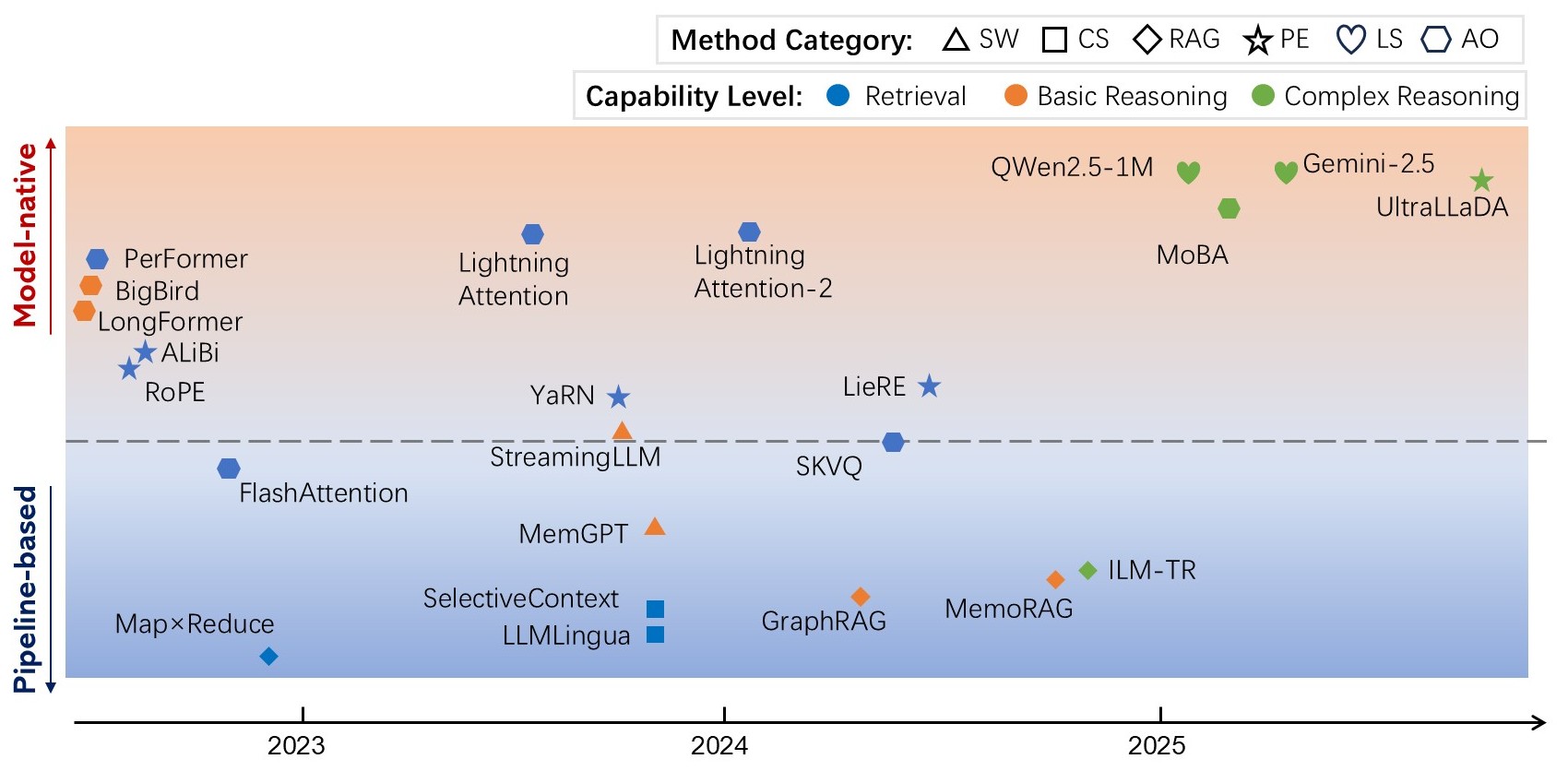} 
  \caption{Overview of long context methods for short-term memory. Method categories include: SW (Sliding Window), CS (Compression \& Summarization), RAG (Retrieval-Augmented Generation), PE (Position Encoding extrapolation), LS (Long-sequence Synthesis), and AO (Attention Optimization).} 
  \label{fig:longcontext} 
\end{figure}

\afterpage{%
\clearpage
\footnotesize 
\begin{longtable}{>{\centering\arraybackslash}p{1.9cm}|l|>{\centering\arraybackslash}p{2.5cm}|>{\centering\arraybackslash}p{1.6cm}|>{\centering\arraybackslash}p{1.4cm}|>{\centering\arraybackslash}p{0.7cm}|>{\centering\arraybackslash}p{0.6cm}}
\caption{Overview of long context methods for short-term memory.}\label{tab:longcontext}\\
\midrule
\rowcolor{blue!5}
\textbf{} &
\textbf{Method} &
\makecell{\textbf{Long-Context}\\ \textbf{Adaptivity}} &
\makecell{\textbf{Capability}\\ \textbf{Level}} &
\textbf{Affiliation} &
\textbf{Access} &
\textbf{Date} \\
\midrule
\endfirsthead

\multicolumn{7}{c}{\tablename~\thetable: Overview of long-context methods for short-term memory.} \\
\midrule
\rowcolor{blue!5}
\textbf{} &
\textbf{Method} &
\makecell{\textbf{Long-Context}\\ \textbf{Adaptivity}} &

\makecell{\textbf{Capability}\\ \textbf{Level}} &
\textbf{Affiliation} &
\makecell{\textbf{Open-}\\ \textbf{Source}} &
\textbf{Date} \\
\midrule
\endhead

\multicolumn{7}{r@{}}{(continued on next page)} \\
\endfoot

\endlastfoot

\rowcolor{black!5}
\multicolumn{7}{c}{\textit{\textbf{Pipeline-based Paradigm}}} \\ \midrule

\multirow{3}{*}{\makecell[c]{\textit{\textbf{Sliding}}\\ \textit{\textbf{Window}}}}

& StreamingLLM~\cite{streamingllm_project} & Streaming to \textasciitilde{} millions; 8K window & Basic Rea. & Academia & \href{https://github.com/mit-han-lab/streaming-llm}{Yes} & 23.09 \\
& MemGPT~\cite{memgpt2024} & Streaming, infinite & Basic Rea. & Academia & \href{https://github.com/letta-ai/letta}{Yes} & 23.10 \\
\midrule

\multirow{2}{*}{\makecell[c]{\textit{\textbf{Compression\&}}\\ \textit{\textbf{Summarization}}}}
& SelectiveContext~\cite{selectivecontext_emnlp} & Compression ($\geq$2$\times$) & Retrieval & Academia & \href{https://github.com/liyucheng09/Selective_Context}{Yes} & 23.10 \\
& LLMLingua~\cite{llmlingua_ms} & Compression (3\textasciitilde{}20$\times$) & Retrieval & Industry & \href{https://github.com/microsoft/LLMLingua}{Yes} & 23.10 \\
\midrule

\multirow{7}{*}{\makecell[c]{\textit{\textbf{RAG}}}}
& MapReduce~\cite{mapreduce_langchain} & Variable (chunked) & Retrieval & Industry & \href{https://github.com/langchain-ai/langchain}{Yes} & 22.10 \\
& GraphRAG~\cite{edge2024local} & Variable (graph retrieval) & Basic Rea. & Industry & \href{https://github.com/microsoft/graphrag}{Yes} & 24.04 \\
& MemoRAG~\cite{memorag_www25} & Variable (memory RAG) & Basic Rea. & Academia/\allowbreak Industry & \href{https://github.com/qhjqhj00/MemoRAG}{Yes} & 24.09 \\
& ILM\mbox{-}TR~\cite{ilmtr} & Variable (iterative retrieval) & Complex Rea. & Academia & No & 24.10 \\
\midrule

\rowcolor{black!5}
\multicolumn{7}{c}{\textit{\textbf{ Model-native Paradigm}}} \\ \midrule

\multirow{6}{*}{\makecell[c]{\textit{\textbf{Position}}\\\textit{\textbf{Encoding}}\\\textit{\textbf{Extrapolation}}}}
& RoPE~\cite{roformer_rope} & Unknown & Retrieval & Academia & \href{https://huggingface.co/docs/transformers/model_doc/roformer}{Yes} & 21.04 \\
& ALiBi~\cite{ALiBiarxiv} & Unknown & Retrieval & Academia & \href{https://github.com/ofirpress/attention_with_linear_biases}{Yes} & 21.08 \\
& YaRN~\cite{peng2024yarn} & 128K & Retrieval & Academia & \href{https://github.com/jquesnelle/yarn}{Yes} & 23.09 \\
& LieRE~\cite{LieREarxiv} & Unknown (multimodal) & Retrieval & Academia & \href{https://github.com/StanfordMIMI/LieRE}{Yes} & 24.06 \\
& UltraLLaDA~\cite{ultrallada25} & 128K & Complex Rea. & Academia & \href{https://github.com/Relaxed-System-Lab/UltraLLaDA}{Yes} & 25.10 \\
\midrule

\multirow{2}{*}{\makecell[c]{\textit{\textbf{Long-sequence}}\\ \textit{\textbf{Synthesis}}}}
& Qwen2.5\mbox{-}1M~\cite{qwen25_1m_2025} & 1M & Complex Rea. & Industry & \href{https://huggingface.co/collections/Qwen/qwen25-1m-679325716327ec07860530ba}{Yes} & 25.01 \\
& Gemini~\mbox{2.5}~\cite{gemini25} & 1M & Complex Rea. & Industry & No & 25.03 \\
\midrule

\multirow{9}{*}{\makecell[c]{\textit{\textbf{Attention}}\\ \textit{\textbf{Optimization}}}}
& Longformer/LED~\cite{longformer_paper} & 16K & Basic Rea. & Academia & \href{https://github.com/allenai/longformer}{Yes} & 20.04 \\
& BigBird~\cite{bigbird_neurips} & 8K & Basic Rea. & Industry & \href{https://github.com/google-research/bigbird}{Yes} & 20.07 \\
& Performer~\cite{performer_arxiv} & 4K+ & Retrieval & Academia & \href{https://github.com/lucidrains/performer-pytorch}{Yes} & 20.09 \\
& FlashAttention~\cite{flashattention_neurips} & N/A & Retrieval & Academia/\allowbreak Industry & \href{https://github.com/Dao-AILab/flash-attention}{Yes} & 22.05 \\
& LightningAttention~\cite{lightningattention} & N/A & Retrieval & Academia & \href{https://github.com/OpenNLPLab/lightning-attention}{Yes} & 23.07 \\
& LightningAttention\mbox{-}2~\cite{lightningattention2} & Theoretically Infinite & Retrieval & Academia & \href{https://github.com/OpenNLPLab/lightning-attention}{Yes} & 24.01 \\
& SKVQ~\cite{skvqarxiv} & 1M & Retrieval & Academia & \href{https://github.com/cat538/SKVQ}{Yes} & 24.05 \\
& MoBA~\cite{moba2025} & Scalable & Complex Rea. & Industry & \href{https://github.com/MoonshotAI/MoBA}{Yes} & 25.02 \\
\midrule

\end{longtable}
}

Consequently, effectively handling long contexts requires coupling the window's physical capacity with training strategies that promote its stable use. When these mechanisms fail, systems may either overlook crucial facts or suffer from attention dilution. This can lead to an illusion of perception without actual use, a phenomenon described as the ``Lost in the Middle'' problem~\cite{liu2023lost}. The overview of long context methods for short-term memory is illustrated in Fig.~\ref{fig:longcontext}. More details are summarized in Table~\ref{tab:longcontext}.

\subsubsection{\textbf{Pipeline-based Paradigm}} \hspace{2mm}
Pipeline-based methods manage long contexts by externally processing information before it is fed to the model. These modular strategies can be grouped into several method categories:

\paragraph{\textbf{Sliding Window}} \hspace{2mm}
The sliding window approach is a simple and broadly applicable baseline for managing context. Under a fixed computational budget, it concatenates recent interaction segments with critical history, such as initial instructions, based on temporal or topical continuity. This method is easy to implement, computationally light, and provides predictable latency, making it suitable for tasks with strong topical continuity like dialogue or code completion. However, its limitations become apparent when relevant evidence is distant or requires cross-segment alignment, as proximity-based truncation can permanently discard key facts. In practice, this baseline is often extended with more sophisticated mechanisms, such as streaming strategies and dedicated memory controllers, which have been developed to create more dynamic and efficient systems~\cite{streamingllm_project, memgpt2024}.

\paragraph{\textbf{Compression and Summarization}} \hspace{2mm}
When the context window is constrained but the history is extensive, hierarchical and selective summarization can increase information density while preserving core semantics. Traditional summarization techniques are often irreversible and produce highly abstract text, saving space at the cost of detail and traceability. To overcome this, contemporary systems increasingly construct structured evidence cards with citation anchors that point back to exact source locations. This approach effectively turns summaries into a session state cache that can be decompressed on demand, a technique explored in systems such as Selective Context~\cite{selectivecontext_emnlp} and LLMLingua~\cite{llmlingua_ms}. During inference, the model can operate on compact summaries while retaining the ability to rapidly follow anchors to verify details, thereby maintaining interpretability and a coherent reasoning chain.

\paragraph{\textbf{Retrieval-Augmented Generation}} \hspace{2mm}
When the dialogue context far exceeds the model's budget and the necessary knowledge resides in external stores, RAG dynamically constructs the working context for each turn. Compared with traditional RAG pipelines, which typically treat retrieval as a stateless, per-query operation, conversational variants are session-aware: they employ query expansion to capture dialogue intent and multi-path retrieval with reranking to optimize the evidence set across turns. This approach excels at knowledge-intensive, needle-in-a-haystack problems like log analysis and technical question answering. Various frameworks have demonstrated this approach, from those employing MapReduce paradigms~\cite{mapreduce_langchain} to more advanced systems that utilize graph-based structures~\cite{edge2024local} or memory-augmented techniques~\cite{memorag_www25, ilmtr}. A key challenge remains that excessive injection of retrieved information can lead to attention dilution. Therefore, practical deployments often treat long windows and session RAG as complementary, using the window to cover retrieval misses and falling back to precise injection when the window becomes too noisy.

\subsubsection{\textbf{Model-native Paradigm}} \hspace{2mm}
Model-native solutions enhance the model's core architecture to directly handle longer sequences. These methods primarily fall into the following three categories: 

\paragraph{\textbf{Position Encoding Extrapolation}} \hspace{2mm}
The first challenge for long-context short-term memory is that models pretrained on shorter sequences struggle to generalize to longer ones. Architectural modifications to positional encodings address this. For instance, Rotary Position Embedding (RoPE)~\cite{roformer_rope} uses rotations to encode relative positions, which inherently improves length extrapolation. Other methods, such as Attention with Linear Biases (ALiBi)~\cite{ALiBiarxiv}, inject distance-aware penalties into attention scores to aid generalization at inference time. Subsequent refinements like YaRN~\cite{peng2024yarn} and LieRE~\cite{LieREarxiv} further stabilize behavior in the extrapolation regime. However, research has shown that these architectural changes are most effective when combined with continued training on long-sequence data. For example, UltraLLaDA~\cite{ultrallada25} couples position encoding extrapolation with targeted training to extend context stability to 128K tokens for retrieval and 32K tokens for complex reasoning, demonstrating that training is critical for activating a model's latent long-context abilities.

\paragraph{\textbf{Long-sequence Synthesis}} \hspace{2mm}
Expanding the context window's capacity does not automatically teach a model how to use it effectively. Therefore, task-driven training curricula are central to developing this capability. These curricula often include tasks like ``needle-in-a-haystack'' extraction and cross-document reasoning, which explicitly incentivize the model to attend to distant information. Reports on large models such as Qwen2.5-1M~\cite{qwen25_1m_2025} and Gemini2.5~\cite{gemini25} confirm that curricula aligned with real-world applications markedly improve performance on ultra-long context tasks. Training typically follows a schedule of increasing difficulty, starting with shorter sequences and gradually extending the length, number of distractors, and reasoning complexity.

This curriculum-centric approach functions as a bridge from pipeline-based engineering to model-native capability. With sufficient training pressure, models can internalize functions once delegated to external modules like retrievers and re-rankers. The result is a shift away from reliance on external workflows and toward robust, end-to-end long-context reasoning.

\paragraph{\textbf{Attention Optimization}} \hspace{2mm}
The quadratic computational cost of the standard self-attention mechanism makes processing very long sequences prohibitively expensive. This has spurred the development of more efficient attention algorithms. Algorithmic methods reduce this cost through sparse connectivity patterns, as seen in models like Longformer \cite{longformer_paper} and BigBird \cite{bigbird_neurips}, or through linear and kernelized approximations such as the Performer \cite{performer_arxiv}. In parallel, systems-level optimizations focus on improving hardware utilization. A notable example is FlashAttention \cite{flashattention_neurips}, which preserves exact attention while reducing high-bandwidth memory access through kernel fusion. This line of work has been extended by subsequent improvements like Lightning Attention \cite{lightningattention, lightningattention2}. Additional cache-aware and shape-adaptive variants further refine attention allocation to better match task structure \cite{skvqarxiv, moba2025}. Together, these advances are shifting long-context capabilities from an exotic feature to a practical default, providing a solid substrate for higher-level algorithms and applications.

\newcommand{\cellL}[1]{\parbox[t]{\linewidth}{\raggedright #1\strut}}
\begin{figure}[t] 
  \centering 
  \includegraphics[width=0.98\textwidth]{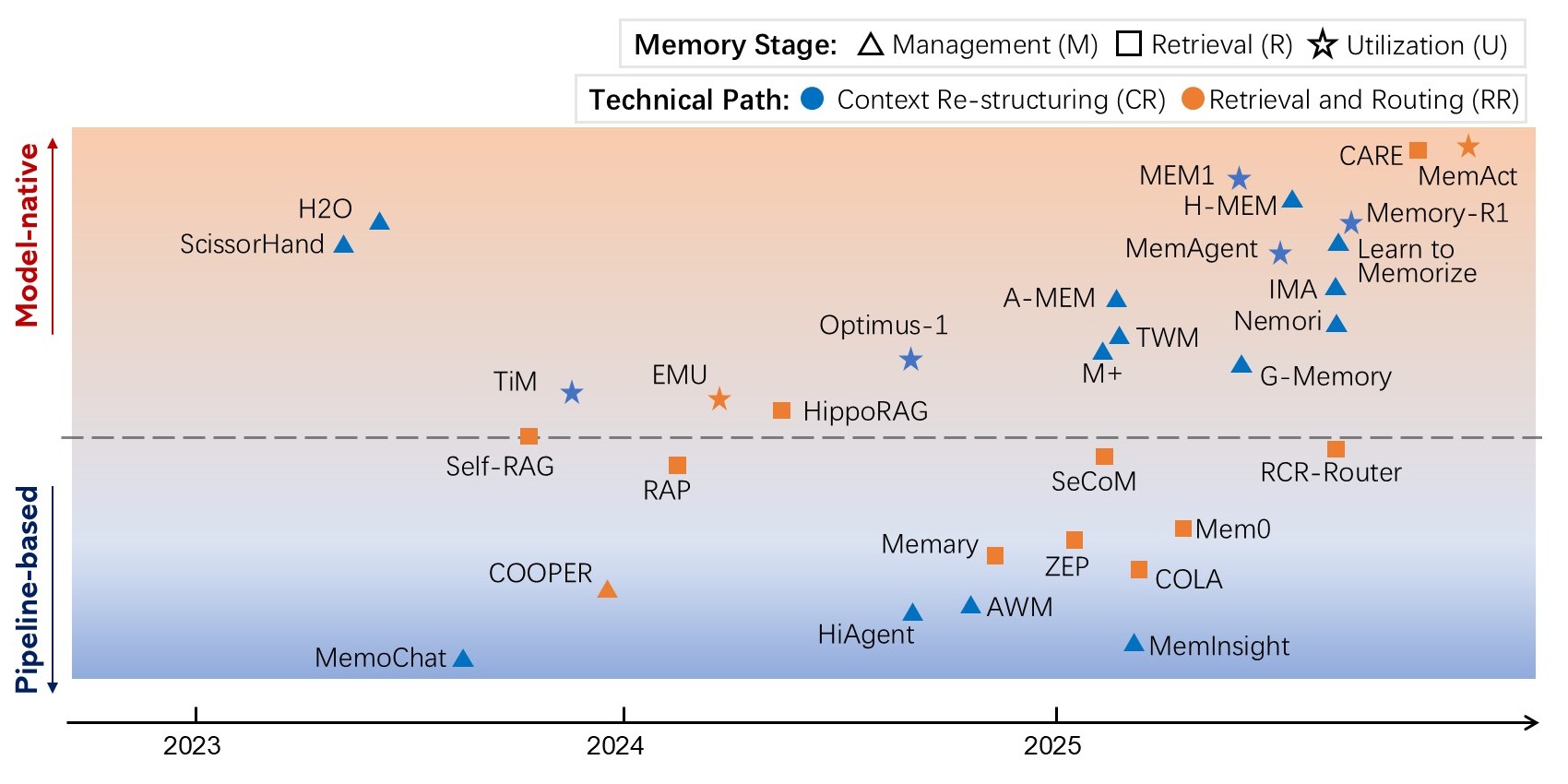} 
  \caption{Overview of context management methods for short-term memory. } 
  \label{fig:ContetxManagement} 
\end{figure}

\subsection{Short-term Memory: Context Management}
While long-context capabilities expand the \textit{volume} of information a model can process, context management addresses the \textit{quality} and \textit{relevance} of that information. The central challenge is attention dilution: even when the total input length is within a model's theoretical limits, irrelevant or redundant data can severely impair reasoning performance. In the ``Needle in a Haystack'' test, for instance, model performance often drops significantly when key information is placed in the middle of a long, distracting context~\cite{liu2023lost,laban2025llms}.

Effective context management aims to produce a sparse yet high-value stream of evidence that maximizes relevance while minimizing attention dilution. The evolution of these techniques can be examined along two core paths: \textit{Context Re-structuring (CR)}, which focuses on optimizing the presentation and ordering of information within the context window, and \textit{Retrieval and Routing (RR)}, which concerns the filtering and recall of information from external sources or internal history.

While the complete memory cycle includes an initial \textit{storing} phase, the core challenge of context management lies in dynamic, in-session processes. Therefore, we categorize the methods in this section based on their primary focus within three main stages: \textit{Management (M)}, \textit{Retrieval (R)}, and \textit{Utilization (U)}. To clarify application scenarios, we also distinguish between single-agent and multi-agent paradigms. In single-agent systems, context management prioritizes evidence selection for a single model, whereas in multi-agent systems, it must also handle inter-agent messaging, role-conditioned retrieval, and shared memory. 

We further divide the progression toward full internalization into two distinct stages: the \textit{hybrid} paradigm and the fully \textit{model-native} paradigm. It reflects the difference between learning to operate components within a system versus the model becoming the system itself. The hybrid paradigm represents a transitional stage where pipeline-based architectures are retained, but individual rule-based modules are replaced with learnable components (\textit{modular optimization}), or the model is trained to execute a predefined external workflow (\textit{workflow internalization}). In contrast, the fully model-native paradigm eliminates this external scaffolding entirely, internalizing the entire memory management process into a single, unified policy learned end-to-end. The overview of context management methods for short-term memory is illustrated in Fig.~\ref{fig:ContetxManagement}. More details are summarized in Table~\ref{tab:long_paradigm_cn}.

\subsubsection{\textbf{Pipeline-based Paradigm}} \hspace{2mm}
The pipeline-based paradigm for context management is often referred to as \textit{context engineering}~\cite{context_engineering_survey_2025}, where information is manipulated externally through rule-based or modular workflows before being passed to the model.

\paragraph{\textbf{Context Re-structuring}} \hspace{2mm} 
In this technical path, context re-structuring involves the static pre-arrangement of input. Techniques include chunking and reordering content based on relevance, such as placing critical information at the beginning or end of a prompt to circumvent the ``lost-in-the-middle'' effect. Structured prompt templates, which assign fixed positions for instructions, examples, and retrieved evidence, are also common, as seen in systems like MemoChat and MemInsight~\cite{memochat2023,meminsight2025}. For specialized domains, hierarchical organization may be used to manage complex evidence and resolve conflicts, as demonstrated by HiAgent and AWM~\cite{hiagent2024,awm2024}.

\paragraph{\textbf{Retrieval and Routing}} \hspace{2mm}
When evidence is drawn from heterogeneous sources, retrieval and routing becomes critical. Retrieval systems typically combine sparse and dense methods to ensure both keyword and semantic coverage~\cite{densepassageretrieval20}. A rule-based routing module then selects the most appropriate information based on user intent or knowledge domain. To increase information density, diversity constraints may be applied to reduce redundancy. This approach is exemplified by systems like Memary, and COLA, which offer auditable selection policies~\cite{memary2024,cola2025}. Production systems like ZEP and Mem0 further integrate joint sparse-dense retrieval for fast and semantically faithful recall~\cite{zep2025,mem0_2025}. While controllable and auditable, the rigidity of these rule-based policies limits their adaptability, motivating a shift toward hybrid, learnable approaches.

\afterpage{%
\clearpage
\footnotesize 
\begin{longtable}{>{\centering\arraybackslash}p{2.6cm}|>{\raggedright\arraybackslash}p{2.0cm}|>{\centering\arraybackslash}p{1.0cm}|>{\centering\arraybackslash}p{0.4cm}|>{\centering\arraybackslash}p{0.4cm}|>{\centering\arraybackslash}p{0.4cm}|>{\centering\arraybackslash}p{1.0cm}|>{\centering\arraybackslash}p{1.2cm}|>{\centering\arraybackslash}p{0.8cm}|l}
\caption{Overview of context management methods for short-term memory}\label{tab:long_paradigm_cn}\\
\midrule
\multirow{2}{*}{\makecell{\textbf{Method} \\ \textbf{Category}}}  &
\multirow{2}{*}{\textbf{Method}} &
\multirow{2}{*}{\makecell{\textbf{Tech.} \\ \textbf{Path}}} &
\multicolumn{3}{c|}{\textbf{Main Stage}} &
\multirow{2}{*}{\makecell{\textbf{Agent} \\ \textbf{Archi.}}} &
\multirow{2}{*}{\textbf{Affiliation}} &
\multirow{2}{*}{\textbf{Access}}  &
\multirow{2}{*}{\textbf{Date}}\\
\cline{4-6} 
& & &\textbf{M} &\textbf{R} &\textbf{U} & & & & \\
\midrule
\endfirsthead

\multicolumn{10}{c}{\tablename~\thetable: Overview of context management methods for short-term memory (continued)} \\
\midrule
\multirow{2}{*}{\makecell{\textbf{Method} \\ \textbf{Category}}} &
\multirow{2}{*}{\textbf{Method}} &
\multirow{2}{*}{\makecell{\textbf{Tech.} \\ \textbf{Path}}} &
\multicolumn{3}{c|}{\textbf{Main Stage}} &
\multirow{2}{*}{\makecell{\textbf{Agent} \\ \textbf{Archi.}}} &
\multirow{2}{*}{\textbf{Affiliation}} &
\multirow{2}{*}{\textbf{Access}} &
\multirow{2}{*}{\textbf{Date}}\\ \cline{4-6} 
& & & \textbf{M} & \textbf{R} & \textbf{U} & & & & \\
\midrule
\endhead

\multicolumn{10}{r@{}}{(continued on next page)} \\
\endfoot

\endlastfoot

\rowcolor{black!5}
\multicolumn{10}{c}{\textit{\textbf{Pipeline-based Paradigm}}} \\ \midrule

\multirow{4}{*}{\textit{\textbf{\makecell{Prompt Structuring}}}}
  & MemoChat~\cite{memochat2023} & CR & \checkmark &  &  & Single & Academia & \href{https://github.com/LuJunru/MemoChat}{Yes} & 23.08 \\
  & HiAgent~\cite{hiagent2024} & CR & \checkmark &  &  & Single & Academia & \href{https://github.com/HiAgent2024/HiAgent}{Yes} & 24.08 \\
  & AWM~\cite{awm2024} & CR & \checkmark &  &  & Single & Academia & \href{https://github.com/zorazrw/agent-workflow-memory}{Yes} & 24.09 \\
  & MemInsight~\cite{meminsight2025} & CR & \checkmark &  &  & Single & Industry & No & 25.03 \\
  
\cline{1-10}

\multirow{5}{*}{\textit{\textbf{\makecell{Rule-based \\ Selection \& Retreival}}}}
  & COOPER~\cite{cooper2023} & RR & \checkmark &  &  & Multi & Academia & \href{https://github.com/YiCheng98/Cooper}{Yes} & 23.12 \\
  & Memary~\cite{memary2024} & RR &  & \checkmark &  & Single & Academia & \href{https://github.com/kingjulio8238/Memary}{Yes} & 24.10 \\
  & ZEP~\cite{zep2025} & RR &  & \checkmark &  & Single & Industry & \href{https://github.com/getzep/zep}{Yes} & 25.01 \\
  & COLA~\cite{cola2025} & RR &  & \checkmark &  & Multi & Academia & \href{https://github.com/Alokia/COLA-demo}{Yes} & 25.03 \\
  & Mem0~\cite{mem0_2025} & RR &  & \checkmark &  & Single & Industry & \href{https://github.com/mem0ai/mem0}{Yes} & 25.04 \\
\midrule

\rowcolor{black!5}
\multicolumn{10}{c}{\textit{\textbf{Hybrid Paradigm}}} \\ \midrule

\multirow{4}{*}{\textit{\textbf{\makecell{Slot-based Memory}}}}
  & A-Mem~\cite{amem2025} & CR & \checkmark &  &  & Single & Academia & \href{https://github.com/agiresearch/A-mem}{Yes} & 25.02 \\
  & G-Memory~\cite{gmemory2025} & CR & \checkmark &  &  & Single & Academia & \href{https://github.com/bingreeky/GMemory}{Yes} & 25.06 \\
  & Intrinsic Memory \allowbreak Agents~\cite{intrinsicmemoryagents2025} & CR & \checkmark &  &  & Multi & Academia & No & 25.08 \\
\cline{1-10}

\multirow{2}{*}{\textit{\textbf{\makecell{Knowledge Synthesis}}}}
  & Optimus-1~\cite{optimus1_2024} & CR &  &  & \checkmark & Single & Academia & \href{https://github.com/JiuTian-VL/Optimus-1}{Yes} & 24.08 \\
  & Nemori~\cite{nemori_2025} & CR & \checkmark &  &  & Single & Academia & No & 25.08 \\
\cline{1-10}

\multirow{7}{*}{\textit{\textbf{\makecell{Modular \\ Optimization}}}} 
  & RAP~\cite{rap_2024} & RR &  & \checkmark &  & Single & Industry & \href{https://github.com/PanasonicConnect/rap}{Yes} & 24.02 \\
  & EMU~\cite{na2024emu} & RR &  &  & \checkmark & Multi & Academia & \href{https://github.com/HyunghoNa/EMU}{Yes} & 24.03 \\
  & SeCom~\cite{secom2025} & RR &  & \checkmark &  & Single & Industry & No & 25.02 \\
  & M+~\cite{mplus_2025} & CR & \checkmark &  &  & Single & Academia & No & 25.02 \\
  & TWM~\cite{wang2025twm} & CR & \checkmark &  &  & Single & Academia & \href{https//github.com/xid32/NAACL_2025_TWM}{Yes} & 25.02 \\
  & RCR-Router~\cite{rcrrouter2025} & RR &  & \checkmark &  & Multi & Academia & No & 25.08 \\
  & Learn-to-\allowbreak Memorize ~\cite{learn2memorize2025} & CR & \checkmark &  &  & Single  & Academia/\allowbreak Industry & \href{https://github.com/nuster1128/learn_to_memorize}{Yes} & 25.08 \\
\cline{1-10}

\multirow{5}{*}{\textit{\textbf{\makecell{Workflow \\Internalization}}}}
  & Self-RAG~\cite{selfrag_2023} & RR &  & \checkmark &  & Single & Academia & \href{https://selfrag.github.io}{Yes} & 23.10 \\
  & TiM~\cite{tim2023} & CR &  &  & \checkmark & Single & Industry & No & 23.11 \\
  & HippoRAG~\cite{hipporag_2024} & RR &  & \checkmark &  & Single & Academia & \href{https://github.com/OSU-NLP-Group/HippoRAG}{Yes} & 24.05 \\
  & MemAgent~\cite{memagent_2025} & CR &  &  & \checkmark & Single & Industry & \href{https://github.com/BytedTsinghua-SIA/MemAgent}{Yes} & 25.07 \\
  & Memory-R1~\cite{memoryr1_2025} & CR &  &  & \checkmark & Single & Academia & No & 25.08 \\
\midrule

\rowcolor{black!5}
\multicolumn{10}{c}{\textit{\textbf{Model-native Paradigm}}} \\ \midrule
\multirow{2}{*}{\textit{\textbf{\makecell{Internal \\State Management}}}}
  & Scissorhands~\cite{scissorhands_2023} & CR & \checkmark &  &  & Single & Academia & No & 23.05 \\
  & H2O~\cite{h2o_2023} & CR & \checkmark &  &  & Single & Academia & \href{https://github.com/FMInference/H2O}{Yes} & 23.06 \\
\cline{1-10}

\multirow{2}{*}{\textit{\textbf{\makecell{Hybrid \\Parameterization}}}}
  & MEM1~\cite{mem1_2025} & CR &  &  & \checkmark & Single & Academia & \href{https://github.com/MIT-MI/MEM1}{Yes} & 25.06 \\
  & H-MEM~\cite{hmem_2025} & CR & \checkmark & & & Single & Academia & No & 25.07 \\
\cline{1-10}

\multirow{2}{*}{\textit{\textbf{\makecell{Unified \\Policy Learning}}}}
  & CARE~\cite{care_2025} & RR &  & \checkmark &  & Single & Academia & \href{https://github.com/FoundationAgents/CARE}{Yes} & 25.09 \\
  & MemAct~\cite{zhang2025memact} & RR &  &  & \checkmark & Single & Academia & No & 25.10 \\
\midrule
\end{longtable}
}

\subsubsection{\textbf{Hybrid Paradigm}} \hspace{2mm} 
The hybrid paradigm enhances adaptability by replacing fixed rules with learned policies while retaining a modular structure. This approach represents a transitional stage, bridging rule-based pipelines and fully internalized models.

\paragraph{\textbf{Context Re-structuring}} \hspace{2mm}
In the hybrid paradigm, context re-structuring evolves from simple information arrangement to dynamic knowledge construction~\cite{awm2024}. A representative technique is the use of learned memory slots, where a lightweight policy network learns to distill and write key information into a structured, finite set of slots rather than passively receiving raw text~\cite{gmemory2025,amem2025,intrinsicmemoryagents2025}. Another approach uses a fine-tuned model to create associations between disparate pieces of knowledge, effectively synthesizing new, structured evidence. Systems such as EMU, Optimus-1, and Nemori illustrate this synthesis-oriented restructuring~\cite{optimus1_2024,nemori_2025}, which has been shown to significantly improve reasoning performance~\cite{amem2025,mem0_2025}.

\paragraph{\textbf{Retrieval and Routing}} \hspace{2mm}
Retrieval and routing in the hybrid paradigm aim to replace fixed rules with learned policies to enhance adaptability. The evolution proceeds mainly along two paths. 
The first path is modular optimization, which retains the classic pipeline structure but replaces individual rule-based components (e.g., a retriever or re-ranker) with trainable neural modules. Retrieval-Augmented Planning(RAP) learns to pull and reuse relevant past trajectories instead of following hand-written heuristics~\cite{rap_2024}, SeCom segments and compresses dialogue into structured memory units and retrieves at the segment level, which reduces noise compared to naive last-k recall in long conversations ~\cite{secom2025}. Temporal Working Memory(TWM) uses query-guided retention over long multimodal streams, keeping only temporally relevant segments rather than forwarding the entire recent history ~\cite{wang2025twm}. RCR-Router learns routing policies that deliver different memory to different agent or roles instead of broadcasting the same context to everyone ~\cite{rcrrouter2025}. Learn-to-Memorize directly optimizes what to store and what to recall as part of the task objective rather than treating storage and retrieval as fixed utilities~\cite{learn2memorize2025}. EMU shows the same tendency in multi-agent reinforcement learning by learning which episodic memory are worth recalling to coordinate future behavior ~\cite{na2024emu}. These works improve performance on specific tasks while maintaining the overall architecture's interpretability.

The second path is workflow internalization, towards a deeper integration. The LLM is trained to execute the steps of a predefined workflow, learning to decide when to retrieve, rewrite, or evaluate information. Self-RAG, for instance, introduced reflection tokens that allow the model to trigger retrieval and self-critique its generated content based on the results~\cite{selfrag_2023}.  HippoRAG integrates retrieval with a structured memory index and incremental consolidation, allowing new information to be linked to existing stored knowledge during generation rather than only before generation ~\cite{hipporag_2024}. MemAgent treats long context as an active working memory management problem that the model ingests long inputs in stages, refreshes what it keeps, and continues reasoning without relying only on a fixed sliding window \cite{memagent_2025}. Related approaches like TiM and Memory-R1 similarly entwine selection and synthesis within the model's decision-making process, improving adaptability while preserving auditability through explicit step definitions~\cite{tim2023memoryr1_2025}.

\subsubsection{\textbf{Model-native Paradigm}} \hspace{2mm}
The fully model-native paradigm eliminates external scaffolding, internalizing the entire memory management process into a single, unified policy learned end-to-end.

\paragraph{\textbf{Context Re-structuring}} \hspace{2mm}
In this paradigm, context re-structuring is deeply integrated with the model's internal state management. This is often achieved by modifying the attention mechanism itself, for example, through a learnable KV-cache policy that proactively decides which historical key-value pairs to retain, discard, or compress. Systems like Scissorhands and H2O exemplify this dynamic cache management~\cite{scissorhands_2023,h2o_2023}. Another direction couples model parameterization with external memory, allowing memory operations to be tied to model weights while remaining grounded in an auditable store, as seen in MEM1 and H-MEM~\cite{mem1_2025,hmem_2025}. In all cases, restructuring moves from an external pre-processing step to an intrinsic, policy-driven behavior during inference.

\paragraph{\textbf{Retrieval and Routing}} \hspace{2mm}
The highest level of internalization merges the entire memory process into a unified policy space, where the model learns not only how to reason but also how and when to access and manage its memory. Very recently, MemAct~\cite{zhang2025memact} proposes to reframe context management as a learnable, intrinsic capability. In MemAct, the agent actively curates its own working memory by executing explicit editing operations as part of its unified decision-making process, trained via a novel DCPO learning algorithm to balance memory curation with long-term task objectives. This approach contrasts with methods like CARE, which focus on co-adapting retrieval and reasoning without giving the agent explicit control over its memory state~\cite{care_2025}. In the fully native paradigm, the retriever and generator become deeply coupled, often sharing parameters and enabling end-to-end optimization. External knowledge repositories remain valuable as a factual foundation for compliance, but the model itself assumes full autonomy over their use, marking the final transition from an assembly of external modules to a system with internally coherent capabilities.

\subsection{Long-term Memory}
\label{sec:long_term_memory}
Long-term memory refers to persistent knowledge and experience maintained across sessions or tasks. Its contents can be held in two primary types of carriers: external repositories, such as documents, vector stores, or graphs, and the model's internal parameters, including adapter modules or the weights of the model itself~\cite{memgpt2024,zheng2023synapse,park2023generative,edge2024local,wu2022memorizing,wang2021kadapter,sheng2023slora,du2025mom}.

Key properties of long-term memory include low write frequency and strong traceability, which support compliance and versioning. The content is typically not editable online during a task; instead, any online optimization focuses on retrieval and utilization strategies. This paradigm provides a factual basis for an agent's planning and tool use but can face challenges with cross-source consistency and information timeliness. Storing knowledge directly in model parameters can yield lower latency and deeper integration, but it often comes at the cost of reduced interpretability and an increased risk of interference between stored facts~\cite{zhao2025lmlm}. An overview of methods for long-term memory is summarized in Table~\ref{tab:ltm_methods}.

\subsubsection{\textbf{External Repository as Memory Carrier}} \hspace{2mm}
\paragraph{\textbf{Retrieval-Augmented Generation.}} \hspace{2mm}
When using external repositories, the emphasis shifts from how to store information to how to retrieve it precisely and use it reliably~\cite{izacard2022atlas,borgeaud2021retro}. A common practice is to build hybrid indexes offline by combining inverted indices, dense vectors, and metadata. During inference, multi-path retrieval and re-ranking are used to select and compress candidate evidence. This process is often accompanied by deduplication and consistency checks to ensure the injected evidence is both relevant and traceable~\cite{memgpt2024,zheng2023synapse}. While this approach is robust for tasks requiring deep knowledge from large corpora, injecting excessive or poorly selected information can dilute the model's attention. Consequently, recent methods have moved from single-path semantic retrieval toward more complex pipelines that incorporate learned re-ranking and delegate decisions about when and what to retrieve to learned policies~\cite{edge2024local}.

\paragraph{\textbf{Structure and Compressed Summarization.}} \hspace{2mm}
Because long-term knowledge sources are often lengthy and heterogeneous, simply expanding the context window is not always effective. Layered summarization and structuring are therefore used to manage information density. Documents can be segmented into structured evidence cards with explicit citations and timestamps for verifiability. This structured layer reduces noise and stabilizes the candidate pool for retrieval systems~\cite{park2023generative,memgpt2024}. The focus has shifted from generating human-readable text summaries to creating machine-usable evidence objects that integrate tightly with retrieval and routing strategies~\cite{edge2024local,izacard2022atlas}.

\begin{table}[t]
\centering
\footnotesize
\setlength{\tabcolsep}{4pt}
\renewcommand{\arraystretch}{1.25}
\caption{Overview of long-term memory methods}\label{tab:ltm_methods}

\begin{tabular}{>{\centering\arraybackslash}p{1.6cm}|%
                >{\raggedright\arraybackslash}p{2.2cm}|%
                >{\centering\arraybackslash}p{1.4cm}|%
                >{\centering\arraybackslash}p{2.0cm}|%
                >{\centering\arraybackslash}p{2.0cm}|%
                >{\centering\arraybackslash}p{1.2cm}|%
                >{\centering\arraybackslash}p{0.8cm}|%
                >{\centering\arraybackslash}p{1.0cm}}
\midrule
\rowcolor{blue!5}
\textbf{Carrier} &
\textbf{Method} &
\makecell{\textbf{Content}\\ \textbf{Structure}} &
\makecell{\textbf{Update}\\ \textbf{Mechanism}} &
\makecell{\textbf{Retrieval}\\ \textbf{Mechanism}} &
\textbf{Affiliation} &
\textbf{Access}&
\textbf{Date} \\
\midrule

\multirow{11}{*}{\makecell[c]{\textit{\textbf{External}}\\ \textit{\textbf{Repository}}}}
& RETRO~\cite{borgeaud2021retro} & Text+Vector & Static & Attention-based & Industry & No & 21.12 \\
\cline{2-8}
& Atlas~\cite{izacard2022atlas} & Text  & Static & Embedding-based & Industry & No & 22.08 \\
\cline{2-8}
& Generative Agents~\cite{park2023generative} & Structured \allowbreak Text & Rule-based\allowbreak (Summarization) & Scoring & Academia & No & 23.04 \\
\cline{2-8}
& Synapse~\cite{zheng2023synapse} & Trajectory Exemplars & Offline (curation) &Embedding-based & Academic & No & 23.06 \\
\cline{2-8}
& MemGPT~\cite{memgpt2024} & Text + Vector & Rule-based\allowbreak (Summarization) & Embedding \allowbreak + Scoring & Academia & \href{https://github.com/letta-ai/letta}{Yes} & 23.10 \\
\cline{2-8}
& GraphRAG~\cite{edge2024local} & \makecell[t]{Knowledge \\ Graph} & Rule-based\allowbreak (Summarization) & Graph-based & Industry & \href{https://github.com/microsoft/graphrag}{Yes} & 24.04 \\
\cline{2-8}
& LMLM~\cite{zhao2025lmlm} & Text  & Static & Instruction-based & Academia & No & 25.05 \\
\midrule

\multirow{11}{*}{\makecell[c]{\textit{\textbf{Model}}\\ \textit{\textbf{Parameters}}}}
& K-Adapter~\cite{wang2021kadapter} & Adapters /\allowbreak LoRA Bank & Offline (Training) & Rule-based\allowbreak + Router & Academia & \href{https://github.com/microsoft/K-Adapter}{Yes} & 21.08 \\
\cline{2-8}
& Memorizing Transformers~\cite{wu2022memorizing} & kNN Cache+\allowbreak KV & Online (KV writes) & Embedding-based & Academia & \href{https://github.com/lucidrains/memorizing-transformers-pytorch}{Yes} & 22.03 \\
\cline{2-8}
& S-LoRA~\cite{sheng2023slora} & Adapters /\allowbreak LoRA Bank & Offline (Training) & Rule-based \allowbreak + Router & Academia & No & 23.11 \\
\cline{2-8}
& MixLoRA~\cite{mixlora_blog} & Adapters /\allowbreak LoRA Bank & Offline (Joint \allowbreak Fine-Tuning) & Router (top k) & Academia & \href{https://github.com/PLUM-Lab/MixLoRA}{Yes} & 24.04 \\
\cline{2-8}
& ELDER~\cite{li2024elder} & Adapters / LoRA Bank & Offline (Automatic \allowbreak Training) & Rule-based \allowbreak + Router & Academia & No & 24.08 \\
\cline{2-8}
& MoM~\cite{du2025mom} & Linear State Layers & Offline (Automatic \allowbreak Training) & Router & Academia & \href{https://github.com/OpenSparseLLMs/MoM}{Yes} & 25.02 \\

\midrule

\end{tabular}
\end{table}

\subsubsection{\textbf{Model Parameters as Memory Carrier}} \hspace{2mm}
\paragraph{\textbf{Global Parameter Internalization.}} \hspace{2mm}
For knowledge that is stable and frequently reused, writing it directly into the model's parameters can reduce inference latency and dependence on external stores. In practice, Continual pretraining and instruction distillation have been used to inject domain-specific knowledge, API schemas, or reasoning patterns into the base model through training data, as in systems that jointly train a generator and retriever for knowledge-intensive QA (Atlas) or couple parametric knowledge with retrieval supervision in large memory language models (LMLM) ~\cite{zhao2025lmlm,izacard2022atlas}. RETRO further shows that some knowledge can remain external but still be treated as internal by conditioning generation on retrieved evidence during training and inference, which lowers the need to scale parameters aggressively ~\cite{borgeaud2021retro}. Data replay and regularization techniques are used to mitigate catastrophic forgetting. This allows the internalized knowledge to serve as a fallback when external retrieval fails or returns incomplete information.

\paragraph{\textbf{Targeted Parameter Intervention.}} \hspace{2mm}
When a small number of critical facts need to be corrected or updated, model editing offers a more targeted approach. Representative methods apply low-rank adjustments to specific layers or intermediate representations, allowing for the injection or replacement of target knowledge with minimal disruption to the model's overall capabilities~\cite{meng2022rome,meng2022memit,mitchell2022serac}. This method is effective for sparse, important factual corrections but is not suitable for large-scale knowledge migration. In practice, this has evolved from single edits to batch pipelines that can be triggered by upstream changes in an authoritative source, enabling near real-time corrections~\cite{mitchell2022serac}.

\paragraph{\textbf{Lightweight Parameter Injection.}} \hspace{2mm}
To balance the high cost of global internalization with the limited capacity of model editing, lightweight modules can be used. This strategy involves freezing the base model and training or attaching small, specialized components that carry specific knowledge. Adapter style methods can encode user preferences or task-specific skills in compact modules, for example K-Adapter for factual andlinguistic knowledge~\cite{wang2021kadapter}, or S-LoRA for User preference and skill personalization~\cite{sheng2023slora}.  Recent variants organize many such modules or linear states as banks or mixtures and select them on a per-input basis~\cite{li2024elder,mixlora_blog,du2025mom}. These approaches are computationally efficient and allow for per-task or per-domain customization.

\subsection{Summary and Discussion}
The evolution of agent memory also follows the paradigm shift from externally orchestrated modules toward an intrinsic, unified capability. This trend reframes memory from a simple information storage mechanism into a system for action-oriented evidence governance, responsible for preserving state, retrieving information, and injecting it into the agent's reasoning process. This shift from pipeline-based to model-native solutions is particularly evident in short-term memory.

Short-term memory was initially developed to compensate for models' limited context processing abilities. Early pipeline-based approaches, such as MemoChat~\cite{memochat2023}, used structured templates and decomposable modules to filter and arrange evidence before it reached the model. While interpretable and easy to implement, these systems were limited by cross-module error accumulation and inflexibility, casting the model as a passive participant. As native context windows have expanded, the focus has shifted from mere capacity to effective context management. The challenge is no longer how much information can be processed, but how well the model can filter noise, locate critical evidence, and utilize it for complex reasoning. This has driven the evolution from coarse-grained pipelines to more intelligent, learned policies.

This transition is not an abrupt replacement but a gradual process of internalizing validated pipeline functions into the model’s native abilities. A key bottleneck in this path lies in long-sequence data synthesis and curriculum design. Future breakthroughs will likely involve training models to endogenously learn operations like retrieval, compression, and verification as explicit objectives~\cite{borgeaud2021retro, izacard2022atlas, zhao2025lmlm}. As models learn attention mechanisms akin to a retriever and resource allocation strategies similar to a router~\cite{rcrrouter2025}, external modules like vector stores will recede from online decision-making. Their role will shift to serving as backend infrastructure for compliance, persistent storage, and offline content preparation, supporting the model's native governance capabilities.

Long-term memory began primarily as external repositories, with engineering practices emphasizing traceability, control, and governance. Retrieval relied on cascaded re-ranking~\cite{densepassageretrieval20}, but this design faced two key limitations: the content was not rewritable online, and there was often a mismatch between retrieved relevance and actual utility in the reasoning chain~\cite{selfrag_2023}. A key recent trend is personalization, where long-term memory evolves from a unified knowledge warehouse into a dynamic, user-specific asset composed of traceable evidence cards~\cite{park2023generative, memgpt2024}. A second trend is the deepening coupling with short-term memory. Although most long-term stores are not yet fully model-native due to their offline nature, methods are emerging that blur this boundary. This includes co-training the retriever and generator to improve evidence utilization~\cite{izacard2022atlas, R4}, using online updates to write back critical facts, and employing lightweight adapters to inject knowledge into model parameters~\cite{wang2021kadapter}. Systems like Mem0~\cite{mem0_2025} sit at this intersection, combining the governability of external memory with the fast read-write capabilities needed for session-level adaptation, pointing toward an eventual unification of long-term and short-term memory systems.

In summary, several key trends are shaping the future of memory-enhanced agents:
\begin{itemize}
\item \textbf{Benchmark evolution}: evaluation will move beyond raw capacity toward stringent reasoning benchmarks that stress-test long-range dependency, interference robustness, and logical closure.
\item \textbf{Capability internalization and unification}: the internalization of capabilities will continue, with the progressive unification of short-term and long-term memory into model-native skills and the default co-training of retrieval and generation components. 
\item \textbf{Personalization with stronger governance}: long-term memory will become highly personalized and dynamic, increasing need for stronger governance, privacy controls, and user-facing correction to ensure consistency and reliability.
\end{itemize}

\section{Applications}\label{sec:6}
\subsection{Deep Research Agent}
\subsubsection{\textbf{Overview}}
\hspace{2mm}
 Recent progress in search technologies has extended model capabilities beyond their internal parameterized knowledge, which is often outdated and unreliable. 
The RAG paradigm~\cite{lewis2020retrieval} enables direct access to external knowledge bases, providing accurate and up-to-date information. 
Building upon the RAG framework, Deep Research agents employ adaptive search strategies that control search iterations and step depth to retrieve relevant evidence and generate definitive answers or generative synthesis. 
Depending on whether the search mechanism is embedded within model parameters, the paradigms of Deep Research agents can be categorized into two complementary yet closely related forms: the pipeline-based and the model-native approaches.

The pipeline-based paradigm provides controllable and interpretable workflows, supporting debugging and updates at any stage, which makes it well-suited for industrial deployment.
However, it depends on predefined, prompt-driven procedures that demand substantial manual design.
The performance of such systems is constrained by the model’s instruction-following capability, long-horizon reasoning stability, and sensitivity to prompt formulation, making them less reliable for complex reasoning tasks.

In contrast, the core idea of model-native methods is to internalize search strategies into model parameters through reinforcement learning, enabling end-to-end autonomous research capability.
This paradigm enhances long-horizon reasoning consistency and allows the model to dynamically adapt to diverse search contexts.
From the perspective of reasoning consistency, internalizing strategies at the parametric level leverages stored knowledge to preserve goal and evidence coherence across multiple turns~\cite{song2025r1plusplus}, thereby improving reasoning stability and reducing phenomena such as goal drift, lost-in-conversation, and information forgetting in extended prompt chains~\cite{laban2025llms,liu2023lost,wang2025illusion}.
With respect to adaptability, the model dynamically adjusts retrieval depth and evidence integration according to task complexity, without relying on predefined workflows, thereby reducing unnecessary tool calls, redundant information retrieval, and overall token costs~\cite{jin2025searchr1trainingllmsreason}.
By shifting from external orchestration to internalized capability, this paradigm ensures stable reasoning across diverse environments and task structures, enhancing the generalization of Deep Research agents to open-domain, multi-hop, and unstructured evidence tasks.

The pipeline-based and model-native approaches differ fundamentally in implementation but converge in their evolution toward more autonomous, efficient, and adaptive research processes, exhibiting consistent trends across key dimensions such as information acquisition, task objectives, base model design, and agent architecture.
The remainder of this paper is organized as follows: Section~\ref{sec:Pipeline-based Paradigm} introduces the pipeline-based paradigm; Section~\ref{sec:Model-native Paradigm} presents the model-native paradigm; and Section~\ref{sec:dr_discussion} compares their evolutionary patterns and discusses future directions of Deep Research agents.

\begin{figure}[t] 
  \centering 
  \includegraphics[width=0.98\textwidth]{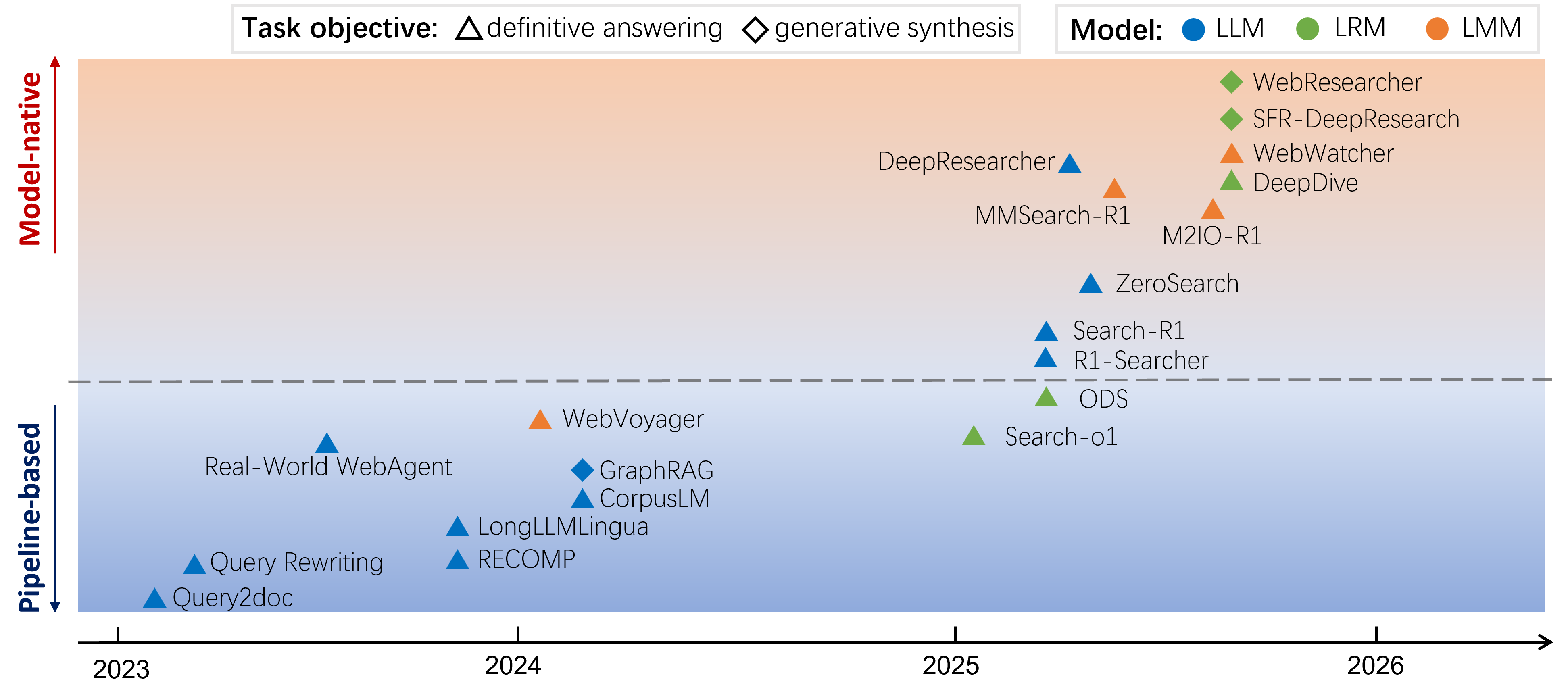} 
  \caption{Overview of deep research agent methods.} 
  \label{fig:DR_Agent} 
\end{figure}

\subsubsection{\textbf{Pipeline-based Paradigm} \label{sec:Pipeline-based Paradigm}}
\hspace{2mm}
Pipeline-based methods enable retrieval-augmented knowledge integration and multi-turn reasoning through carefully designed prompt templates and orchestrated tool chains.
Two main implementation paths exist:
(1) single-turn retrieval and synthesis, exemplified by RAG~\cite{lewis2020retrieval}, which optimizes the ``retrieve–integrate–generate'' pipeline; and
(2) multi-turn reasoning–action cycles, exemplified by ReAct~\cite{yao2023react}, which integrates retrieval into iterative reasoning, allowing models to decide when and what to retrieve through a ``think–retrieve–rethink'' process.

\paragraph{\textbf{Single-turn Retrieval and Synthesis}}
\hspace{2mm}
Single-turn retrieval and synthesis methods perform one-shot retrieval and generation, introducing external knowledge into LLM reasoning to address the limitations of static, parameterized memory.
These methods typically follow a two-stage process: query enhancement and evidence processing.
During the first phase, the model or its external components construct or rewrite task-specific queries to retrieve candidate evidence from search engines, vector databases, or knowledge bases.
Subsequently, the retrieved content is filtered, compressed, and structurally organized to preserve essential and relevant context, which is then injected into the model’s context for answer synthesis or report generation~\cite{jung2024familiarity}.
The two stages are elaborated in detail below.
\begin{itemize}
    \item \textbf{Query enhancement.}\quad \setlength{\parindent}{1em}
    Query enhancement improves retrieval effectiveness by rewriting, expanding, or constructing task-specific queries~\cite{wang2023generative}, thereby increasing retrieval quality and answerability.
    Query2doc~\cite{wang2023query2doc} prompts an LLM in a few-shot manner to generate pseudo-documents, which are concatenated with the original query for search, effectively expanding query scope.
    \citet{ma2023query} introduce a query rewriting framework for RAG, where an LLM or a lightweight model first rewrites the query, retrieves results via the Bing API, and then passes the retrieved content to a reader model for final generation.
    FreshLLMs~\cite{vu2023freshllms} focus on time-sensitive knowledge by combining multiple evidence sources with structured metadata, and guide attention according to temporal order.

    \item \textbf{Evidence processing.}\quad \setlength{\parindent}{1em}
    To ensure that retrieved information is relevant, concise, and structurally coherent before being injected into the model context, evidence processing refines and organizes raw retrieval outputs. \citet{xu2023recomp} propose RECOMP, which introduces a compression layer prior to injection, combining extractive and abstractive compressors that can selectively omit uninformative content. \citet{jiang2023longllmlingua} develop LongLLMLingua, applying global reranking and span-level extraction to mitigate context explosion. \citet{vu2023freshllms} normalize search results into structured evidence lists for more consistent integration. \citet{li2024corpuslm} design CorpusLM, which treats a unified corpus as fixed external memory, retrieving fragments online and directly reading from a clean repository during inference. 
    \citet{edge2024local} proposed GraphRAG that replace scattered documents with knowledge graphs and community summaries, supporting path-level interpretability and global evidence aggregation.\citet{jin2024bider} present \textit{BIDER}, which refines retrieval documents into Key Supporting Evidence (KSE) through knowledge synthesis, supervised fine-tuning, and preference alignment, thereby improving LLM's answer quality while reducing input length.
\end{itemize}

\paragraph{\textbf{Multi-turn Reasoning-Action} }
\hspace{2mm}
To overcome the limitations of single turn retrieval and synthesis, including insufficient evidence, hallucinations, and error propagation, multi turn reasoning and action methods introduce iterative retrieval within a closed reasoning loop.
\citet{yao2023react} propose ReAct, a representative framework that interleaves reasoning and acting, forming a feedback process where reasoning guides actions and observations refine subsequent reasoning.
This paradigm enhances reasoning reliability through two key mechanisms: the orchestration of reasoning and retrieval, which enables iterative evidence acquisition, and context management with evidence refinement, which maintains focus and reduces noise accumulation.
Building on this foundation, recent studies have advanced these two mechanisms through various designs and optimization strategies, as detailed below.

\begin{itemize}
    \item \textbf{Orchestration of reasoning and retrieval.}\quad \setlength{\parindent}{1em}
    This line of work focuses on coordinating reasoning and retrieval to iteratively accumulate relevant evidence. \citet{he2024webvoyager}, \citet{gur2023real}, and \citet{alzubi2025open} develop agents that solve tasks step by step, repeatedly invoking search tools and feeding textual observations or web snapshots back into the reasoning process. \citet{li2025search} propose \textit{Search o1}, which employs a large reasoning model without handcrafted templates and introduces a \textit{Reason in Documents} mechanism that orchestrates pause, retrieve, and resume operations for retrieval augmented answering.

    \item \textbf{Context management and evidence refinement.}\quad \setlength{\parindent}{1em}
    Managing large amounts of retrieved information is essential to preserve reasoning coherence and reduce redundancy.  
    \citet{he2024webvoyager} prune historical context, retaining only the three most recent observations and the complete reasoning trace to maintain recency and relevance.  
    \citet{gur2023real} train HTML T5 to handle long HTML documents using hybrid local and global attention with long span denoising pretraining, extending input length and context window during fine tuning. 
    \citet{li2025search} design an in document reasoning and refinement module that compresses retrieved content and reinjects only the most relevant knowledge.  
    \citet{alzubi2025open} adopt chunking and reranking, segmenting web pages into paragraph level chunks and keeping only those above a relevance threshold.  
    \citet{wu2025resum} mitigate information overload through periodic context summarization, which removes redundant content while preserving key evidence, enabling continuous exploration.
\end{itemize}

\subsubsection{\textbf{Model-native Paradigm} \label{sec:Model-native Paradigm}}
\hspace{2mm}
The model-native paradigm redefines how Deep Research agents perform reasoning and tool use.
Instead of relying on externally orchestrated workflows such as step-by-step prompting in ReAct, model-native agents internalize planning and retrieval within their parameters. This integration allows end-to-end, proactive, and dynamically adaptive information seeking, where the model itself acts as both planner and executor.
Depending on the interaction environment during training, model-native agents can be developed through offline or online training.

\paragraph{\textbf{Offline Training} }
\hspace{2mm}
Offline training conducts external information retrieval within closed, locally hosted databases that operate without internet access.
This setup provides a stable and controllable training environment, allowing the model to learn retrieval and reasoning behaviors under low noise and fixed data conditions.
Representative works such as Search-R1~\cite{jin2025searchr1trainingllmsreason}, R1-Searcher~\cite{song2025r1}, ReSearch~\cite{chen2025researchlearningreasonsearch}, R-Search~\cite{zhao2025r}, and R1-Searcher++~\cite{song2025r1plusplus} adopt Wikipedia as their static retrieval corpus, enabling supervised or reinforcement learning over consistent textual contexts.
Similarly, M2IO-R1~\cite{xiao2025m2io} employs a pre-constructed dataset (M2IO-Inserter) to simulate information insertion and retrieval tasks in a controlled manner.
Such designs ensure reproducibility and avoid issues commonly encountered in online environments, including network instability, API rate limitations, and unpredictable web responses.

However, the static and bounded nature of local datasets constrains the scope of knowledge and limits exposure to real-time information. As a result, models trained exclusively offline may struggle with information obsolescence and reduced generalization in open environments.
To overcome these limitations, ZeroSearch~\cite{sun2025zerosearchincentivizesearchcapability} introduces a simulated search engine that uses an LLM-generated document space to emulate online retrieval. This approach maintains controllable training dynamics while approximating the variability of real web search, achieving a balance between data freshness, quality, and stability.

\paragraph{\textbf{Online Training} }
\hspace{2mm}
Online training situates model learning within real-world, connected environments, allowing agents to retrieve and reason over dynamic and continuously updated information sources.
This paradigm provides direct access to the open web, enabling models to acquire real-time knowledge and effectively extend their informational coverage beyond static corpora.
DeepResearcher~\cite{zheng2025deepresearcherscalingdeepresearch} demonstrates large-scale reinforcement learning in live web environments, where agents interact directly with search engines and webpages during training.
Following this direction, WebThinker~\cite{li2025webthinker}, WebWatcher~\cite{geng2025webwatcher}, SFR-DeepResearch~\cite{nguyen2025sfr}, MMSearch-R1~\cite{wu2025mmsearchr1}, and DeepDive~\cite{lu2025deepdive} perform end-to-end reinforcement learning under realistic web settings, where model actions directly determine query formulation, browsing behaviors, and document reasoning.

Although online training captures the complexity of open-world information retrieval, it also introduces significant engineering and optimization challenges.
Factors such as network fluctuations, page latency, API quotas, anti-crawling mechanisms, and advertisement noise can disrupt the learning signal and escalate computational costs.
Building robust and scalable online training pipelines therefore requires careful system design, adaptive control strategies, and efficient reward shaping to ensure stability and performance in dynamic environments.

\afterpage{ 
\clearpage  
\footnotesize

\begin{longtable}{C{1.5cm}|l|C{1.2cm}|c|C{1.2cm}|C{1.0cm}|C{1.2cm}|c|C{0.8cm}} 
\caption{An overview of deep research agent methods, with their task objective (\textit{Def.} for definitive answering and \textit{Gen.} for generative synthesis), base model, information acquisition method (\textit{Info.}), agent architecture (\textit{Arch.}) and among other attributes.}\label{tab:deep research}\\ 

\toprule
\rowcolor{blue!5}
\textbf{} & \textbf{Method} & \textbf{Objective} & \textbf{Model} & \textbf{Info.} & \textbf{\makecell{Arch.}} & \textbf{Affiliation} & \textbf{Access} & \textbf{Date}\\
\midrule
\endfirsthead
\multicolumn{9}{c}{\tablename~\thetable\ -- {Overview of deep research agent methods (continued).}}\\
\toprule
\rowcolor{blue!5}
\textbf{} & \textbf{Method} & \textbf{Task Obj.} & \textbf{Model} & \textbf{Info. Acq.} & \textbf{\makecell{Agent\\Para.}} & \textbf{Affi.} & \textbf{Access} & \textbf{Date}\\
\midrule
\endhead

\midrule
\multicolumn{9}{r}{\textit{continued on next page}}\\
\endfoot

\bottomrule
\endlastfoot

\rowcolor{black!5}\multicolumn{9}{c}{\textit{\textbf{Pipeline-based Paradigm}}}\\
\midrule

\multirow{9}{*}{\makecell[c]{\textit{\textbf{Single-turn}}\\ \textit{\textbf{Retrieval\&}}\\  \textit{\textbf{Synthesis}}}}
& Query2doc ~\cite{wang2023query2doc}     & Def. & LLM & API    & Single & Industry & \href{https://huggingface.co/datasets/intfloat/query2doc_msmarco}{Yes} & 23.03 \\
& Query Rewriting ~\cite{ma2023query}           & Def.    & LLM & API & Single & Academia & \href{https://github.com/xbmxb/RAG-query-rewriting}{Yes}     & 23.05 \\
& RECOMP ~\cite{xu2023recomp}            & Def.    & LLM & API     & Single & Academia & \href{https://github.com/carriex/recomp}{Yes} & 23.10 \\
& LongLLMLingua ~\cite{jiang2023longllmlingua}& Def.    & LLM & API     & Single & Industry & \href{https://github.com/microsoft/LLMLingua}{Yes} & 23.10 \\
& FreshLLMs ~\cite{vu2023freshllms}  & Def.    & LLM & API    & Single & Academia & \href{https://github.com/freshllms/freshqa}{Yes} & 23.10 \\
& BIDER ~\cite{jin2024bider}          & Def. & LLM & API     & Single  & Academia & {No} & 24.02 \\
& CorpusLM ~\cite{li2024corpuslm} & Def.   & LLM & API    & Single & Academia & {No} & 24.02 \\
& GraphRAG ~\cite{edge2024local}          & Gen.   & LLM & API    & Single & Industry & \href{https://github.com/microsoft/graphrag}{Yes} & 24.04 \\
& RetroLLM ~\cite{li2024retrollm}      & Def.   & LLM & API     & Single & Academia & \href{https://github.com/sunnynexus/RetroLLM}{Yes} & 24.12 \\
\midrule

\multirow{5}{*}{\makecell[c]{\textit{\textbf{Multi-turn}}\\ \textit{\textbf{Reasoning-}}\\ \textit{\textbf{Action}}}}
& Real-World WebAgent ~\cite{gur2023real} & Def. & LLM & Browser & Multi & Academia & {No} & 23.07 \\
& WebVoyager ~\cite{he2024webvoyager} & Def. & LMM & Browser & Single & Academia & \href{https://github.com/MinorJerry/WebVoyager}{Yes} & 24.01 \\
& Search-o1 ~\cite{li2025search}& Def. & LRM & API  & Single  & Academia & \href{https://github.com/RUC-NLPIR/Search-o1}{Yes} & 25.01 \\
& ODS ~\cite{alzubi2025open} & Def. & LRM & API  & Single  & Academia & \href{https://github.com/sentient-agi/OpenDeepSearch}{Yes} & 25.03 \\
& ReSum ~\cite{wu2025resum} & \makecell{Def.+Gen.} & LRM & API & Multi  & Industry & \href{https://github.com/Alibaba-NLP/DeepResearch//}{Yes} & 25.09 \\
\midrule

\rowcolor{black!5}\multicolumn{9}{c}{\textit{\textbf{Model-Native-Based Paradigm}}}\\
\midrule

\multirow{6}{*}{\makecell[c]{\textit{\textbf{Offline}}\\ \textit{\textbf{Training}}}}
& R1-Searcher ~\cite{song2025r1}& Def. & LLM & API & Single & Academia & \href{https://github.com/RUCAIBox/R1-Searcher}{Yes} & 25.03 \\
& Search-R1 ~\cite{jin2025searchr1trainingllmsreason}& Def. & LLM & API & Single & Academia & \href{https://github.com/PeterGriffinJin/Search-R1}{Yes} & 25.03 \\
& ReSearch ~\cite{chen2025researchlearningreasonsearch}& Def. & LLM & API & Single & Academia & \href{https://github.com/Agent-RL/ReCall}{Yes} & 25.03 \\
& R1-Searcher++ ~\cite{song2025r1plusplus}& Def. & LLM & API & Multi & Academia & \href{https://github.com/RUCAIBox/R1-Searcher-plus}{Yes} & 25.05 \\
& R-Search ~\cite{zhao2025r}& Def. & LLM & API & Single & Academia & \href{https://github.com/QingFei1/R-Search}{Yes} & 25.06 \\
& M2IO-R1 ~\cite{xiao2025m2io}& Def. & LMM & API & Single & Academia & {No} & 25.08 \\
\midrule

\multirow{10}{*}{\makecell[c]{\textit{\textbf{Online}}\\ \textit{\textbf{Training}}}}
& DeepResearcher ~\cite{zheng2025deepresearcherscalingdeepresearch}& Def. & LLM & \makecell{API\\+Browser} & Multi & Academia & \href{https://github.com/GAIR-NLP/DeepResearcher}{Yes} & 25.04 \\
& WebThinker ~\cite{li2025webthinker} & Def. & \makecell{LRM\\+LLM} & \makecell{API\\+Browser} & Multi & Academia & \href{https://github.com/RUC-NLPIR/WebThinker}{Yes} & 25.04 \\
& ZeroSearch ~\cite{sun2025zerosearchincentivizesearchcapability}& Def. & LLM & API & Multi & Academia & \href{https://github.com/Alibaba-NLP/ZeroSearch}{Yes} & 25.05 \\
& MMSearch-R1 ~\cite{wu2025mmsearchr1} & Def. & LMM & API & Single & Industry & \href{https://github.com/EvolvingLMMs-Lab/multimodal-search-r1}{Yes} & 25.06 \\
& WebWatcher ~\cite{geng2025webwatcher}& Def. & LMM & API & Single & Industry & \href{https://github.com/Alibaba-NLP/DeepResearch}{Yes} & 25.09 \\
& SFR-DeepResearch ~\cite{nguyen2025sfr}&  Gen. & LRM & API & Single & Industry & {No} & 25.09 \\
& DeepDive ~\cite{lu2025deepdive}& Def. & LRM & \makecell{API\\+Browser} & Single & Academia & \href{https://github.com/THUDM/DeepDive}{Yes} & 25.09 \\
& WebResearcher ~\cite{qiao2025webresearcher}& Gen. & LRM & \makecell{API\\+Browser} & Multi & Industry & \href{https://github.com/Alibaba-NLP/DeepResearch}{Yes} & 25.09 \\

\end{longtable}
}


\subsubsection{\textbf{Summary and Discussion}}
\hspace{2mm}
\label{sec:dr_discussion}
The preceding subsections have separately introduced the implementation mechanisms and representative works of the pipeline-based and model-native paradigms.
In this subsection, we synthesize these two trajectories from a broader evolutionary perspective. Specifically, we analyze four converging dimensions that characterize the ongoing transformation of the Deep Research paradigm:
(1) \textit{information acquisition} methods diversifying across API-based and browser-based approaches;
(2) \textit{task objectives} expanding from deterministic question answering to open-ended generative synthesis;
(3) \textit{base models} evolving from LLMs to LRMs and further to LMMs~(Large Multimodal Models); and
(4) \textit{agent architectures} shifting from single-agent coordination to multi-agent collaborative systems. An overview of deep research agents is summarized in Table~\ref{tab:deep research}.

\paragraph{\textbf{Information Acquisition} }
\hspace{2mm}
Information acquisition denotes the use of external tools to obtain information from the environment during the reasoning and answering process of Deep Research Agents. 
According to the mode of interaction, existing approaches can be broadly categorized into API-based and browser-based methods.

\begin{itemize} \item \textbf{API-based methods.}\quad \setlength{\parindent}{1em} 
These methods acquire external information through APIs that interface with either static, local databases or live, third-party search engines. 

\indent A significant line of work focuses on interacting with live search engines to access real-time, open-domain information. For example, systems like Search-o1~\cite{li2025search}, ODS~\cite{alzubi2025open}, and ReSum~\cite{wu2025resum} utilize APIs from commercial search engines like Bing or Google, often in combination with web page parsing tools like Jina Reader. This approach has also been extended to multimodal search: MMSearch-R1~\cite{wu2025mmsearchr1} and WebWatcher~\cite{geng2025webwatcher} employ SerpAPI for both image and text retrieval. To handle more specialized information needs, WebResearcher~\cite{qiao2025webresearcher} integrates multiple APIs, including Google Search and Google Scholar, and uses a dedicated summarization model to process the results.

\indent Another line of research prioritizes reproducibility and training stability by using static databases, such as a Wikipedia dump, as the external knowledge source. This controlled environment allows for more rigorous evaluation of an agent's reasoning and retrieval strategies without the noise and non-stationarity of the open web. Representative works in this category include Search-R1~\cite{jin2025searchr1trainingllmsreason}, ReSearch~\cite{chen2025researchlearningreasonsearch}, R-Search~\cite{zhao2025r}, and the R1-Searcher series~\cite{song2025r1, song2025r1plusplus}. M2IO-R1~\cite{xiao2025m2io} further adapts this approach for multimodal scenarios by retrieving from a pre-constructed database of text-image pairs. To bridge the gap between static and live environments, ZeroSearch~\cite{sun2025zerosearchincentivizesearchcapability} introduces a novel training strategy where an LLM is used to simulate a search engine, generating both relevant and noisy documents. This allows the agent to be trained with curriculum-based difficulty, improving robustness while avoiding the high cost and instability of live API calls.

\item \textbf{Browser-based methods.}\quad \setlength{\parindent}{1em}
Complementary to API-based approaches, browser-based methods enable more human-like web interaction through automated browser control, supporting dynamic page rendering, element interaction, and the handling of complex web scenarios. Related methods can be distinguished by their level of abstraction.

\indent Some works focus on the low-level mechanics of direct browser interaction. For example, WebVoyager~\cite{he2024webvoyager} renders HTML into visual web pages and interacts with elements using a combination of textual information and screenshots. This allows it to handle dynamic elements like floating advertisements, thereby improving robustness in real-world web environments.

\indent Other approaches operate at a higher level of abstraction, focusing on information and workflow management. DeepResearcher~\cite{zheng2025deepresearcherscalingdeepresearch} invokes a dedicated browsing agent to sequentially read and extract relevant information, managing a short-term memory to condense the information before returning it to the main agent. Similarly, WebThinker~\cite{li2025webthinker} uses an auxiliary LLM to summarize crawled pages based on the agent's click intent, and in its report generation mode, stores explored pages in a document memory for later retrieval. Finally, some frameworks represent a hybrid approach; DeepDive~\cite{lu2025deepdive}, for instance, abstracts browser interactions into a set of discrete function calls like $(search, click, open)$, combining the direct control of browser-based methods with the structured nature of API-based invocation.

\end{itemize}

\paragraph{\textbf{Task Objectives} }
\hspace{2mm}
According to the goal of information research, Deep Research agents can be broadly divided into two categories: \textit{definitive answering} and \textit{generative synthesis}.

\begin{itemize}
    \item \textbf{Definitive answering.}\quad \setlength{\parindent}{1em}
    This category focuses on concise, verifiable factual questions that emphasize rapid information localization and multi-hop verification.Representative works such as WebVoyager~\cite{he2024webvoyager}, Real-World WebAgent~\cite{gur2023real}, Search-o1~\cite{li2025search}, ODS~\cite{alzubi2025open}, and Search-R1~\cite{jin2025searchr1trainingllmsreason} effectively employ single-hop or multi-hop search strategies to locate precise answers to user queries.

    \item \textbf{Generative synthesis.}\quad \setlength{\parindent}{1em}
    This category addresses complex, open-ended tasks that require integrating multiple sources of evidence and constructing coherent arguments around thematic questions.  
    Such works often employ multimodal or structured representations—such as tables and images—to enhance information density and presentation quality.   WebThinker~\cite{li2025webthinker} generates research-style reports incorporating Markdown-formatted tables for structured summarization.  
    M2IO-R1~\cite{xiao2025m2io} produces interleaved text–image outputs, combining retrieved multimodal content with contextual reasoning.  
    SFR-DeepResearch~\cite{nguyen2025sfr} supports long-form report generation, synthesizing retrieved knowledge into extended, well-organized documents. 
\end{itemize}

\paragraph{\textbf{Base Models} }
\hspace{2mm}
Across both pipeline-based and model-native paradigms, Deep Research agents have evolved in base model design, from LLMs to LRMs and further to LMMs.
Pipeline-based methods rely on LLMs or LRMs guided by explicit prompting frameworks (e.g., ReAct), while model-native approaches use reinforcement learning to internalize reasoning and retrieval.
With the emergence of reasoning-optimized LRMs such as OpenAI-o1~\cite{openai_o1_2024} and DeepSeek-R1~\cite{deepseek_r1_2025}, models increasingly integrate tool use within their architecture.
To handle multimodal environments, LMMs extend these capabilities to images, webpages, and videos, enabling richer evidence acquisition.

\begin{itemize}
    \item \textbf{LLM.}\quad \setlength{\parindent}{1em}
    Search-R1~\cite{jin2025searchr1trainingllmsreason}, R1-Searcher~\cite{song2025r1}, ReSearch~\cite{chen2025researchlearningreasonsearch}, R-Search~\cite{zhao2025r} utilize a single LLM that, after reinforcement learning for deep research work, becomes an LRM that internalizes tool usage mastery, responsible for completing the entire process of reasoning, generating search queries, understanding retrieval results, and generating answers.

    \item \textbf{LRM.}\quad \setlength{\parindent}{1em}
    Search-o1~\cite{li2025search} enables the model to autonomously decide when to retrieve and refine documents into compact ``knowledge steps'' that support current reasoning.  
    WebThinker~\cite{li2025webthinker} performs end-to-end reasoning, invoking tools such as search, navigation, and report generation within its reasoning chain.  
    SFR-DeepResearch~\cite{nguyen2025sfr} and DeepDive~\cite{lu2025deepdive} adopt multiple reasoning-optimized LRMs as their core base models.
    \item \textbf{LMM.}\quad \setlength{\parindent}{1em}
    Multimodal extensions further enhance Deep Research capabilities.  
    WebVoyager~\cite{he2024webvoyager} defaults to gpt-4-vision-preview model.
    MMSearch-R1~\cite{wu2025mmsearchr1} uses three types of LMMs. Qwen2.5-VL-7B-Instruct serves as the policy model, learning to execute on-demand search through RL training. Qwen3-32B serves as the webpage summarization model for content summarization in text search pipelines.
    GPT-4o serves as the judge model for evaluating model response accuracy. WebWatcher~\cite{geng2025webwatcher} uses two types of LMMs. Qwen2.5-VL-7B and Qwen2.5-VL-32B serve as policy models trained through SFT and RL. 
    GPT-4o serves as an auxiliary model for data generation, trajectory annotation, and evaluation.
    M2IO-R1~\cite{xiao2025m2io} uses Qwen2.5-VL series for image insertion tasks, with GPT-4o for text answer generation and auxiliary data construction.
\end{itemize}

\paragraph{\textbf{Agent Architectures} }
\hspace{2mm}
Whether in pipeline-based or model-native paradigms, Deep Research Agents have developed along two architectural lines: single-agent and multi-agent.
Single-agent systems integrate planning, retrieval, reasoning, and generation within one model, featuring simplicity, efficiency, and low cost.
In contrast, multi-agent architectures assign specialized roles such as planning, retrieval, summarization, and writing to multiple collaborating models, improving scalability, robustness, and adaptability for complex, long-horizon, or multimodal tasks.

\begin{itemize}
    \item \textbf{Single-agent architecture.}\quad \setlength{\parindent}{1em}
    Single-agent architectures unify planning, retrieval, comprehension, and generation within one model, achieving end-to-end execution with minimal computational overhead and a shorter training pipeline. 
    Search-R1~\cite{jin2025searchr1trainingllmsreason}, R1-Searcher~\cite{song2025r1}, and ReSearch~\cite{chen2025researchlearningreasonsearch} follow this pattern, using a single LLM trained through reinforcement learning to perform reasoning, generate search queries, interpret retrieved content, and produce final answers. 
    Such an approach is well suited to factual or task-specific applications.

    \item \textbf{Multi-agent architecture.}\quad \setlength{\parindent}{1em}
    Compared with single agents that are limited by context capacity and role overload in complex reasoning, multi-agent architectures distribute specialized roles such as planning, retrieval, summarization, and writing to enhance scalability, robustness, and overall performance.
    Real-World WebAgent~\cite{gur2023real} separates planning and summarization using HTML-T5, while Flan-U-PaLM generates executable programs, improving coordination and efficiency. DeepResearcher~\cite{zheng2025deepresearcherscalingdeepresearch} employs a Web Browsing Agent for segmented retrieval and summarization, reducing context load on the main agent while preserving essential information for reasoning. 
    ReSum~\cite{wu2025resum} integrates a reasoning policy model with a summarization tool that compresses historical context when limits are reached, enabling continuous exploration. 
    WebThinker~\cite{li2025webthinker} assigns its main LRM to reasoning and planning, and an auxiliary LLM to search intent generation, content summarization, and report composition, improving logical coherence. 
    R1-Searcher++~\cite{song2025r1plusplus} adopts dual-model collaboration during reinforcement learning, where a rewrite model transforms retrieved documents into internal reasoning paths, allowing the policy model to internalize external knowledge and reduce retrieval frequency without compromising accuracy.
\end{itemize}

\paragraph{\textbf{Reward Design} }
\hspace{2mm}
Beyond the common trends shared by both pipeline-based and model-native paradigms, reward design in reinforcement learning is a mechanism unique to the latter.
It determines whether models can internalize retrieval and reasoning strategies—when to search, how to filter evidence, and how to generate reliable results.
Through fine-grained reward shaping, model-native approaches transform externally guided behaviors into self-regulated reasoning capabilities.
Existing studies mainly explore three types of reward signals: outcome, format, and process rewards.

\begin{itemize}
    \item \textbf{Outcome rewards.}\quad \setlength{\parindent}{1em}
    Outcome rewards assess only the final result, encouraging correct outputs without considering intermediate reasoning quality.  
    DeepSeek-R1~\cite{deepseek_r1_2025} demonstrates that outcome rewards alone can effectively guide reasoning.  Similarly, Search-R1~\cite{jin2025searchr1trainingllmsreason}, ZeroSearch~\cite{sun2025zerosearchincentivizesearchcapability}, and SFR-DeepResearch~\cite{nguyen2025sfr} adopt simple outcome-based supervision.

    \item \textbf{Format rewards.}\quad \setlength{\parindent}{1em}
    Format rewards ensure structural correctness within reasoning chains, aligning thinking steps, tool calls, and outputs.  R1-Searcher~\cite{song2025r1}, ReSearch~\cite{chen2025researchlearningreasonsearch}, DeepResearcher~\cite{zheng2025deepresearcherscalingdeepresearch}, WebWatcher~\cite{geng2025webwatcher}, and M2IO-R1~\cite{xiao2025m2io} constrain models to wrap reasoning, retrieval, and final answers within predefined tags.  
    DeepDive~\cite{lu2025deepdive} further applies a strict binary reward that requires both correct format and correct answer simultaneously.

    \item \textbf{Process rewards.}\quad \setlength{\parindent}{1em}
    Process rewards provide stepwise feedback, guiding models to balance reasoning depth, efficiency, and traceability.  
    R1-Searcher++~\cite{song2025r1plusplus} combines format, outcome, and group rewards, encouraging accurate answers with minimal retrievals.  
    R-Search~\cite{zhao2025r} extends this approach with multi-stage reward mechanisms to jointly optimize reasoning quality and resource efficiency.
\end{itemize}

\paragraph{\textbf{Future Directions} }
\hspace{2mm}
Building on the evolutionary framework of Deep Research agents, we identify key challenges and outline future research trajectories. 
These directions collectively aim to advance agents from instruction-following tools to intelligent research partners that understand problems, explore knowledge boundaries, and co-evolve with humans.

\begin{itemize} \item \textbf{Information acquisition.} 
Agents should dynamically choose acquisition strategies—using APIs for structured, time-sensitive queries and browsers for interactive, multimodal scenarios. 
Emerging AI-native browsers (e.g., Browserbase, Browser Use, Perplexity’s Comet) are narrowing this gap by providing stable DOM access, asynchronous execution, and automated anti-scraping, bringing browser-based methods closer to API-level reliability. 

\item \textbf{Task objectives.} 
Future research may progress in three directions:
(1) Interactive research with user feedback loops to refine intent and deepen exploration;
(2) Continuous ``Living Research'' for persistent topic tracking and periodic incremental reporting;
(3) Predictive analysis that anticipates future trends and potential breakthroughs from historical data.

\item \textbf{Agent architecture.} 
Agents should assess task complexity, evaluate interdependencies, and autonomously decide between serial or parallel collaboration. This can be achieved by modeling workflows as mutable graphs and optimizing them through evolutionary or adaptive algorithms.

\item \textbf{Reward design.} 
Incorporating curiosity-driven and self-reward mechanisms~\cite{pathak2017curiosity,yuan2024self} can encourage exploration and self-assessment. Intrinsic rewards promote uncertainty reduction, while self-evaluation reduces reliance on external supervision. 
\end{itemize}

\subsection{GUI Agent}
\subsubsection{\textbf{Overview}}
\hspace{2mm}
Graphical User Interface (GUI) agents refer to intelligent entities capable of autonomously perceiving, planning, and executing tasks in GUI environments. Their development also reflects a clear evolution from an external, pipeline-based paradigm to an internalized, model-native paradigm. 

Early GUI automation relied on a \textit{pipeline-based} paradigm, initially driven by system-based workflows like record-and-replay~\cite{barman2016ringer, guo2019sara} or rule-based scripts~\cite{AGOSTINELLI2022103721, li2017sugilite}. While stable for repetitive tasks, the hard-coded system workflow lacked the necessary understanding of interface semantics and user intent. The advent of LLMs introduced prompt-based methods, replacing rigid scripts with natural language instructions and enabling agents like AppAgent~\cite{zhang2023appagent} and Mobile-Agent~\cite{wang2024mobileagent} to exhibit a preliminary degree of autonomy. However, these agents still operated within manually orchestrated frameworks, invoking LLMs as external tools that were not specifically optimized for the nuances of GUI interaction, thus failing to fully leverage their inherent potential.

\begin{figure}[t] 
  \centering 
  \includegraphics[width=0.98\textwidth]{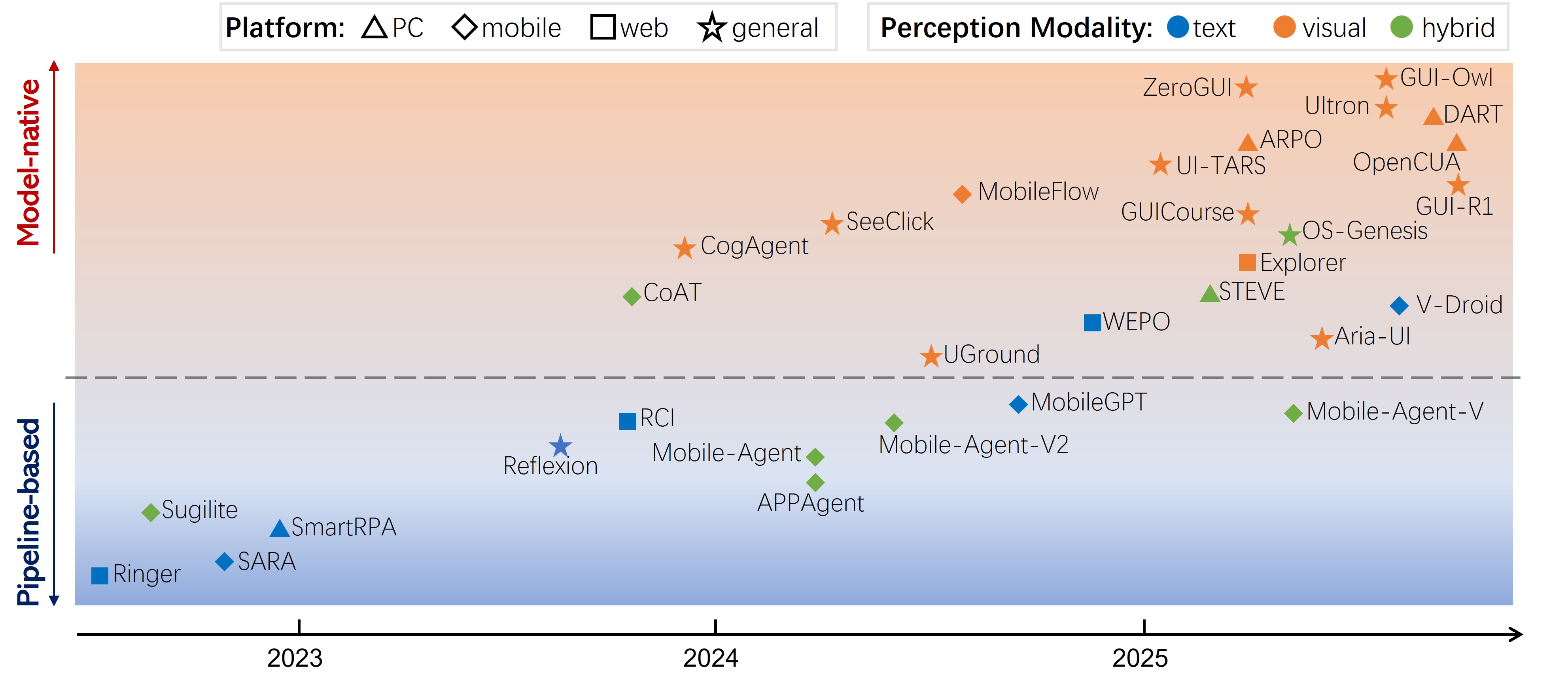} 
  \caption{Overview of GUI agent methods.} 
  \label{fig:GUI_Agent} 
\end{figure}

To overcome these limitations, current research is progressively moving towards the \textit{model-native} paradigm, which internalizes the core capabilities of a GUI agent within the model's parameters. This transition has progressed along a path of increasing integration, beginning with \textit{modular training}. This approach solved critical bottlenecks by individually training components, such as the perception-focused models in UGround~\cite{gou2025navigating} and Aria-UI~\cite{yang-etal-2025-aria}, or by fusing modules as seen in CogAgent~\cite{10655402}. This evolutionary path is ultimately leading toward \textit{end-to-end training}, where one single model learns the direct mapping from perceptual inputs to executable actions. This fully integrated approach, demonstrated by recent methods like UI-TARS~\cite{qin2025ui_tars} and GUI-owl~\cite{ye2025mobileagentv3}  significantly enhances an agent's autonomy and generalization, allowing it to move beyond its dependency on external frameworks. An overview of GUI Agent methods is illustrated in Fig.~\ref{fig:GUI_Agent}.

\subsubsection{\textbf{Pipeline-based Paradigm}}
\hspace{2mm}
In their early development, GUI Agents primarily relied on manually orchestrated pipelines that decomposed tasks into fixed modules linked by explicit logic. Within this framework, models were assigned only limited functions at specific steps of the process. Based on the autonomy of the agent's decision-making, the pipeline paradigm can be further divided into two sub-stages: system-based and prompt-based. The former, corresponding to the traditional Robotic Process Automation (RPA) paradigm, relies on record-and-playback or rule-based scripts to replay a fixed sequence of operations. The latter utilizes prompts to drive LLMs to autonomously handle functions such as task understanding, planning, and tool calling to complete a GUI task.

\paragraph{\textbf{System-based Workflows}} 
\hspace{2mm}
In the early development of GUI Agents, models did not yet have the ability to perceive interface states or make independent decisions. Consequently, automation relied on system-level orchestration to reproduce human operations in fixed environments, aiming for greater efficiency and repeatability. According to the control method, this stage can be divided into two types: record-and-replay and rule-based scripting automation.
\begin{itemize}
    \item \textbf{Record-and-replay.}\quad \setlength{\parindent}{1em}
    This class of methods achieves task automation by recording a user's action sequence (e.g. clicks and inputs) and replaying it verbatim within an identical environment. For instance, CoScripter\cite{leshed2007coscripter} first proposed a method for users to record actions directly within a web browser, which could then be replayed as scripts to enable the automation and sharing of web-based workflows. Subsequently, RERAN\cite{gomez2013reran} extended this concept to mobile platforms, achieving high-precision touch replay by capturing low-level Linux input events. 
    
    \indent To improve robustness, subsequent research introduced more intelligent matching and waiting mechanisms. Ringer\cite{barman2016ringer}, for example, enhanced replay success rates across different website versions by recording multi-dimensional attributes of UI elements and calculating their similarity during the replay phase. It could also automatically insert wait conditions to handle asynchronous scenarios. In the mobile domain, SARA\cite{guo2019sara} introduced adaptive replay and context-capturing mechanisms, enabling recorded scripts to be executed across heterogeneous devices. 
    
    \indent Although these improvements enhanced the environmental adaptability of the scripts, this approach still depended on fixed environments and specific examples. Consequently, it suffered from poor generalization and was unable to support open-ended tasks.

    \item \textbf{Rule-based scripting.}\quad \setlength{\parindent}{1em}
    To overcome the brittleness of record-and-replay methods and to support more complex and stable automation, the research community shifted towards rule-based scripting. This approach involves controlling the operational logic of an agent through explicit rules or conditional statements. For example, Chickenfoot\cite{bolin2005automation} pioneered the use of natural language-like JavaScript to manipulate web elements, which enabled users to control tasks through semantic commands such as `` click(login)''. Subsequently, Sikuli\cite{yeh2009sikuli} advanced this approach by using screenshot matching instead of coordinate-based positioning, making scripts more robust against layout changes. 
    
    \indent As this technology progressed towards industrial applications, the research focus shifted to automating and enhancing the intelligence of rule-based scripts. SmartRPA\cite{AGOSTINELLI2022103721}, for instance, proposed automatically synthesizing rule-based scripts from user interface logs to reduce manual effort. Meanwhile, Sugilite\cite{li2017sugilite} explored a collaborative human-computer programming paradigm that combined voice and demonstration, allowing end-users to generate conditional scripts from natural language instructions. 
    
    \indent While rule-based scripting offered greater flexibility and a degree of logical control compared to record-and-replay, it was still critically constrained by its lack of perceptual and semantic understanding capabilities.
    \end{itemize} 
    
In summary, the system-based workflow is defined by its reliance on deterministic control through either explicit rules or direct action replay. This approach offered high stability and interpretability but was ultimately limited by its lack of semantic understanding and its inability to generalize to novel tasks.
    
\paragraph{\textbf{Prompt-based Methods}} 
\hspace{2mm}
With the advent of LLMs, the research focus for GUI Agents shifted from scripted execution to language-driven decision generation. This prompt-based stage leverages the in-context learning capabilities of a pre-trained LLM, using prompts to guide the model in understanding task instructions, parsing interface states, and generating action plans without requiring additional training. This transition enabled agents to adapt to multi-platform GUI environments using a unified language interface, thereby achieving a preliminary form of generality. As work in this area progressed to enhance the coherence and reliability of planning, four primary developmental paths emerged:

\begin{itemize}
    \item \textbf{Single-step reactive agents.}\quad \setlength{\parindent}{1em}
    Early prompt-based GUI agents often adopted a single-step reactive approach, where the agent generates and executes a single action immediately following each observation of the interface. The ReAct framework~\cite{yao2023react} pioneered this by having an LLM generate a ``Thought'' and an ``Action'' in each round, enabling zero-shot completion of tasks in text-based environments like ALFWorld. Building on this, Auto-GUI\cite{zhang2024lookscreensmultimodalchainofaction} constructed a multimodal context that included action history to inform its single-step decisions. This concept was further adapted to mobile platforms by MobileAgent~\cite{wang2024mobileagent}, which concatenates screenshots, OCR information, and user instructions into a multimodal prompt. This process allows the agent to produce short-term reasoning and an action at each step, achieving low-latency reactive operations. 
    
    \indent However, a significant limitation of purely reactive methods is their lack of a global plan, which can lead to inefficient looping behaviors or failure to address the overall task objective.
    
    \item \textbf{Hierarchical planning agents.}\quad \setlength{\parindent}{1em}
    To address the limitations of purely reactive methods, researchers introduced the hierarchical planning paradigm. In this paradigm, the agent first generates a high-level, global plan and then executes it, with the ability to dynamically modify sub-plans. For instance, WebAgent~\cite{gur2024a} employs a three-stage process where an LLM first decomposes a task, a program synthesizer converts the plan into a script, and the script is then executed. In the mobile domain, the MobileAgent-v2~\cite{wang2024mobileagentv2} series extended this concept into a multi-agent, layered architecture where a high-level planner summarizes task progress to guide a separate low-level decision-making agent. 
    
    \indent Such hierarchical structures compensate for the lack of global control found in reactive strategies. However, their reliance on pre-defined plans can limit their ability to generalize across platforms or to correct errors during execution. 
    
    \item \textbf{Reflective agents.}\quad \setlength{\parindent}{1em}
    This approach introduces a self-evaluation and correction loop, enabling the agent to identify and rectify its own errors.
    The Reflexion~\cite{NEURIPS2023_1b44b878} framework allows an agent to analyze task failures and store these reflections in memory to avoid repeating mistakes, which has been shown to improve success rates. Similarly, the RCI~\cite{kim2023language} framework demonstrated increased accuracy in GUI form-filling tasks by incorporating a self-correction mechanism. In mobile scenarios, Mobile-Agent-V~\cite{wang2025mobile} achieves self-correction through a fusion of visual backtracking and linguistic reflection, reviewing past demonstration videos to evaluate its actions and generate improvement strategies.
    
    \item \textbf{Exploratory agents.}\quad \setlength{\parindent}{1em}
    In contrast to the post-execution error correction of reflective agents, exploratory agents proactively generate and evaluate multiple candidate paths via simulation or search before selecting the optimal one for execution. The Tree-of-Thoughts (ToT)~\cite{yao2023tree} framework, for example, extends the LLM's reasoning process into a tree structure, enabling parallel reasoning and self-evaluation to prune less promising paths at critical decision points. AppAgent~\cite{zhang2023appagent} improves stability by leveraging a knowledge base built during an initial, task-agnostic exploration phase. Furthering this idea, MobileGPT~\cite{lee2024mobilegpt} combined exploratory decision-making with a hierarchical memory to enable long-term knowledge accumulation, demonstrating improved generalization and stability in zero-shot, cross-application scenarios.
\end{itemize}

The prompt-based methods demonstrated the potential of using LLMs for GUI agent tasks, freeing them from the constraints of manual scripting and enabling zero-shot completion of complex tasks via natural language. 

However, this paradigm also revealed significant limitations: (1) perceptual constraints, due to a reliance on textual descriptions of the GUI; (2) insufficient state tracking, because of limited context windows; and (3) high inference costs from multi-round reflection and exploration. These collectively motivated the research community to advance toward the model-native paradigm, with the aim of improving representational stability and generalization through direct model training.

\subsubsection{\textbf{Model-native Paradigm}}
\hspace{2mm}

To overcome the limitations of the pipeline-based approaches, research has shifted toward the model-native paradigm, which aims to internalize core GUI agent capabilities into the model's parameters through data-driven training. These capabilities can be trained either individually or jointly, leading to the development along two primary paths distinguished by their degree of integration: \textit{modular training} and \textit{end-to-end training}.

\paragraph{\textbf{Modular Training}} 
\hspace{2mm}
Modular training serves as a critical transitional stage between pipeline-based systems and fully end-to-end learning. In this approach, the complex GUI interaction process is decomposed into several learnable functional components such as perception, planning, and execution, which are trained either independently or in a partially coupled manner. This methodology enhances interpretability and data efficiency by allowing individual capabilities to mature under targeted supervision before their eventual integration. To provide a clearer overview of development in this stage, we categorize the related work into three main directions:
\begin{itemize}
    \item \textbf{Foundational modules: perception and execution.}\quad \setlength{\parindent}{1em}
    In the initial phase of modular training, research focused on establishing the agent's fundamental interaction loop, which consists of perception, planning, and execution. In this loop, the perception module analyzes the structure and semantic elements of the graphical interface; the planning module decomposes high-level user goals into sub-tasks and generates executable steps; and the execution module maps these abstract plans into concrete actions.

    \indent Early work on the perception module focused on parsing the screen into structured elements. For instance, OmniParser\cite{lu2024omniparser} converts screen content into readable bounding boxes and semantic labels. Subsequent research began to focus more specifically on grounding. UGround\cite{gou2025navigating} trained a ``universal GUI grounding'' model on a large-scale dataset consisting of 1.3 million screenshots and tens of millions of UI elements. Further advancing this work, Aria-UI\cite{yang-etal-2025-aria} achieved purely visual grounding across diverse instruction formats and in dynamic contexts, setting new state-of-the-art results on multiple online agent benchmarks. More recently, research has begun to introduce reinforcement learning paradigms into perception training. For example, SE-GUI\cite{yuan2025enhancing} utilized self-evolving fine-tuning with dense, point-level rewards through GRPO to stably improve localization robustness. Overall, this line of research has collectively established a stable correlation: improvements in the performance of the grounding-centric perception module are consistently linked to higher overall task success rates.

    \indent Research in planning module follows two main approaches: imitative and generative planning. Imitative planning learns from expert demonstrations. For instance, CoAT\cite{zhang-etal-2024-android} makes the ``chain of action'' and ``chain of thought'' explicit for supervised fine-tuning, enhancing interpretability and robustness. Similarly, WebAgent\cite{gur2024a} constructs a modular pipeline to normalize and orchestrate natural language instructions. In contrast, generative planning enables agents to create plans more autonomously. WMA-Agent\cite{chae2024web}, for example, allows the planner to ``imagine before deciding'' by learning the environment's dynamics, showing performance gains on benchmarks like WebArena. Another approach, WebSynthesis\cite{gao2025websynthesisworldmodelguidedmctsefficient}, uses a world model with MCTS to synthesize high-quality training trajectories, thereby augmenting its policy learning process.

    \indent Execution modules can be divided into two categories based on their action space: atomic-level and widget-level executors. Atomic-level executors decode primitive, coordinate-based actions like click and type, focusing on precision and reliability. Research in this area includes UI-R1\cite{lu2025ui},  which enhanced atomic action accuracy using rule-based rewards; and OS-Kairos\cite{cheng-etal-2025-os}, which enhances the robustness of its atomic-level actions by jointly optimizing for confidence estimation. Widget-level executors, in contrast, perform semantic selection on UI elements to align with human intent. For instance, WEPO\cite{10.1609/aaai.v39i25.34863} used preference learning to reduce selection errors caused by element ambiguity. In summary, the two executor types offer a clear trade-off. Atomic-level executors provide high precision and generality but are vulnerable to perceptual noise and lack semantic context. Conversely, widget-level executors align well with human intent and offer better generalization, but they depend on high-quality perception and struggle with fine-grained operations.
    
    \item \textbf{Advanced modules: reflection and exploration.}\quad \setlength{\parindent}{1em}
    As foundational perception and execution capabilities matured, the research focus shifted to higher-level agentic capabilities to self-assess, self-correct, and proactively explore. The reflection module helps the model identify and correct erroneous decisions through continuous evaluation and feedback on the execution process. In parallel, the exploration module is responsible for autonomously generating high-quality interaction data to support continuous learning and generalization. The introduction of these two module types marks the evolution of GUI Agents from passive executors into proactive entities capable of learning and adaptation.

    \indent Training for the reflection module follows two main paradigms. Discriminative Reflection uses supervised signals to judge the correctness of local actions, enabling real-time self-checking and rollbacks. In contrast, Evaluative Reflection assesses entire sequences through reward modeling or preference learning to guide policy optimization.

    In discriminative reflection, the focus is on step-by-step verification. For instance, STEVE\cite{lu2025steve} reflects on each action's validity by using a large multimodal model to verify the screen transition (previous screen, action, next screen). V-Droid\cite{dai2025advancing}, on the other hand, reflects by using a verifier to evaluate a set of candidate actions based on learned task progress preferences. In evaluative reflection, the goal is to assess the overall quality of a sequence. GUI-PRA\cite{xiong2025gui} employs a Process Reward Model (PRM) to reflect on entire action sequences, using this evaluation to shape rewards during training and re-rank actions during inference. Similarly, UI-Genie\cite{xiao2025ui} uses a learned reward model to score actions. This reflective scoring serves a dual purpose: re-ranking actions during inference and providing a reward signal for fine-tuning, thereby creating a self-improvement loop.

     \indent The exploration module aims to autonomously produce high-quality interaction data for GUI agents, creating a ``data flywheel'' that improves sample efficiency and robustness. Early research focused on the method of exploration, evolving from broad synthesis to more targeted and efficient strategies. Initially, the focus was on scalable trajectory synthesis. For instance, Explorer\cite{pahuja-etal-2025-explorer} explored the web using heuristic and self-learning strategies to synthetically generate large-scale, diverse interaction data. Subsequently, the research shifted to backward task synthesis, where agents explore more freely. OS-Genesis\cite{sun-etal-2025-os}, for example, explores a real environment without a predefined task and then derives high-quality, labeled trajectories by backtracking from its interaction logs, using a reward model for quality control. To maximize efficiency, GUI-Xplore\cite{11093114} demonstrated that an agent can perform just one or a few explorations to learn strategies that are generalizable across sub-tasks, highlighting the power of targeted, few-shot exploration.

    \item \textbf{Module fusion.}\quad \setlength{\parindent}{1em}
    To mitigate the information mismatches and error propagation that can arise from training modules in isolation, researchers have attempted to incorporate multiple sub-modules into a unified training objective. The related studies can be categorized by which modules are fused.
    
    \indent One line of work focuses on \textit{perception-planning fusion}. For example, CoCo-Agent~\cite{ma-etal-2024-coco} uses joint training of separate modules, AGUVIS~\cite{xu2025aguvis} connects them via a structured ``Inner Monologue'' to improve information flow, and CogAgent~\cite{10655402} unifies both capabilities within a single model through multi-task pre-training. 
    
    \indent  A second line, \textit{planning-execution fusion}, aims to bridge the gap between high-level intent and low-level action. MagicGUI~\cite{tang2025magicguifoundationalmobilegui} achieves this by training a single, unified policy network with reinforcement learning. In contrast, PILOT-RL~\cite{lu2025pilotrltraininglanguagemodel} fuses these stages by embedding a world model into the decision loop, guiding the policy search with imagined outcomes. These fusion strategies represent a critical step toward reducing inter-module error and are a clear progression towards fully end-to-end training paradigms.    
\end{itemize}

The future progression of this research is toward more efficient joint training and cross-task generalization, while preserving interpretability. However, the inherent need to reduce error propagation continues to drive the field from module fusion toward fully end-to-end training paradigms.

\newcolumntype{C}[1]{>{\centering\arraybackslash}p{#1}}
\newcolumntype{M}[1]{>{\centering\arraybackslash}m{#1}}

\afterpage{
\clearpage
\footnotesize

\begin{longtable}{
M{0.118\textwidth}| 
  l|
  M{0.08\textwidth}| 
  M{0.07\textwidth}| 
  M{0.08\textwidth}| 
  M{0.098\textwidth}| 
  M{0.06\textwidth}| 
  M{0.06\textwidth}  
}
\caption{Overview of GUI Agents.}\label{tab:GUI Agent}\\
\toprule

\rowcolor{blue!5}
\textbf{} & 
\multicolumn{1}{|l|}{\cellcolor{blue!5}\textbf{Method}} & 
\textbf{Platform} & 
\textbf{Modal} & 
\textbf{Action} & 
\textbf{Affiliation} & 
\textbf{Access} & 
\textbf{Date}\\
\midrule
\endfirsthead

\multicolumn{8}{c}{\tablename~\thetable\ -- \textit{Overview of GUI Agents (continued).}}\\
\toprule
\rowcolor{blue!5}
\textbf{} & 
\multicolumn{1}{|l|}{\cellcolor{blue!5}\textbf{Method}} & 
\textbf{Platform} & 
\textbf{Modal} & 
\textbf{Action} & 
\textbf{Affiliation} & 
\textbf{Access} & 
\textbf{Date}\\
\midrule
\endhead

\midrule
\multicolumn{8}{r}{\textit{continued on next page}}\\
\endfoot

\bottomrule
\endlastfoot

\rowcolor{black!5}\multicolumn{8}{c}{\textit{\textbf{Pipeline-based Paradigm}}}\\
\midrule

\multirow{4}{*}{\makecell[c]{\textit{\textbf{System-based}}\\ \textit{\textbf{Workflow}}}}
& Ringer ~\cite{barman2016ringer} & Web & Text & Widget & Academia & \href{https://github.com/sbarman/webscript}{Yes} & 16.11 \\
& Sugilite ~\cite{li2017sugilite} & Mobile & Hybrid & Widget & Academia & \href{https://github.com/tobyli/Sugilite_development}{Yes} & 17.05 \\
& SARA ~\cite{guo2019sara} & Mobile & Text & Widget & Academia & \href{https://github.com/microsoft/SARA}{Yes} & 19.07 \\
& SmartRPA ~\cite{AGOSTINELLI2022103721} & PC & Text & Widget & Academia & \href{https://github.com/bpm-diag/smartRPA}{Yes} & 22.11 \\
\midrule

\multirow{7}{*}{\makecell[c]{\textit{\textbf{Prompt-based}}\\ \textit{\textbf{Methods}}}}
& Reflexion ~\cite{NEURIPS2023_1b44b878} & General & Text & Widget & Academia & \href{https://github.com/noahshinn/reflexion}{Yes} & 23.10 \\
& RCI ~\cite{kim2023language} & Web & Text & Widget & Academia & \href{https://github.com/posgnu/rci-agent}{Yes} & 23.11 \\
& AppAgent ~\cite{zhang2023appagent} & Mobile & Hybrid & Widget & Industry & \href{https://github.com/TencentQQGYLab/AppAgent}{Yes} & 24.04 \\
& Mobile-Agent ~\cite{wang2024mobileagent} & Mobile & Hybrid & Widget & Academia & \href{https://github.com/X-PLUG/MobileAgent/tree/main/Mobile-Agent-v1}{Yes} & 24.04 \\
& Mobile-Agent-V2 ~\cite{wang2024mobileagentv2} & Mobile & Hybrid & Atomic & Academia & \href{https://github.com/X-PLUG/MobileAgent/tree/main/Mobile-Agent-v2}{Yes} & 24.06 \\
& MobileGPT ~\cite{lee2024mobilegpt} & Mobile & Text & Atomic & Academia & \href{https://github.com/mobilegptsys/MobileGPT}{Yes} & 24.11 \\
& Mobile-Agent-V ~\cite{wang2025mobile} & Mobile & Hybrid & Widget & Academia & No & 25.06 \\
\midrule

\rowcolor{black!5}\multicolumn{8}{c}{\textit{\textbf{Model-native Paradigm}}}\\
\midrule

\multirow{9}{*}{\makecell[c]{\textit{\textbf{Modular}}\\ \textit{\textbf{Training}}}}
& CogAgent ~\cite{10655402} & General & Visual & Atomic & Academia & \href{https://github.com/zai-org/CogAgent}{Yes} & 23.12 \\
& UGround ~\cite{gou2025navigating} & General & Visual & Atomic & Academia & \href{https://github.com/OSU-NLP-Group/UGround}{Yes} & 24.10 \\
& CoAT ~\cite{zhang-etal-2024-android} & Mobile & Hybrid & Widget & Academia & \href{https://github.com/IMNearth/COAT}{Yes} & 24.11 \\
& WEPO ~\cite{10.1609/aaai.v39i25.34863} & Web & Text & Widget & Academia & \href{https://github.com/KLGR123/WEPO}{Yes} & 24.12 \\
& STEVE ~\cite{lu2025steve} & PC & Hybrid & Widget & Academia & \href{https://github.com/FanbinLu/STEVE}{Yes} & 25.03 \\
& Explorer ~\cite{pahuja-etal-2025-explorer} & Web & Hybrid & Widget & Academia & \href{https://osu-nlp-group.github.io/Explorer/}{Yes} & 25.05 \\
& OS-Genesis ~\cite{sun-etal-2025-os} & General & Hybrid & Widget & Academia & \href{https://qiushisun.github.io/OS-Genesis-Home/}{Yes} & 25.06 \\
& Aria-UI ~\cite{yang-etal-2025-aria} & General & Visual & Atomic & Academia & \href{https://ariaui.github.io}{Yes} & 25.07 \\
& V-Droid ~\cite{dai2025advancing} & Mobile & Text & Widget & Industry & \href{https://github.com/V-Droid-Agent/V-Droid}{Yes} & 25.09 \\
\midrule

\multirow{12}{*}{\makecell[c]{\textit{\textbf{End-to-end}}\\ \textit{\textbf{Training}}}}
& SeeClick ~\cite{cheng2024seeclickharnessingguigrounding} & General & Visual & Atomic & Academia & \href{ https://github.com/njucckevin/SeeClick}{Yes} & 24.02 \\
& MobileFlow ~\cite{nong2024mobileflowmultimodalllmmobile} & Mobile & Visual & Atomic & Industry & No & 24.07 \\
& UI-TARS ~\cite{qin2025ui_tars} & General & Visual & Atomic & Industry & \href{https://github.com/bytedance/UI-TARS}{Yes} & 25.01 \\
& GUICourse ~\cite{chen-etal-2025-guicourse} & General & Visual & Atomic & Academia & \href{https://github.com/RUCBM/GUICourse}{Yes} & 25.05 \\
& ZeroGUI ~\cite{yang2025zeroguiautomatingonlinegui} & General & Visual & Atomic & Academia & \href{https://github.com/OpenGVLab/ZeroGUI}{Yes} & 25.05 \\
& ARPO ~\cite{lu2025arpoendtoendpolicyoptimizationgui} & PC & Visual & Atomic & Academia & \href{https://github.com/dvlab-research/ARPO}{Yes} & 25.05 \\
& UItron ~\cite{zeng2025uitronfoundationalguiagent} & General & Visual & Atomic & Industry & \href{https://github.com/UITron-hub/UItron}{Yes} & 25.08 \\
& GUI-Owl ~\cite{ye2025mobileagentv3} & General & Visual & Atomic & Industry & \href{https://github.com/X-PLUG/MobileAgent/tree/main/Mobile-Agent-v3}{Yes} & 25.08 \\
& DART ~\cite{li2025efficientmultiturnrlgui} & PC & Visual & Atomic & Academia & \href{https://computer-use-agents.github.io/dart-gui/}{Yes} & 25.09 \\
&  GUI-R1 ~\cite{luo2025guir1generalistr1style} & General & Visual & Atomic & Academia & \href{https://github.com/ritzz-ai/GUI-R1}{Yes} & 25.10 \\
&  OpenCUA ~\cite{wang2025opencuaopenfoundationscomputeruse} & PC & Visual & Atomic & Academia & \href{https://github.com/xlang-ai/OpenCUA}{Yes} & 25.10 \\
\end{longtable}
} 

\paragraph{\textbf{End-to-end Training}} 
\hspace{2mm}
While modular training advanced individual components, its limitations such as error accumulation and weak generalization became increasingly apparent. To address these challenges, the research focus has shifted toward a higher level of integration with end-to-end training. In this paradigm, a single, unified model learns the complete mapping directly from perceptual inputs (like screen images) to executable actions, eliminating the need for manually designed intermediate modules. The work in this area can be categorized by the training environment into two primary approaches: offline and online training.

\begin{itemize}
    \item \textbf{Offline training.}\quad \setlength{\parindent}{1em}
    In the early exploration of offline end-to-end training, researchers primarily relied on data-driven imitation learning and multimodal pre-training. The objective was for a model to learn the complete mapping from interface perception to action execution within a single, unified structure. 
    
    \indent The initial breakthrough in this area came from GUIDE\cite{chawla2024guidegraphicaluserinterface}, which was the first work to systematically construct an execution-oriented GUI behavioral dataset. It unified interface screenshots, task descriptions, the previous action, and the spatial location of the next action into supervised training samples. This enabled a multimodal model to directly learn the vision-to-action transformation in an offline setting, laying the data foundation for subsequent end-to-end agents. Building on this idea, GUICourse\cite{chen-etal-2025-guicourse} further proposed a systematic training curriculum that included sub-datasets for interface understanding (GUIEnv), action knowledge (GUIAct), and task interaction (GUIChat). Through a process of staged supervised fine-tuning, this curriculum enabled a general-purpose vision-language model to progressively acquire integrated capabilities for perception, planning, and execution. As a result, the model could perform task-level operations without the need for an explicitly designed workflow. More recently, GUI-R1\cite{luo2025guir1generalistr1style} built upon this foundation by introducing a rule-based Reinforcement Fine-Tuning (RFT) mechanism. It performed policy-level optimization on a pre-trained model using a small amount of high-quality offline data. Through a unified action space and a relative reward design, it further improved the stability and generalization of end-to-end execution. 
    
    \indent Overall, this body of work signifies the evolution of offline GUI agents from data-driven imitation learning to policy-level adaptive optimization. In this new paradigm, the model no longer relies on explicit modules or a pipeline structure. Instead, it directly learns the mapping from perception, through planning, to execution within a unified architecture, laying a solid foundation for future online reinforcement learning and hybrid training paradigms.
    
    \item \textbf{Online training.}\quad \setlength{\parindent}{1em}
    Online end-to-end training aims to overcome the coverage limitations of offline data. It enables a GUI agent to autonomously optimize its entire decision-making chain, from visual perception to action execution, through continuous interaction with a real or simulated environment. 
    
    \indent The earliest work in this area, UI-TARS\cite{qin2025ui_tars}, was the first to implement end-to-end online reinforcement learning in a native application environment. It demonstrated the online learnability of a ``pixels-to-actions'' policy by having a model take screenshots as direct input and produce action commands as output, using real interactions and environmental feedback to optimize its policy function. Subsequently, ARPO\cite{lu2025arpoendtoendpolicyoptimizationgui} built upon this foundation by proposing a mechanism that decouples experience replay from policy updates. By converting collected GUI trajectories into reusable samples and introducing a high-quality task filtering strategy, it achieved efficient and stable convergence for end-to-end reinforcement learning. This allowed the model to continuously improve its perception, planning, and execution capabilities through sustained online interaction. ZeroGUI\cite{yang2025zeroguiautomatingonlinegui} further advanced this line of research by breaking the dependence on human supervision. It achieved a ``zero human cost'' online end-to-end training pipeline through automatic task generation, automatic reward signal evaluation, and an adaptive sampling mechanism. This enabled the model to self-explore and correct its policy within a closed loop. In parallel, DART\cite{li2025efficientmultiturnrlgui} significantly improved the efficiency and stability of online training in multi-task, long-sequence environments by introducing a decoupled multi-turn reinforcement learning framework that separates environment exploration from policy updates. Building on these advances, UI-TARS-2\cite{wang2025uitars2technicalreportadvancing} extended the training to multi-turn interactions and task transfer. It constructed a complete multi-task online optimization loop, which enabled the agent to achieve policy generalization and self-evolution over long-term interactions. More recent works, such as UItron\cite{zeng2025uitronfoundationalguiagent}, Mobile-Agent-v3/GUI-Owl\cite{ye2025mobileagentv3} and OpenCUA~\cite{wang2025opencuaopenfoundationscomputeruse}, represent a culmination of this research direction. Both approaches combine reinforcement learning with large multimodal models, allowing a single architecture to handle the three core functions of perception, planning, and execution. By continuously fine-tuning with online reward signals, these agents achieve adaptive, end-to-end capabilities across different platforms and tasks. 
    
    \indent Overall, this line of research clearly illustrates the evolutionary path of online end-to-end GUI agents, progressing from being merely ``interactive'' to ``optimizable,'' and finally to ``self-evolving.'' This marks a fundamental shift in GUI automation, moving from static models to dynamic learning systems.
\end{itemize}

\subsubsection{\textbf{Summary and Discusion}}
\hspace{2mm}
The evolution of GUI Agents can be understood as two major leaps. The first was the shift from system-based workflow to prompt-based methods with the introduction of LLMs, which enabled agents to generate operational plans from natural language. The second, more significant leap, has been the transition to the model-native paradigm. This stage focuses on internalizing the ``perceive-plan-act'' loop through data-driven training, progressing from modular training of separate capabilities to fully end-to-end training. This has culminated in recent methods like UItron\cite{zeng2025uitronfoundationalguiagent} and GUI-Owl\cite{ye2025mobileagentv3}, which leverage RL for continuous, real-world improvement.

Despite this significant progress, the field of GUI Agents is still in its early stages, and several fundamental challenges remain. The two core properties of GUI interaction, fine-grained/low-level and dynamic/non-stationary, give rise to the primary bottlenecks like the scarcity of high-quality labeled data, the insufficient prior knowledge of foundational models in GUI scenarios, and the lack of realistic interactive environments. As emphasized in Ultron\cite{zeng2025uitronfoundationalguiagent}, overcoming these issues requires robust data engineering and a sophisticated interaction infrastructure, both of which are still developing.

Future development will likely continue to focus on the model-native paradigm, with the objective of creating learning-driven, end-to-end GUI agents. Based on recent progress, key research directions are emerging to address the aforementioned challenges:
\begin{itemize}
    \item \textbf{Data efficiency and generalization:} \setlength{\parindent}{1em}
    Future work will aim to mitigate data scarcity issues through techniques such as domain knowledge injection, dense reward design, self-supervised annotation, and superior exploration strategies. The goal is to enable agents to adapt to new interfaces with a minimal number of samples.
    \item \textbf{Enhanced GUI perception: } \setlength{\parindent}{1em}
    This involves developing more fine-grained visual-language encoders for the recognition and localization of complex UI elements, as seen in works like MobileFlow\cite{nong2024mobileflowmultimodalllmmobile}. It also includes fusing screenshots, OCR, and structured metadata to achieve more robust GUI grounding.

    \item \textbf{Scalable evaluation and realistic environment: } 
    The development of unified benchmarks, such as A3\cite{chai2025a3} and MLA-Trust\cite{yang2025mla_trust}, and automated evaluation pipelines is essential. These tools will allow for the systematic assessment of agent capabilities across multiple platforms and tasks, thereby promoting fair comparisons and rapid iteration among different models. Future work in this area will likely move beyond static benchmarks toward creating realistic environments that better simulate the dynamic and non-stationary nature of real-world applications.
\end{itemize}

\begin{figure}[t] 
  \centering 
  \includegraphics[width=0.9\textwidth]{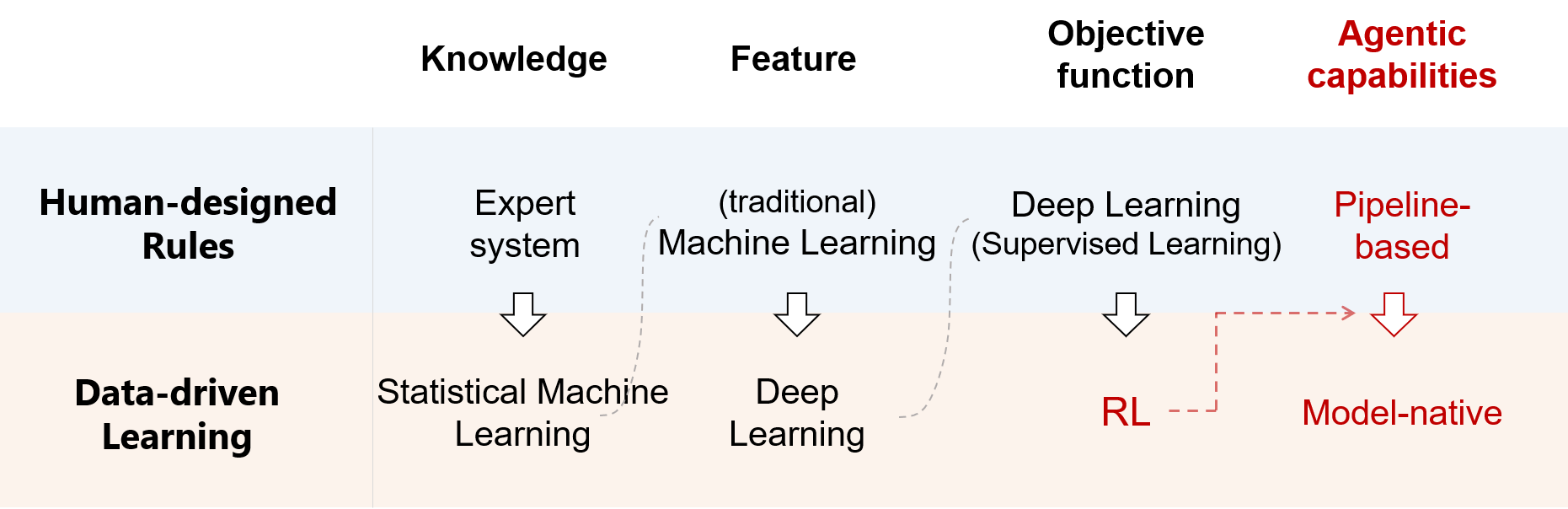} 
  \caption{Paradigm shift in AI: from \textit{human-designed rules} to \textit{data-driven learning}.} 
  \label{fig:trend} 
\end{figure}

\section{Future Direction and Discussion}\label{sec:7}
We have observed the shift in agentic capabilities and the two main forms of agent application, transitioning from pipeline-based to model-native paradigms. This transition actually coincides with the recurring trend from human-designed logic toward data-driven learning that has defined the progress of AI for decades.

As illustrated in Fig.~\ref{fig:trend}, we have witnessed this pattern repeatedly: knowledge acquisition from expert-defined rules to a more automated, data-driven process through statistical machine learning; feature representation from hand-crafted filters in traditional machine learning to deeply layered, automatically learned features in deep learning. The recent evolution from supervised learning to reinforcement learning mirrors a similar pattern: the objective function transformed from fitting explicit, human-provided labels to self-guided exploration and optimization based on environmental feedback. The common drivers behind these repeated transitions are twofold: the inherent cost and scalability limitations of manual design, and the concurrent advancement of computational power and data infrastructure.

In this context, we now turn to the future of Agentic AI by discussing two questions: (1) What other core agentic capabilities are likely to become model-native? (2) As these capabilities are gradually internalized, how will the role of the system layer evolve within agentic AI?

\subsection{Emerging Model-native Agentic Capabilities}
As previously established, the $LLM + RL + Task$ solution provides a promising, unified path to internalize a spectrum of agentic capabilities. Beyond planning, tool use, and memory, we envision a landscape of other agentic capabilities that will progressively transition from pipeline-based to model-native implementations.

\begin{figure}[t] 
  \centering 
  \includegraphics[width=0.95\textwidth]{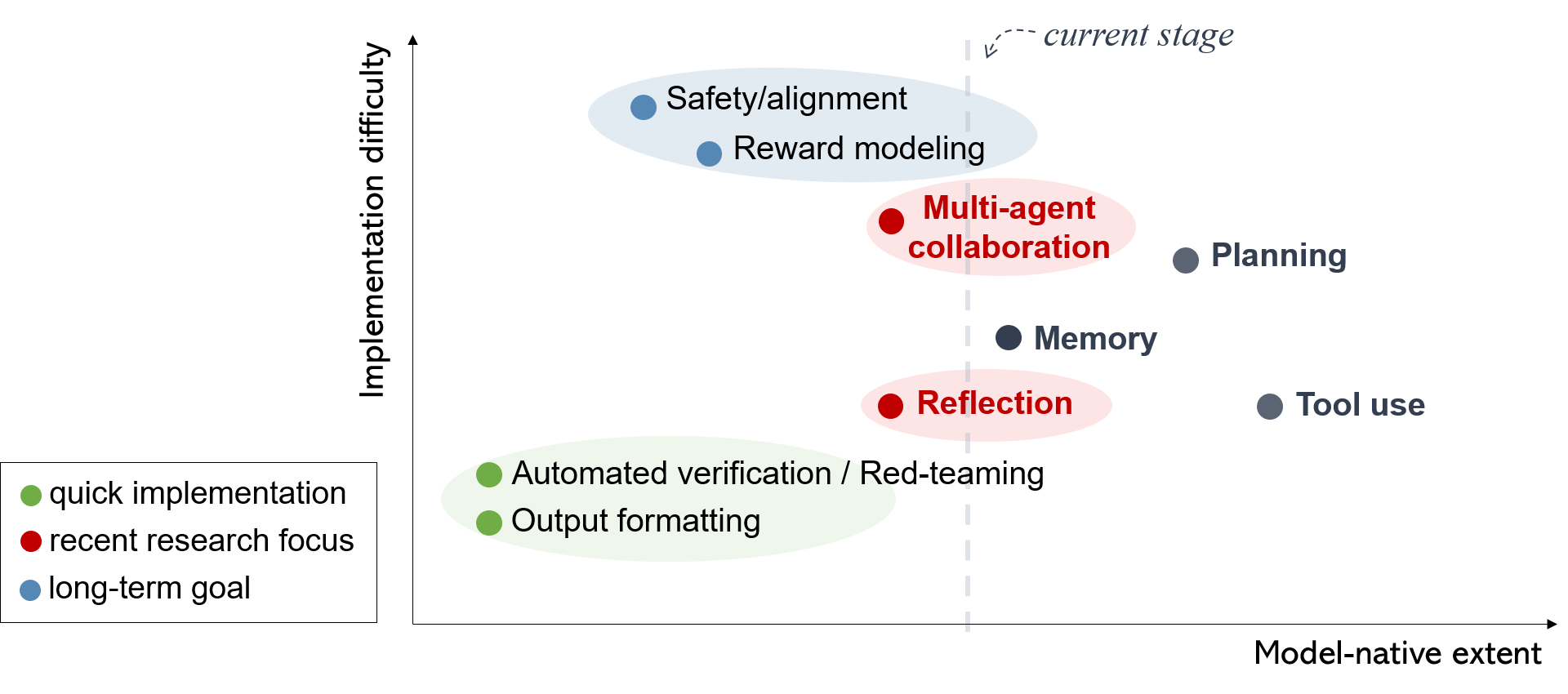} 
  \caption{A roadmap for the model-native internalization of agentic capabilities.} 
  \label{fig:trend2} 
\end{figure}

To illustrate this evolution, Fig.~\ref{fig:trend2} presents a predictive roadmap, categorizing these future capabilities into three groups based on their implementation difficulty and current model-native progress. This landscape spans from quick implementation of internalization (e.g., output formatting) to long-term challenges (e.g., safety/alignment). In the following, we focus on the critical mid-term research frontier, briefly reviewing recent attempts to internalize two key agentic capabilities of \textit{multi-agent collaboration} and \textit{reflection}.

\subsubsection{\textbf{Multi-agent Collaboration}}
Recent advances have transformed LLMs from generative text tools into agentic systems endowed with planning, tool use and memory capabilities. Organizing multiple LLM-based agents into collaborative systems enables the solution of tasks that exceed the capacity of any single agent. However, achieving transferability, robustness and controllability in such collaborations requires a shift from manual orchestration toward internalized, training-time learning. Multi-agent reinforcement learning (MARL) serves as a key catalyst for this transition. It can not only optimize individual agent policies but also learn system-level organizational structures and communication topologies, thereby enabling the shift from pipeline-based paradigm to model-native paradigm.

\afterpage{
\clearpage 
{ 
\footnotesize
\begin{longtable}{l|C{2.0cm}|C{1.8cm}|C{2.5cm}|C{1.0cm}|C{1.0cm}}
\caption{Representative studies on multi-agent collaboration.}\label{tab:mas_overview}\\
\toprule
\rowcolor{blue!5}
\textbf{Method} & \textbf{Task} & \textbf{Algorithm} & \textbf{Affiliation} & \textbf{Access} & \textbf{Date} \\
\midrule
\endfirsthead

\caption[]{Representative studies on multi-agent collaboration (continued). }\\
\toprule
\rowcolor{blue!5}
\textbf{Method} & \textbf{Task} & \textbf{Algorithm} & \textbf{Affiliation} & \textbf{Access} & \textbf{Date} \\
\midrule
\endhead

\bottomrule
\endlastfoot

\rowcolor{black!5}
\multicolumn{6}{c}{\textit{\textbf{Pipeline-based Paradigm}}} \\
\midrule

CAMEL~\cite{li2023camel} & General & — & Academia & \href{https://github.com/camel-ai/camel}{Yes} & 23.03 \\
MetaGPT~\cite{hong2023metagpt} & Programming & — & Academia/Industry & \href{https://github.com/FoundationAgents/MetaGPT}{Yes} & 23.08 \\
MAD~\cite{liang2023encouraging} & Reasoning & — & Academia/Industry & \href{https://github.com/Skytliang/Multi-Agents-Debate}{Yes} & 23.05 \\
MoA~\cite{wang2024mixture} & General & — & Academia/Industry & \href{https://github.com/togethercomputer/moa}{Yes} & 24.06 \\
AFlow~\cite{zhang2024aflow} & Planning & — & Academia/Industry & \href{https://github.com/FoundationAgents/AFlow}{Yes} & 24.10 \\
\midrule

\rowcolor{black!5}
\multicolumn{6}{c}{\textit{\textbf{Model-native Paradigm}}} \\
\midrule

MALT~\cite{motwani2024malt} & Reasoning & SFT+DPO & Academia & No & 24.12 \\
CORY~\cite{ma2024coevolving} & Interaction & PPO-like RL & Academia/Industry & \href{https://github.com/Harry67Hu/CORY}{Yes} & 24.10 \\
MARFT~\cite{liao2025marft} & General & MARFT & Academia/Industry & \href{https://github.com/jwliao-ai/MARFT}{Yes} & 25.04 \\
MAGRPO~\cite{liu2025llm} & General & MAGRPO & Academia & No & 25.08 \\
RLCCF~\cite{yuan2025wisdom} & Reasoning & GRPO & Academia & No & 25.08 \\
MATPO~\cite{mo2025multi} & General & MATPO & Industry & \href{https://github.com/mzf666/MATPO}{Yes} & 25.10 \\
MasHost~\cite{yang2025mashost} & Collaboration & HRPO & Academia & No & 25.06 \\
G-Designer~\cite{zhang2024g} & Topology & VGAE & Academia & \href{https://github.com/yanweiyue/GDesigner}{Yes} & 24.10 \\
ARG-Designer~\cite{li2025assemble} & Topology & ARG & Academia/Industry & \href{https://github.com/Shiy-Li/ARG-Designer}{Yes} & 25.07 \\
MAGDi~\cite{chen2024magdi} & Reasoning & SFT & Academia & \href{https://github.com/dinobby/MAGDi}{Yes} & 24.02 \\
CoA~\cite{li2025chain} & General & SFT + RL & Industry & \href{https://github.com/OPPO-PersonalAI/Agent_Foundation_Models}{Yes} & 25.08 \\

\end{longtable}
} 
} 

\paragraph{\textbf{Pipeline-based Paradigm.}} \hspace{2mm}
Methods in this category orchestrate multi-agent collaboration without modifying the underlying model parameters, instead relying on mechanisms such as prompt engineering, role assignment, and predefined workflow architectures. This hand-crafted orchestration is well-suited for low-cost, rapid-iteration applications. However, its fundamental limitation is the inability to internalize collaborative strategies into the model's parameters, which constrains generalization.

Early multi-agent systems primarily depended on manually defined roles and static interaction protocols. While straightforward to implement for well-defined tasks, these approaches struggle with the variability and long-horizon nature of complex problems. For example, CAMEL~\cite{li2023camel} utilizes a role-playing framework where agents coordinate through dialogue, guided by inception prompting to maintain alignment. Similarly, MetaGPT~\cite{hong2023metagpt} employs an assembly-line structure, decomposing tasks and assigning specialized roles to different agents to reduce error propagation in multi-step processes.

To mitigate the biases inherent in single-chain reasoning, a subsequent line of work introduced mechanisms such as debate and arbitration to foster multi-perspective thinking. While these methods substantially improve the quality of reasoning and question-answering tasks, their effectiveness remains bounded by the design of the prompts and the arbitration strategy itself. For instance, MAD~\cite{liang2023encouraging} formalizes a debate where multiple agents argue iteratively before a judge aggregates the final outcome. MoA~\cite{wang2024mixture} employs a multi-layer agent structure where agents at each level exchange auxiliary inputs to refine the final output.

More recent pipeline-based methods have begun to explore the algorithmic generation of workflows to create more dynamic team structures. AFLOW~\cite{zhang2024aflow}, for example, formulates workflow optimization as a search problem, applying Monte-Carlo Tree Search to explore and refine candidate collaboration structures. Such approaches lay the methodological groundwork for treating the system's architecture itself as an action space, setting the stage for the more adaptive strategies found in the model-native paradigm.

\paragraph{\textbf{Model-native Paradigm.}}
\hspace{2mm}
To overcome the limitations of the pipeline-based paradigm, research has shifted toward incorporating reinforcement learning to internalize collaborative strategies directly into model parameters. This progression has generally occurred in two stages: first, by enhancing the core capabilities of individual agents within a multi-agent system, and second, by elevating the learning objective to the architectural level of the system itself.

Early efforts in this paradigm focused on improving the core reasoning and collaborative capabilities of the agents. For instance, MALT~\cite{motwani2024malt} uses an offline multi-agent RL approach to improve reasoning by decomposing tasks into distinct generator, verifier, and refiner roles. CORY~\cite{ma2024coevolving} employs a two-role replication strategy within a cooperative MARL framework, where a single LLM plays both a ``leader'' and an ``observer'', swapping them during training to improve stability. Building on this, more general frameworks have emerged. MARFT~\cite{liao2025marft} and MAGRPO~\cite{liu2025llm} adapt algorithms like GRPO to the multi-agent setting to improve training stability without requiring individual value networks, while MATPO~\cite{mo2025multi} introduces a principled credit assignment mechanism for training a single model to perform distinct roles.

A more advanced line of research has moved beyond optimizing agent behavior to learning the system's construction and architecture. Works like MasHost~\cite{yang2025mashost}, G-Designer~\cite{zhang2024g}, and ARG-Designer~\cite{li2025assemble} treat the multi-agent system's topology---including the number of agents, their roles, and their communication links---as a learnable action space. By using reinforcement learning or graph generation techniques, these methods can autonomously assemble a bespoke multi-agent system tailored to a specific task, elevating the learning objective from behavior to architecture.

A related approach involves distilling the emergent behaviors of a complex multi-agent system back into a single, powerful model. In this paradigm, collaboration data is first generated through multi-agent interaction and then used to fine-tune a student model. As demonstrated by MAGDi~\cite{chen2024magdi} and Chain-of-Agents~\cite{li2025chain}, the resulting distilled model internalizes multi-role cognitive patterns, allowing it to execute complex, coordinated tasks at inference time without external orchestration.

By leveraging multi-agent reinforcement learning, model-native multi-agent system allows for the direct internalization of collaborative behaviors. However, significant challenges remain. First, the credit assignment problem is exacerbated; with multiple agents acting concurrently, it becomes exceedingly difficult to attribute the final team outcome to the specific actions of any single agent, which can dilute or misdirect the learning signal. Second, the training environment becomes highly non-stationary. From the perspective of any one agent, the environment is constantly changing due to the evolving policies of all other agents. This dynamic instability can destabilize the learning process, as an optimal action at one point in training may become suboptimal as other agents adapt.

To tackle the credit assignment challenge, research will likely focus on more sophisticated attribution techniques, such as developing reward structures that better reflect individual contributions to the collective goal. To manage the non-stationary environment, a key direction will be the development of more robust MARL algorithms that can maintain stable learning amidst shifting policies, potentially incorporating opponent modeling or adaptive learning rates. Ultimately, progress in this domain will depend on creating training methodologies that can effectively navigate the complex interplay between individual contribution and collective success in dynamic, multi-agent settings.

\afterpage{
\clearpage 
{ 
\footnotesize
\begin{longtable}{l|C{2.6cm}|C{1.3cm}|C{1.8cm}|C{2.1cm}|C{0.8cm}|C{0.7cm}}
\caption{Representative studies on agent reflection.}\label{tab:long_paradigm_en}\\
\toprule
\rowcolor{blue!5}
\rowcolor{blue!5}
\textbf{Method} & \textbf{Task} & \textbf{Algorithm} & \textbf{Data Source} & \textbf{Affiliation} & \textbf{Access} & \textbf{Date} \\
\midrule
\endfirsthead

\caption[]{Representative studies on agent reflection (continued). }\\
\toprule
\rowcolor{blue!5}
\textbf{Method} & \textbf{Task} & \textbf{Algorithm} & \textbf{Data Source} & \textbf{Affiliation} & \textbf{Access} & \textbf{Date} \\
\midrule
\endhead

\bottomrule
\endlastfoot

\rowcolor{black!5}
\multicolumn{7}{c}{\textit{\textbf{Pipeline-based Paradigm}}} \\
\midrule

Reflexion~\cite{NEURIPS2023_1b44b878} & Interaction/Coding & — & — & Academia & \href{https://github.com/noahshinn/reflexion}{Yes} & 23.03 \\
Self-Refine~\cite{madaan2023self} & Text Gen./Reasoning & — & — & Academia/Industry & \href{https://selfrefine.info/}{Yes} & 23.03 \\
Critic~\cite{gou2023critic} & Reasoning & — & — & Academia/Industry & \href{https://github.com/microsoft/ProphetNet/tree/master/CRITIC}{Yes} & 23.05 \\
Conf. Matters~\cite{li2024confidence} & Reasoning & — & — & Academia & \href{https://github.com/MBZUAI-CLeaR/IoE-Prompting}{Yes} & 24.02 \\
SCoRe~\cite{kumar2024training} & Math/Coding & REINFORCE & Self-gen. & Industry & No & 24.09 \\
CoV~\cite{he2024retrieving} & Open-domain QA & SFT & Self-gen. + Ext. & Academia/Industry & No & 24.10 \\
DPSDP~\cite{yuan2025reinforce} & Math/Reasoning & SFT+DPO & Self-gen. + Ext. & Academia & No & 25.06 \\
\midrule

\rowcolor{black!5}
\multicolumn{7}{c}{\textit{\textbf{Model-native Paradigm}}} \\
\midrule

Agent-R~\cite{yuan2025agent} & Interaction/Agent Task & SFT & Self-gen. & Industry & \href{https://github.com/bytedance/Agent-R}{Yes} & 25.01 \\
AgentRefine~\cite{fu2025agentrefine} & Interaction/Agent Task & SFT & Ext. & Academia/Industry & \href{https://agentrefine.github.io/}{Yes} & 25.01 \\
STeP~\cite{chen2025training} & Interaction/Agent Task & SFT & Self-gen. + Ext. & Academia/Industry & No & 25.05 \\
KnowSelf~\cite{qiao2025agentic} & Interaction/Agent Task & SFT+RPO & Self-gen. + Ext. & Academia/Industry & \href{https://github.com/zjunlp/KnowSelf}{Yes} & 25.04 \\

\end{longtable}
} 
} 

\subsubsection{\textbf{Reflection}}
Agent reflection is also undergoing a systematic shift from reliance on hand-crafted, run-time pipelines to the internalization of reflective processes. Early approaches used prompt engineering, role assignment, and fixed pipelines to trigger self-checks at inference time. These methods are easy to prototype but limited by templates. Subsequent work introduced supervised and reinforcement learning to improve the reliability of these external modules. Recent research focuses on making reflection an intrinsic behavior of policy so that models can autonomously assess, verify, and correct their outputs without external orchestration.

\paragraph{\textbf{Pipeline-based Paradigm.}} \hspace{2mm}
In early work on reflective behavior, researchers found that prompt engineering and fixed pipelines can elicit self-checking and iterative revision in LLMs without changing model parameters. Representative studies such as Reflexion~\cite{NEURIPS2023_1b44b878}, Self-Refine~\cite{madaan2023self}, Critic~\cite{gou2023critic} and Confidence Matters~\cite{li2024confidence} implement reflection by verbalizing the reflection process and iterating loops such as generate, self-evaluate and revise. These methods demonstrate that pipeline-based reflection can reduce errors and improve robustness on tasks such as question answering, code debugging, and factual verification. However, pipeline-based approaches often depend heavily on the base model’s internal evaluation abilities, e.g., the reliability of confidence estimates. They also rely on manually designed templates and templates scale poorly as task complexity or interaction length increases. Crucially, pipelines do not embed successful collaborative strategies into model parameters, which limits long-term generalization.

To address the robustness and generalization limits of early pipeline-based paradigms, researchers began to introduce training signals that improve a model’s ability to execute a prescribed pipeline. Typical approaches generate high-quality self-correction examples or reward signals via online or offline reinforcement learning, preference learning, or adversarial data synthesis. For instance. SCoRe~\cite{kumar2024training} uses multi-turn RL and regularization to strengthen self-correction. Chain-of-Verification~\cite{he2024retrieving} augments retrieval-augmented generation with verify-and-rewrite chains to align retrieval and generation. DPSDP~\cite{yuan2025reinforce} formalizes multi-turn improvement as a Markov decision process and applies actor-critic style policy search. These studies show that training can substantially broaden the applicability and effectiveness of external pipelines. Still, such methods treat reflection as an external module: they remain constrained by the pipeline’s predefined structure, and they face practical bottlenecks in the cost and quality of the data required for training.

\paragraph{\textbf{Model-native Paradigm.}} \hspace{2mm}
Recently, some pioneering studies have begun to explore the internalization of reflection, aiming to make self-correction an intrinsic component of the model's generation policy. For example, Agent-R~\cite{yuan2025agent} builds training examples by iteratively self-training and using tree search to convert erroneous trajectories into corrected ones. AgentRefine~\cite{fu2025agentrefine} and STeP~\cite{chen2025training} enhance generalization by fine-tuning on partially masked self-reflection trajectories, teaching the model to identify and fix its own mistakes. Furthermore, KnowSelf~\cite{qiao2025agentic} emphasizes situational awareness, enabling the agent to dynamically regulate its use of knowledge and verification procedures based on the context of the task.

In summary, the model-native paradigm for reflection represents a critical shift toward greater agent autonomy. By internalizing the self-correction process, these methods have the potential to improve inference efficiency and long-term robustness. However, this direction also introduces significant challenges, including higher demands on the design of training data, the complexity of reward and value modeling, and the need to maintain interpretability. Future work will likely focus on developing more efficient methods for training reflective capabilities, potentially by exploring self-supervised techniques to reduce the dependency on explicit error labels. A key research frontier will be to investigate the trade-off between the computational cost of performing these internal self-checks and the resulting gains in task performance and reliability.

\subsection{System vs. Model: Evolving Roles in Agentic AI}
As we explore the ongoing shift from pipeline-based to model-native agentic systems, a fundamental divergence emerges between the optimistic visions within academic research and the engineering-centric industry. While researchers eagerly anticipate a future where agentic capabilities are fully internalized in models, with intelligence becoming intrinsic to the system, many industry practitioners still believe that production-ready agentic systems are ``90\% engineering, 10\% intelligence.''~\cite{gohel2024agents} This pragmatic view reflects the current reality of practical systems, which treat the AI as a functional component or a specialized tool, rather than as the central driver of the system itself.

This conflict is actually a common phenomenon in the early stage of a technological revolution. History offers a compelling parallel. The  architectural evolution of Internet clearly demonstrates a dynamic shift in the balance between engineering and intelligence: from an early, purely engineering stage of hand-coded HTML/CSS to a dynamic interaction stage where applications like search engines began to incorporate statistics-based intelligence. This has finally culminated in the present day, where cloud services and MLOps have standardized the underlying engineering, allowing AI-driven personalization and recommendation to become the core value of web applications.

The evolution of web applications suggests that the current engineering-heavy agentic AI is transitory. As the underlying technologies discussed in this survey mature, the applications built upon them will likely follow a similar path. We envision the development of these agent systems in three stages:
\begin{itemize}
    \item \textbf{Pipeline design} ($2023–2025$). In the current era, LLMs are often treated as components, which are callable functions within a human-designed pipeline. 
    \item \textbf{Model-native transition} ($2025–2027$). We are now entering a transitional phase where the model is increasingly becoming the driving force behind core agentic functions like planning, tool use, and memory. At the same time, task automation is evolving from procedural process automation toward more sophisticated levels of intent-driven and role-level autonomy.

During this period, scaffolding frameworks such as LangChain, AutoGen, AgentCore and Agent Builder will continue to play an important role. They will act as intermediate layers that bridge the gap between model-native intelligence and real-world deployment.

    \item \textbf{Autonomous evolution} ($2027-$). This stage anticipates the maturation of the engineering stack into a standardized and automated practice, or \textit{AgentOps}. The focus will shift to fostering highly autonomous agents, enabling agents exhibiting capabilities such as emergent task discovery, on-demand AI-generated software, and even architectural self-evolution.
\end{itemize}

\begin{figure}[t]
 \centering
\includegraphics[width=0.98\textwidth]{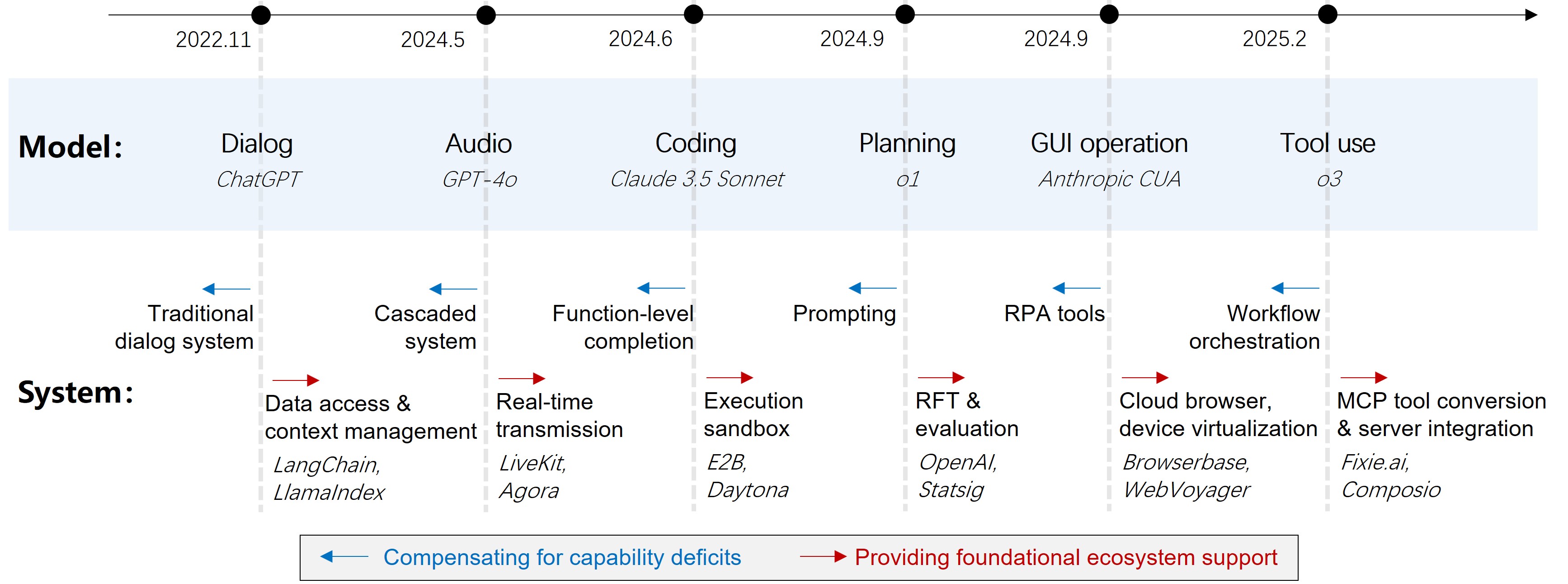}
\caption{The evolving role of the system layer (2022-2025): from \textit{capability compensation} to \textit{ecosystem support}}
 \label{fig:model_system_evolution_past}
\end{figure}

As agentic models become more self-sufficient, the role of the system layer is undergoing a critical transformation: from \textit{compensating for capability deficits} to \textit{providing foundational ecosystem support}. This transition is actually a recurring pattern which is evident in the development of LLM capabilities over the past few years, as illustrated in Fig.~\ref{fig:model_system_evolution_past}. At each stage, as a new capability becomes model-native, the system layer's role evolves from ``compensator'' to ``supporter'':

\begin{itemize}
    \setlength{\itemsep}{2pt}
    \setlength{\parskip}{0pt}
    \item \textbf{Dialogue}: Traditional dialogue systems often split into modules like intent understanding, dialogue management, and response generation. With the rise of ChatGPT, the model layer internalized this entire flow. Consequently, the system layer's role shifted from capability compensation to ecosystem support: providing external data access and context management, exemplified by frameworks like LangChain and LlamaIndex.

    \item \textbf{Planning }:  As models like \textit{o1} began to internalize reasoning through RL, the system layer's role transformed to support this new training paradigm by providing robust Reinforcement Fine-Tuning (RFT), and evaluation frameworks, such as those from OpenAI and Statsig.

    \item \textbf{Tool Use }: As models like o3 internalize the ability to plan and execute tool calls, the system layer's role moves away from workflow orchestration and toward providing a standardized ecosystem for tool integration, such as MCP-based tool transformation and scalable server integration (e.g., Fixie.ai, Compositio).
\end{itemize}

\begin{figure}[t]
 \centering
\includegraphics[width=0.92\textwidth]{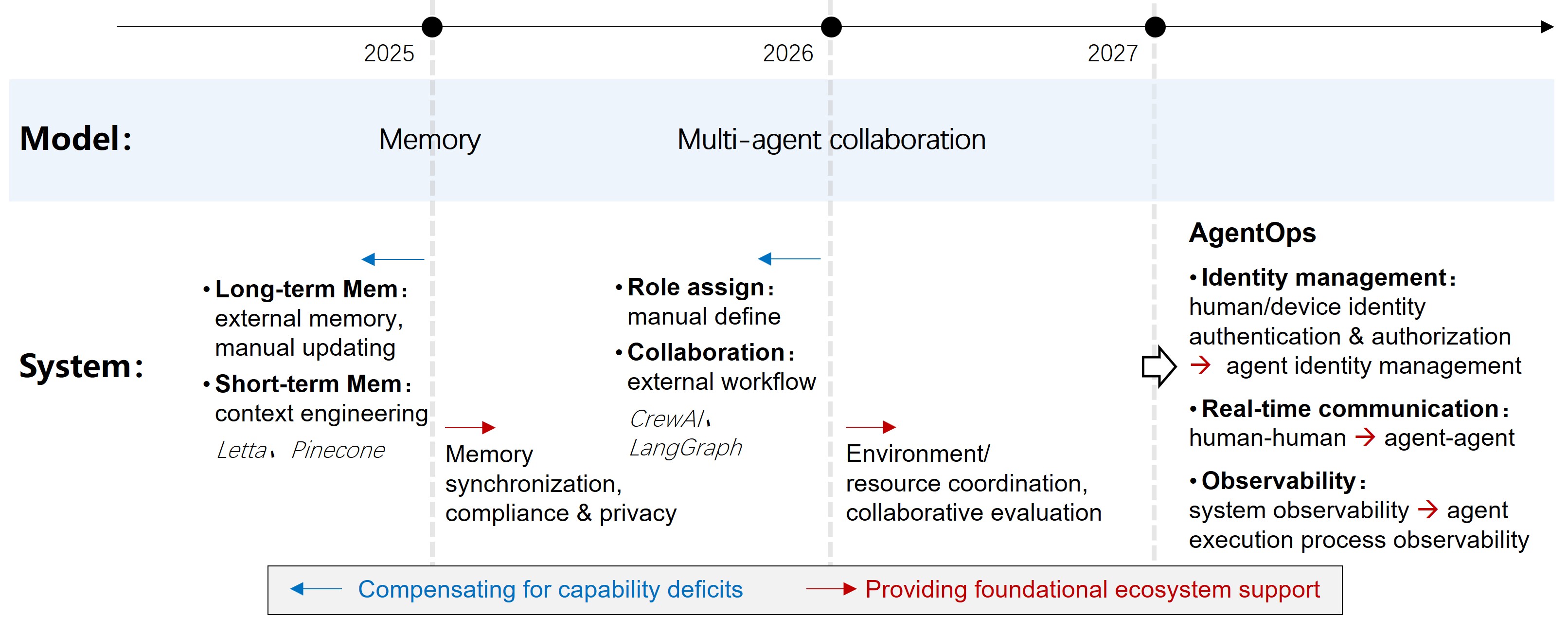}
\caption{The future role of the system layer (2025-): \textit{ecosystem support} and \textit{foundational AgentOps}}
 \label{fig:model_system_evolution_future}
\end{figure}

This established pattern provides a view for understanding the system layer's future evolution, which will be driven by the continued internalization of other agentic capabilities, as illustrated in Fig.~\ref{fig:model_system_evolution_future}. The next stages of this evolution are likely to be:

\begin{itemize}
    \setlength{\itemsep}{2pt}
    \setlength{\parskip}{0pt}
    \item \textbf{Memory:} The system layer will shift from its current role of compensating for the model's statelessness with external vector databases and context engineering (e.g., Letta, Pinecone). It will transition to supporting the model's native memory by providing a foundational ecosystem for cross-platform memory synchronization, data compliance, and privacy controls.

    \item \textbf{Multi-Agent Collaboration:} The system layer will move beyond compensating for a lack of collaboration capability with pre-defined, handcrafted roles and externally orchestrated workflows (e.g., CrewAI, LangGraph). It will transform to support emergent collaboration by providing essential services like environment and resource scheduling, managing inter-agent communication, and offering a platform for collaborative evaluation.
\end{itemize}

\indent Ultimately, as core agentic capabilities are progressively internalized, the system layer will evolve from its current state as a collection of capability-specific patches into a foundational \textit{AgentOps} infrastructure. This future system layer will focus on providing domain-agnostic, scalable services essential for a robust agent ecosystem, evolving existing infrastructure to meet new, autonomous demands:
\begin{itemize}
    \setlength{\itemsep}{2pt}
    \setlength{\parskip}{0pt}
    \item \textbf{Agent Identity Management:} The paradigm for human/device identity and authorization (e.g., Okta) will be extended to manage agent identity, enabling secure, auditable authentication for autonomous entities.
    
    \item \textbf{Agent Communication:} The infrastructure for human-to-human communication (e.g., Twilio) will evolve to support high-bandwidth, structured agent-to-agent communication protocols.
    
    \item \textbf{Agent Execution Observability:} The focus will expand from traditional system-level observability (e.g., Datadog) to the much more complex challenge of \textit{agent execution process observability}, which involves tracing and debugging the emergent, non-deterministic decision-making paths of autonomous agents.
\end{itemize}

\section{Conclusions}
The evolution of agentic AI reflects a deeper transformation in how intelligence itself is conceived, trained, and deployed. From \textit{pipeline-based} systems, where reasoning, memory, and action were orchestrated by external scaffolds, to \textit{model-native} paradigms that internalize these capabilities, we are witnessing a fundamental redefinition of agentic AI.
Reinforcement learning, acting as the engine of experience, bridges perception and action, turning static models into adaptive, goal-directed entities capable of learning from interactions with the environment.

Through this survey, we have reviewed how planning, tool use, and memory are progressively being absorbed into the model’s intrinsic policy. The unifying principle $LLM + RL + Task$ is emerging as the methodological singularity of modern AI. This framework is transforming compute into intelligence through cycles of pretraining, post-training, and inference.

Ultimately, the trajectory of agentic AI is not merely toward greater autonomy, but toward a deeper synthesis between models and the environments they inhabit. Therefore, the paradigm shift from external pipeline to model-native marks a transition from building systems that use intelligence to systems that grow intelligence. The next era of AI will be defined less by how we design agents, and more by how we enable them to learn, collaborate, and evolve through experience.


\bibliographystyle{ACM-Reference-Format}
\bibliography{sample-base}



\end{document}